\pdfoutput=1
\documentclass[11pt]{article}

\usepackage{authblk}
\usepackage[]{acl}

\usepackage{times}
\usepackage{latexsym}
\usepackage{amsmath}
\usepackage{multicol}
\usepackage{booktabs}
\usepackage{dcolumn}
\usepackage{CJKutf8}
\usepackage{enumitem}
\usepackage{cleveref}
\usepackage{graphicx}
\usepackage{hyperref}
\usepackage{subfigure}
\usepackage{subcaption}
\usepackage[normalem]{ulem}
\PassOptionsToPackage{table,xcdraw}{xcolor}
\useunder{\uline}{\ul}{}

\definecolor{mycolor}{RGB}{1,149,255}
\definecolor{lightgray}{RGB}{200,200,200}
\newcommand{\texten}[1]{\textcolor{black}{#1}}
\newcommand{\textzh}[1]{\textcolor{gray}{\tiny \textbf{#1}}}
\newcommand{\textblue}[1]{\textcolor{mycolor}{\textbf{#1}}}
\usepackage[T1]{fontenc}
\usepackage[utf8]{inputenc}

\usepackage{microtype}

\title{Crafting a Good Prompt or Providing Exemplary Dialogues? A Study of In-Context Learning for Persona-based Dialogue Generation
}

\author[1]{Jiashu Pu}
\author[1]{Yajing Wan}
\author[1]{Yuru Zhang}
\author[2]{Jing Chen}
\author[3]{Ling Cheng}
\author[3]{Qian Shao}
\author[1]{\authorcr Yongzhu Chang}
\author[1]{Tangjie Lv}
\author[1,2,*]{Rongsheng Zhang}

\affil[1]{Fuxi AI Lab, NetEase Inc., Hangzhou, China}
\affil[2]{Zhejiang University, Hangzhou, China}
\affil[3]{Singapore Management University, Singapore}
\begin{document}
\maketitle
% \vspace*{1.5cm}
\begingroup\def\thefootnote{*}\footnotetext{Corresponding Author}\endgroup
\begin{abstract}
% In-context learning (ICL) is a new paradigm that has emerged in recent years, where large language models (LLMs) can directly learn from demonstrations (demos) provided in the context through forward propagation.
Previous in-context learning (ICL) research has focused on tasks such as classification, machine translation, text2table, etc., while studies on whether ICL can improve human-like dialogue generation are scarce. Our work fills this gap by systematically investigating the ICL capabilities of large language models (LLMs) in persona-based dialogue generation, conducting extensive experiments on high-quality real human Chinese dialogue datasets. From experimental results, we draw three conclusions: 1) adjusting prompt instructions is the most direct, effective, and economical way to improve generation quality; 2) randomly retrieving demonstrations (\mbox{demos}) achieves the best results, possibly due to the greater diversity and the amount of effective information; counter-intuitively, retrieving demos with a context identical to the query performs the worst; 3) even when we destroy the multi-turn associations and single-turn semantics in the demos, increasing the number of demos still improves dialogue performance, proving that LLMs can learn from corrupted dialogue demos. Previous explanations of the ICL mechanism, such as $n$-gram induction head, cannot fully account for this phenomenon
(Code is available at \href{https://anonymous.4open.science/r/dialog_icl_acl2024}{Dialog ICL}).
\end{abstract}

\section{Introduction}
Current chatbots based on LLMs have superior performance on question answering, polishing documents, etc.~\citep{srivastava2022beyond}. However, for the task of persona-based dialogue generation, these universal LLMs still lag far behind real humans\footnote{As shown in Table~\ref{tab:gpt4_zeroshot_example}, despite numerous specifications, GPT-4 tends to generate unnatural responses in most cases.}. Persona-based dialogue generation is crucial and has practical application value. For instance, optimizing persona dialogue generation is of great significance for empathetic and medical chatbots~\citep{de2020effectiveness,liu2022persona}, and it can bring trust to users~\citep{huang2023personalized,de2020effectiveness}.

% even with the most advanced language models (such as GPT-4~\citep{OpenAI2023GPT4TR}), the performance of language models in persona-based dialogue generation still lags far behind that of real humans\footnote{As shown in Table~\ref{}, despite numerous specifications provided to GPT-4, it still tends to generate rigid and template-like responses in most cases.}. Endowing chatbots with persona is valuable as it can bring trust to users~\citep{huang2023personalized,de2020effectiveness}. 
% % Optimizing persona dialogue generation is of great significance for empathetic and medical chatbots~\citep{de2020effectiveness,liu2022persona}.

On the other hand, fine-tuning a high-quality persona-based dialogue model usually requires a proprietary dataset, and the cost of manually writing dialogues is very high~\citep{cao2022model,huang2023personalized}. Since the valid context length of LLMs is constantly growing~\citep{xiong2023effective}, we believe that ICL may be an effective way to generate high-quality dialogues at a low cost. However, research papers on how ICL affects dialogue generation are few~\citep{dong2022survey}. Only one paper~\citet{xu2023towards} investigates the ability of LLM to learn from dialogue demos, but these demos only contain character background information, which is quite different from real human conversations. In this paper, we experiment with the ICL capability of LLMs on dialogues of real-human conversations, systematically exploring the capabilities of ICL in persona-based dialogue generation, including how we should choose the demo retrieval method, whether we should focus more on the text quality of the demo itself or the input-output mapping, whether providing more demos can improve the results, and give insight on what the LLM learns and does not learn from the provided demos.

\section{Problem Formulation}
We investigate the LLMs' ICL ability to perform turn-level persona-based dialogue generation using the prompt with instructions and demos. We begin by defining notations. We denote the persona description as $\mathit{per}$ and the query context before the $t$-th round of dialogue as $\mathit{c_{:t-1}}$, where $\mathit{c_{:t-1}}=(r_0,\dots,r_{t-1})$ and $r_{t-1}$ is the most recent dialogue turn. A dialogue demo of $\ell$ turns can be represented as $x=(per,c_\ell)$, where $c_{:\ell-1}$ can be interpreted as the input part of the demo while $r_\ell$ as the output ($y$ label) part. A set of $k$ dialogue demos is denoted as $\mathbf{x}^k_{demo}=(x^0,\dots,x^k)$. We define $\mathbf{p}$ as the current task's prompt. Additionally, we have a template function $\mathcal{T}$ for integrating $k$ demos, the task prompt, query persona, and query context. We can represent the LLM's output response at current round $t$ as ${\hat{r_t}=\mathcal{LLM}(\mathcal{T}(\mathbf{x}^k_{demo},\mathbf{p},per,c_{:t-1}))}$. See \S\ref{sec:examples_filled_templates} for filled templates.

Our work primarily investigates three research questions. \textbf{RQ1}: For dialogue generation via ICL, should we focus on tuning the task prompt, providing high-quality demos, or both? \textbf{RQ2}: What is the impact of different demo retrieval methods on ICL? Do the number of demos and their context length make a difference? \textbf{RQ3}: From which aspects do LLMs learn useful information from demos? We analyze this from the perspectives of multi-turn correlation, single-turn semantics, input-label mapping, and token distributions.

\section{Evaluation Metrics for Generation}
In the experiment, given a persona description $per$, a context $c_{:t-1}$ and five expert-written responses, we evaluate a set of model-generated responses from three dimensions: \textbf{intra-diversity}, \textbf{inter-similarity} to expert-written responses, and \textbf{response quality}. Intra-diversity is the average of multiple traditional metrics, including Distinct-n~\cite{distinct}, Entropy-n~\cite{serban2017hierarchical}, self-bleu~\cite{zhu2018texygen}, and cosine similarity based on sentence embeddings. The similarity to expert responses mainly considers rougeL~\cite{lin-2004-rouge}, bleu~\cite{papineni-etal-2002-bleu}, and cosine similarity based on sentence embeddings. The response quality is scored by our self-trained Response Evaluator, which takes into account persona consistency, context logic, interestingness, and colloquial expressions. Our self-trained Response Evaluator significantly outperforms crowd-sourced evaluations and other automated evaluation schemes in terms of consistency with expert ratings. See \S\ref{sec:eval_metrics} for details of how each evaluation metric is calculated and \S\ref{sec:train_response_eval} for how the Response Evaluator is trained.

\section{Different Prompt and ICL Settings}
Below we show settings to concatenate prompts and demos (concrete examples are provided in \Cref{tab:llm_input_example_context_only,tab:llm_input_example_prompt_only,tab:llm_input_example_few_shot_demo_random,tab:llm_input_example_few_shot_demo_random_with_prompt}).

\noindent \textbf{Context Only}: We only provide the persona description and context containing ${t-1}$ turns. The output response is produced as ${\mathcal{LLM}(\mathcal{T}(\mathbf{x^0},\mathbf{p^{\emptyset}},per,c_{:t-1}))}$, where $\mathbf{p}^{\emptyset}$ is an empty string.

\noindent \textbf{Prompt Only}: We only provide the persona description, the task prompt, and the context $c_{:t-1}$. The output response is produced as ${\mathcal{LLM}(\mathcal{T}(\mathbf{x^0},\mathbf{p},per,c_{:t-1})}$.

\noindent \textbf{Few Shot Demo}: We provide $k$ demos, the persona description and the context $c_{:t-1}$. The output response is produced as ${\mathcal{LLM}(\mathcal{T}(\mathbf{x^k},\mathbf{p^{\emptyset}},per,c_{:t-1}))}$.

\noindent \textbf{Few Shot Demo+Prompt}: We provide $k$ demos, the task prompt, the persona, and the context $c_{:t-1}$. The output response is produced as ${\mathcal{LLM}(\mathcal{T}(\mathbf{x^k},\mathbf{p},per,c_{:t-1}))}$.

% \noindent \textbf{Context Only}: We only provide the persona description and context containing ${t-1}$ turns. The output response is produced as ${\mathcal{LLM}(\mathcal{T}(\mathbf{x^0},\mathbf{p^{\emptyset}},per,c_{:t-1}))}$, where $\mathbf{p}^{\emptyset}$ is an empty string.

% \noindent \textbf{Prompt Only}: We only provide the persona description , the task prompt and context containing ${t-1}$ turns. The output response is produced as ${\mathcal{LLM}(\mathcal{T}(\mathbf{x^0},\mathbf{p},per,c_{:t-1})}$.

% \noindent \textbf{Few Shot Demo}: We provide $k$ demos, the persona description and context containing ${t-1}$ turns. The output response is produced as ${\mathcal{LLM}(\mathcal{T}(\mathbf{x^k},\mathbf{p^{\emptyset}},per,c_{:t-1}))}$.

% \noindent \textbf{Few Shot Demo+Prompt}: We provide $k$ demos, the task prompt, the persona description and context containing ${t-1}$ turns. The output response is produced as ${\mathcal{LLM}(\mathcal{T}(\mathbf{x^k},\mathbf{p},per,c_{:t-1}))}$.

For the setting of demo retrieval, we attempt three methods. The first method \textbf{randomly} selects a demo with the same context length, the second method (following~\citep{su2022selective}) retrieves the most similar demos (with the same length as the query) based on the cosine similarity of sentence \textbf{embeddings}, and the third method provides demos that contain the exact \textbf{same} context ($c_{:\ell-1}$) as the query context, each with a high-quality response $r_\ell$ written by a human expert.
 
\section{Experimental Settings}

\begin{table*}[h]
\centering
\small
\begin{tabular}{@{}lllll@{}}
\toprule
 & Prompt Construction Method & Intra-Diversity & Inter-Similarity & Response Quality \\ \midrule
GPT-3.5 & Context Only$^{0}$ & \textbf{0.671} & \textbf{0.269$\uparrow^{1}$} & \textbf{0.147$\uparrow^{4}$} \\
 & Prompt Only$^{1}$ & \textbf{0.684$\uparrow^{0,4}$} & \textbf{0.228} & \textbf{0.191$\uparrow^{0,3,4}$} \\
 & Few Shot Demo (Random)$^{2}$ & \textbf{0.686$\uparrow^{0,4}$} & \textbf{0.277$\uparrow^{1}$} & \textbf{0.160$\uparrow^{4}$} \\
 & \hspace{1em} - w/o correct y label & {$\phantom{-}\Delta$0.008$\dag$} & {$-\Delta$0.040$\dag$} & $\phantom{-}\Delta$0.004 \\
 & \hspace{1em} - w/o contextual semantic & $\phantom{-}\Delta$0.005 & $-\Delta$0.013$\dag$ & {$\phantom{-}\Delta$0.026$\dag$} \\
 & \hspace{1em} - w/o turn semantic & $-\Delta$0.005 & $-\Delta$0.021$\dag$ & $\phantom{-}\Delta$0.010 \\
 & \hspace{1em} - w/o contextual/turn semantic & {$-\Delta$0.007} & {$-\Delta$0.023$\dag$} & {$-\Delta$0.000} \\
 & Few Shot Demo (Emebdding)$^{3}$ & \textbf{0.680$\uparrow^{0,4}$} & \textbf{0.269$\uparrow^{1}$} & \textbf{0.156$\uparrow^{4}$} \\
 & \hspace{1em} - w/o correct y label & $\phantom{-}\Delta$0.006 & $-\Delta$0.020$\dag$ & $-\Delta$0.019 \\
 & Few Shot Demo (Same Persona/Context)$^{4}$ & \textbf{0.667} & {\ul \textbf{0.286$\uparrow^{all}$}} & \textbf{0.094} \\
 & \hspace{1em} - w/o correct y label & $-\Delta$0.014$\dag$ & $-\Delta$0.079$\dag$ & $-\Delta$0.042$\dag$ \\
 & Few Shot Demo (Random) + Prompt$^{5}$ & {\ul \textbf{0.696$\uparrow^{all}$}} & \textbf{0.243$\uparrow^{1}$} & {\ul \textbf{0.222$\uparrow^{all}$}} \\ \midrule
GPT-4 & Context Only$^{0}$ & \textbf{0.655} & \textbf{0.269$\uparrow^{1,2,3}$} & \textbf{0.155} \\
 & Prompt Only$^{1}$ & \textbf{0.704$\uparrow^{all}$} & \textbf{0.248} & {\ul \textbf{0.476$\uparrow^{0,2,3,4}$}} \\
 & Few Shot Demo (Random)$^{2}$ & \textbf{0.679$\uparrow^{0,4}$} & \textbf{0.258$\uparrow^{1}$} & \textbf{0.235$\uparrow^{0,4}$} \\
 & \hspace{1em} - w/o correct y label & {$\phantom{-}\Delta$0.002} & {$-\Delta$0.034$\dag$} & $-\Delta$0.035$\dag$ \\
 & \hspace{1em} - w/o contextual semantic & $\phantom{-}\Delta$0.005 & {$\phantom{-}\Delta$0.001} & {$-\Delta$0.010} \\
 & \hspace{1em} - w/o turn semantic & {$-\Delta$0.018$\dag$} & $-\Delta$0.002 & $-\Delta$0.034$\dag$ \\
 & \hspace{1em} - w/o contextual/turn semantic & $-\Delta$0.008$\dag$ & $-\Delta$0.015$\dag$ & {$-\Delta$0.062$\dag$} \\
 & Few Shot Demo (Emebdding)$^{3}$ & \textbf{0.680$\uparrow^{0,4}$} & \textbf{0.261$\uparrow^{1}$} & \textbf{0.243$\uparrow^{0,4}$} \\
 & \hspace{1em} - w/o correct y label & $\phantom{-}\Delta$0.004 & $-\Delta$0.024$\dag$ & $-\Delta$0.029$\dag$ \\
 & Few Shot Demo (Same Persona/Context)$^{4}$ & \textbf{0.658} & {\ul \textbf{0.294$\uparrow^{all}$}} & \textbf{0.193$\uparrow^{0}$} \\
 & \hspace{1em} - w/o correct y label & $\phantom{-}\Delta$0.003 & $-\Delta$0.070$\dag$ & $-\Delta$0.095$\dag$ \\
 & Few Shot Demo (Random) + Prompt$^{5}$ & {\ul \textbf{0.710$\uparrow^{0,2,3,4}$}} & \textbf{0.264$\uparrow^{1,2}$} & \textbf{0.470$\uparrow^{0,2,3,4}$} \\ \midrule
Ernie & Context Only$^{0}$ & \textbf{0.690} & \textbf{0.265} & \textbf{0.345} \\
 & Prompt Only$^{1}$ & {\ul \textbf{0.720$\uparrow^{0,2,3,4}$}} & \textbf{0.263} & \textbf{0.519$\uparrow^{0,4}$} \\
 & Few Shot Demo (Random)$^{2}$ & \textbf{0.706$\uparrow^{0,3}$} & \textbf{0.269} & \textbf{0.487$\uparrow^{0,4}$} \\
 & \hspace{1em} - w/o correct y label & {$\phantom{-}\Delta$0.023$\dag$} & {$-\Delta$0.025$\dag$} & {$-\Delta$0.075$\dag$} \\
 & \hspace{1em} - w/o contextual semantic & $\phantom{-}\Delta$0.001 & {$\phantom{-}\Delta$0.007} & {$\phantom{-}\Delta$0.007} \\
 & \hspace{1em} - w/o turn semantic & {$-\Delta$0.005$\dag$} & $-\Delta$0.001 & $-\Delta$0.054$\dag$ \\
 & \hspace{1em} - w/o contextual/turn semantic & $-\Delta$0.004 & $-\Delta$0.001 & $-\Delta$0.042$\dag$ \\
 & Few Shot Demo (Emebdding)$^{3}$ & \textbf{0.698$\uparrow^{0}$} & \textbf{0.281$\uparrow^{0,1,2,5}$} & \textbf{0.500$\uparrow^{0,4}$} \\
 & \hspace{1em} - w/o correct y label & $\phantom{-}\Delta$0.012$\dag$ & $-\Delta$0.023$\dag$ & $-\Delta$0.053$\dag$ \\
 & Few Shot Demo (Same Persona/Context)$^{4}$ & \textbf{0.701$\uparrow^{0}$} & {\ul \textbf{0.330$\uparrow^{all}$}} & \textbf{0.340} \\
 & \hspace{1em} - w/o correct y label & $\phantom{-}\Delta$0.009 & $-\Delta$0.120$\dag$ & $-\Delta$0.166$\dag$ \\
 & Few Shot Demo (Random) + Prompt$^{5}$ & \textbf{0.718$\uparrow^{0,2,3,4}$} & \textbf{0.271$\uparrow^{1}$} & {\ul \textbf{0.544$\uparrow^{all}$}} \\ \bottomrule
\end{tabular}
\caption{Results of dialogue generation when few-shot $k$ is set to $5$. For each LLM, the best results are {\ul underlined}. The up-arrow $\uparrow$ indicates statistical significance ($p<0.05$ with Wilcoxon signed-rank test~\cite{woolson2007wilcoxon}) when comparing two methods. For example, \textbf{0.684$\uparrow^{0,4}$} in the second row indicates \emph{Prompt Only} method is better than \emph{Context Only} method and \emph{Few Shot Demo (Same Persona/Context)} method, and is statistically significant. We use $\Delta$ and $-\Delta$ to represent absolute increase and decrease in scores for ablations ($\dag$ indicates statistical significance). The text in parentheses (such as \emph{Random}) represents the demo's retrieval method.}
\label{tab:result_main_creative_k5}
\end{table*}

\subsection{Evaluation LLMs \& Dataset}
For LLMs, we select GPT-3.5-turbo~\cite{ye2023comprehensive}, GPT-4~\cite{achiam2023gpt}, and Ernie~\cite{sun2021ernie} because they have the abilities to follow instructions and perform in-context learning. We employ nucleus sampling~\cite{Holtzman2020The} for decoding, setting both of the top-$p$ and temperature parameters to $0.9$.

Regarding the evaluation dataset $\mathcal{D}_{eval}$, we manually selected $6$ personas with distinct personality backgrounds, with an average of 150 Chinese characters per persona, and an average of 15 turns of conversation. Each turn has an average of 35 characters. All the dialogues are written by students from the Department of Chinese Language and Literature. Compared to previous open-source Chinese dialogue data~\cite{papangelis2020plato}, our persona descriptions are much more complex, and conforming to the majority of commercial chat-bot-based products\footnote{English products include \href{https://beta.character.ai}{character.ai} and \href{https://inworld.ai}{inworld.ai}, while  Chinese products include Minimax's \href{https://new.qq.com/rain/a/20231120A00OOS00}{Xingye}.}, with higher dialogue quality and richer content (see more examples in \S\ref{sec:examples_filled_templates}). We keep the test set compact due to our limited budget (GPT-4 and ernie-bot-4 are very expensive). We also provide a demo dataset $\mathcal{D}_{demo}$ where demos are retrieved (for \textbf{random/embedding} method). It contains $800$ dialogues, with each written by a human expert. $\{\mathcal{D}_{demo}\} \cap \{\mathcal{D}_{eval}\}$ equals $\emptyset$.

\subsection{Ablation Settings}
In order to investigate from which aspects of the demo LLM learned useful information using ICL, we design several ablation settings as below. For concrete examples, please refer to \Cref{tab:demo_example_wo_correct_y_label,tab:demo_example_wo_contextual_semantic,tab:demo_example_wo_turn_semantic,tab:demo_example_wo_contextual_turn_semantic}.

\noindent \textbf{w/o correct y label}: response $r_l$ for each demo context $c_{l-1}$ is replaced with a random response sampled from from $\mathcal{D}_{demo}$. 

\noindent \textbf{w/o contextual semantic}: turns in each demo context $c_{l-1}$ are shuffled. 

\noindent \textbf{w/o turn semantic}: tokens\footnote{each turn is tokenized by \href{https://github.com/fxsjy/jieba}{jieba}.} in each demo turn $r_i$ are shuffled (including the last label turn).

\noindent \textbf{w/o contextual/turn semantic}: tokens in each demo turn $r_i$ and turns in each demo context $c_{:l-1}$ are both shuffled (denoted as $\widetilde{c_{:l-1}}$).

% The reason for choosing these three models is that they exhibit progressive capabilities in Chinese dialogue generation.

\section{Results Analysis}

We present the performance of various settings (few-shot $k$ is fixed to $5$) for dialogue generation in Table~\ref{tab:result_main_creative_k5}. We present the response quality scores averaged over varying few-shot settings in Table~\ref{tab:result_main_creative_k_all}. For more detailed chart data, please refer to \S\ref{sec:appendix}.

\begin{figure*}
    \centering
    \includegraphics[width=0.325\textwidth]{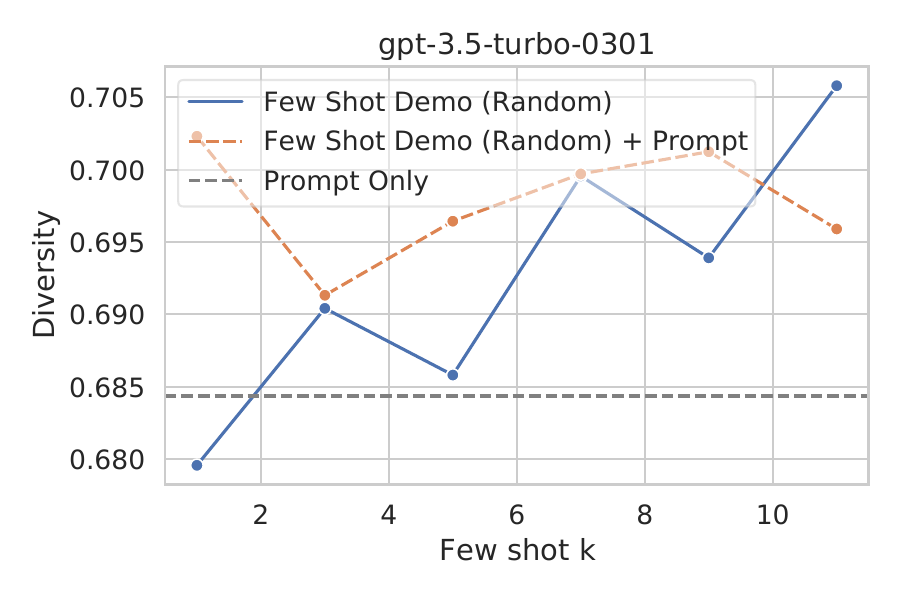}
    \includegraphics[width=0.325\textwidth]{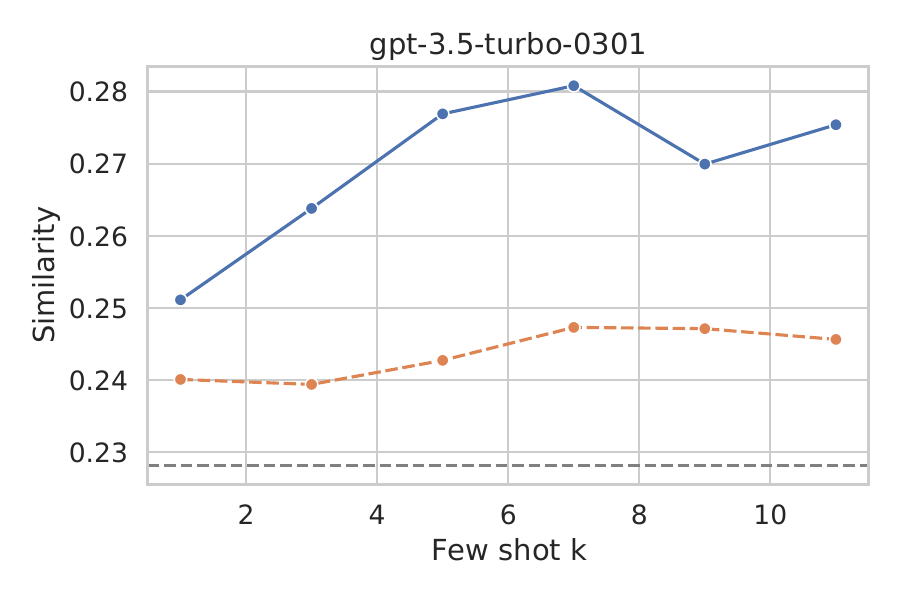}
    \includegraphics[width=0.325\textwidth]{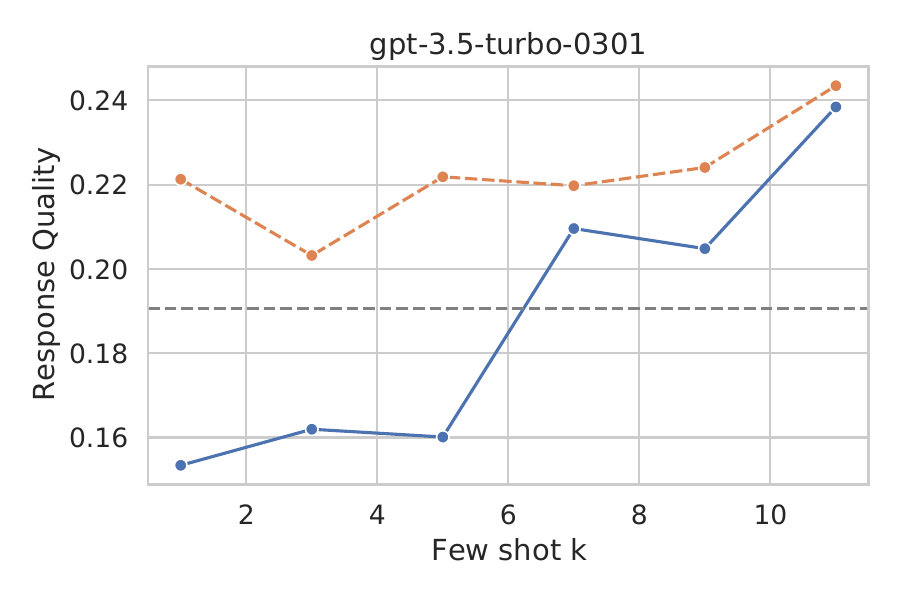}
    \includegraphics[width=0.325\textwidth]{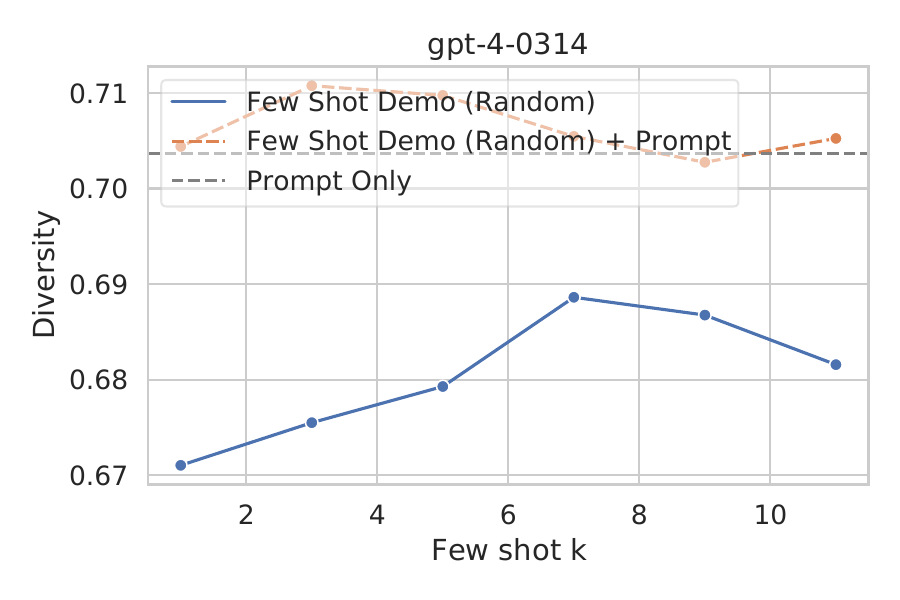}
    \includegraphics[width=0.325\textwidth]{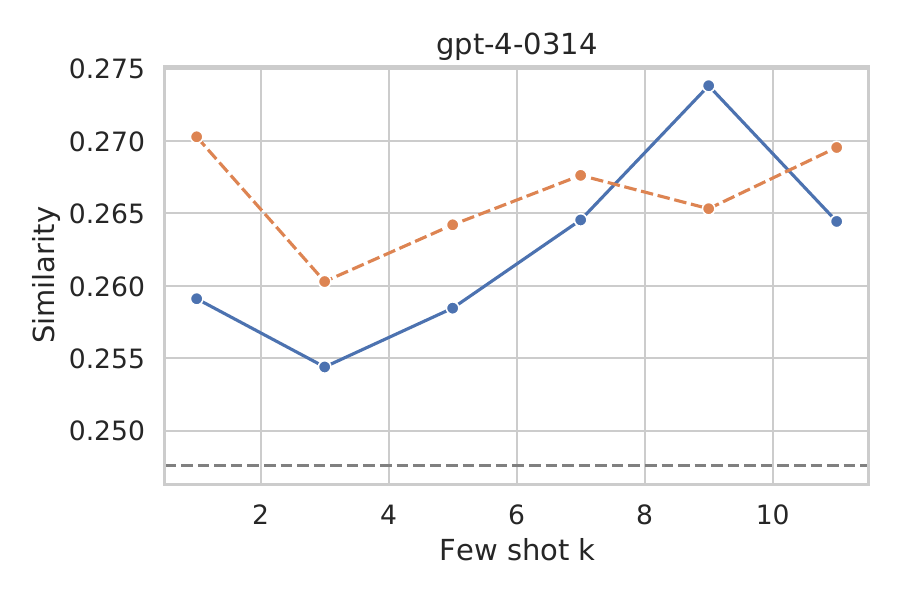}
    \includegraphics[width=0.325\textwidth]{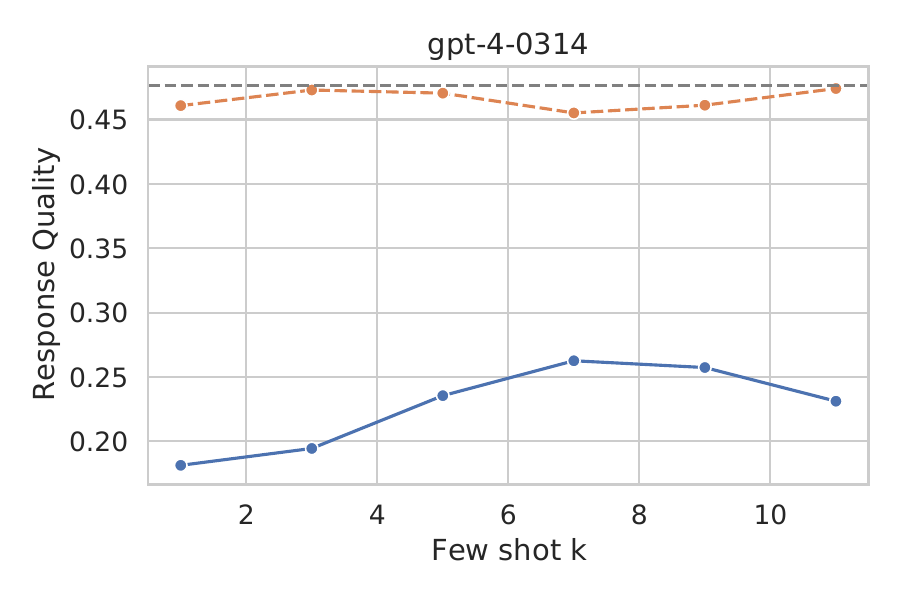}
    \includegraphics[width=0.325\textwidth]{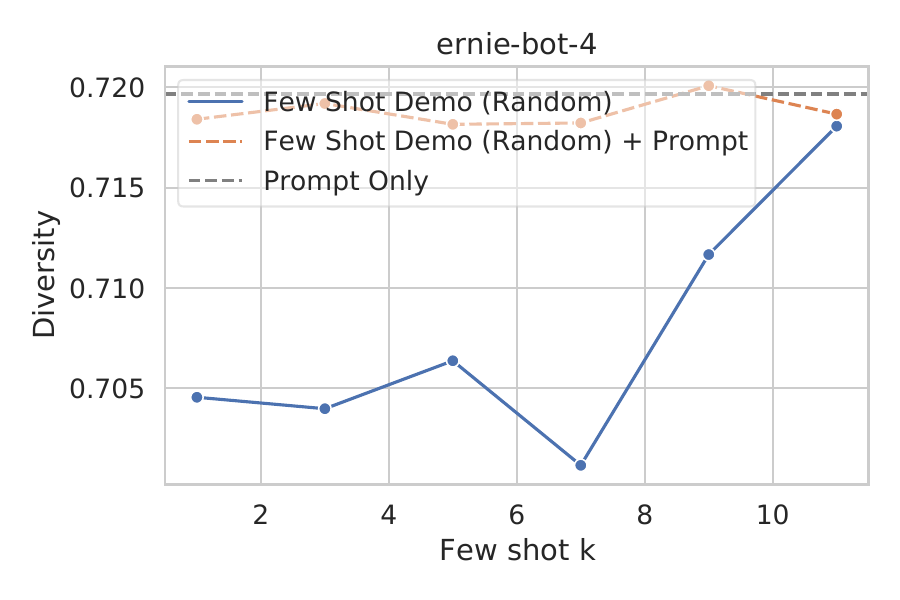}
    \includegraphics[width=0.325\textwidth]{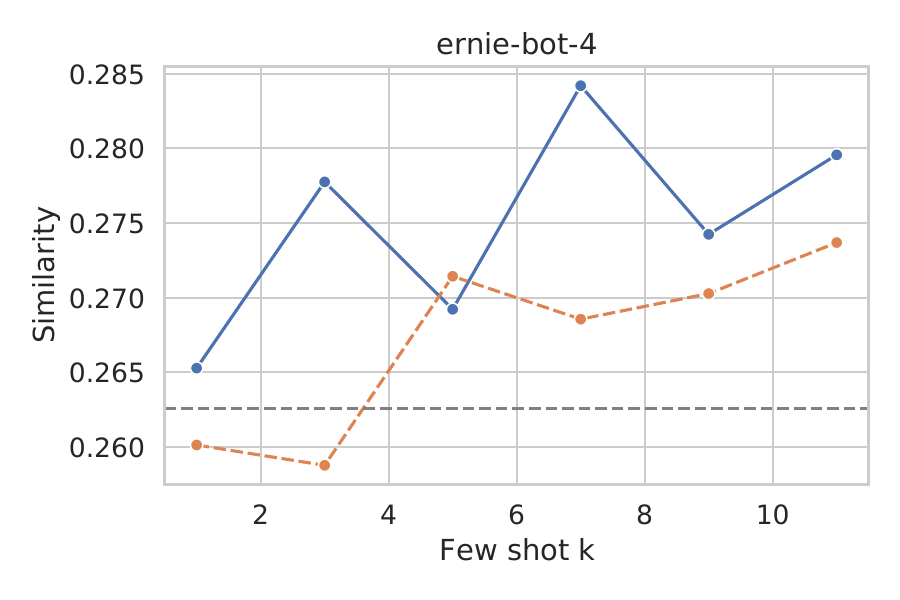}
    \includegraphics[width=0.325\textwidth]{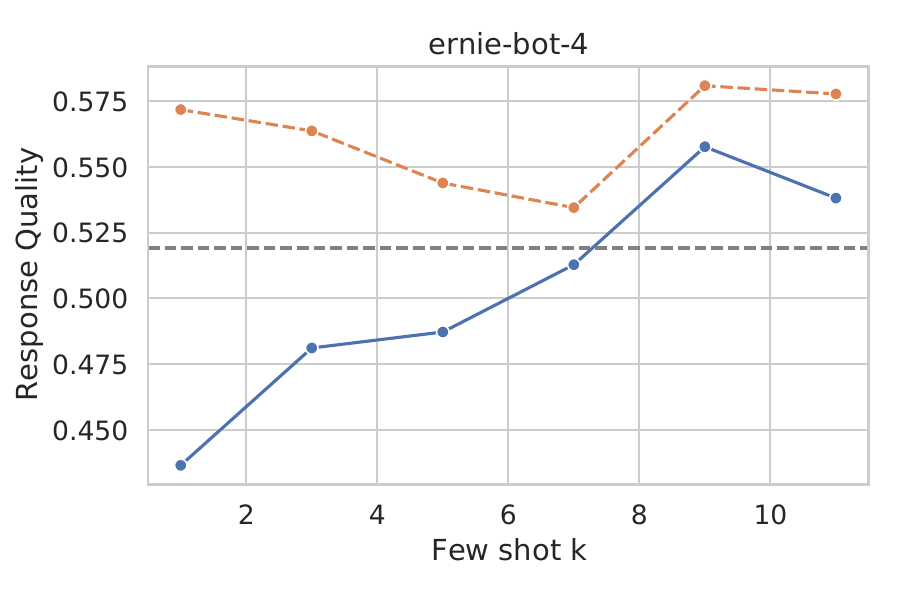}
    \caption{The performance comparison among \emph{Context Only} method, \emph{Prompt Only} method, and \emph{using both prompt and demonstration} when the number of few-shot demonstrations $k$ varies. All values are averaged cross different context lengths.}
    \label{fig:few_shot_context_vs_prompt_vs_demo}
\end{figure*}

\noindent \textbf{Discussion of RQ1}: From Table~\ref{tab:result_main_creative_k5}, we observe that for all LLMs, \emph{Prompt Only} method scores much higher in response quality than using context or few-shot demos alone. From Table~\ref{tab:result_main_creative_k_all} and Figure~\ref{fig:few_shot_context_vs_prompt_vs_demo}, we can see that the few-shot ICL method requires a large number of demos (at least 7) to catch up with using only the prompt. \emph{Overall, we believe that optimizing the prompt alone is the most cost-effective choice. But if funding permits, you may consider adding more demos to the prompt. At least for models gpt-3.5 and Ernie, we find that adding demos can improve the quality of responses.}

\begin{figure}
    \centering
    \includegraphics[width=0.48\textwidth]{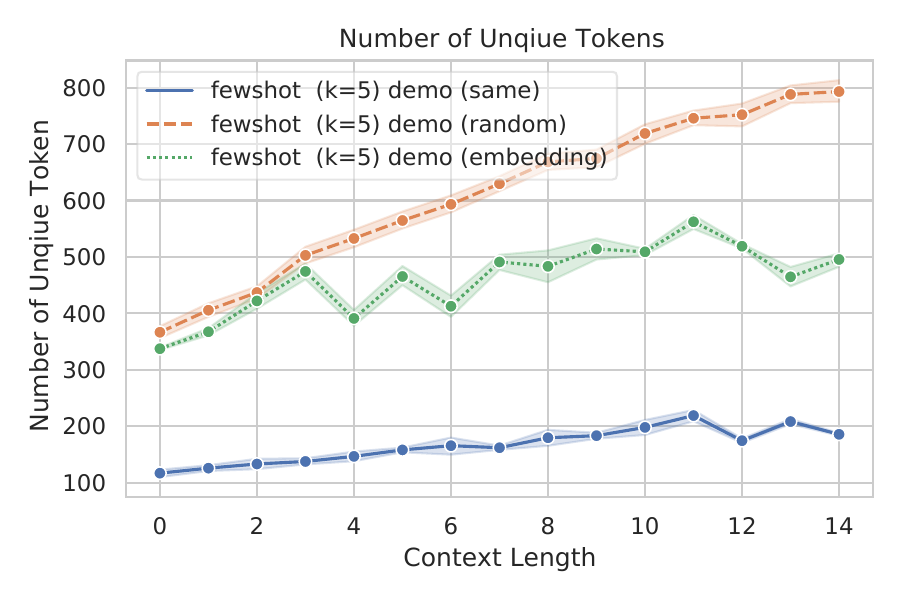}
    \caption{\textbf{X-axis}: length of the demonstration context. \textbf{Y-axis}: number of unique tokens in demonstrations' context for different methods.}
    \label{fig:num_unique_token_for_diff_methods}
\end{figure}

\begin{figure*}
    \centering
    \includegraphics[width=0.325\textwidth]{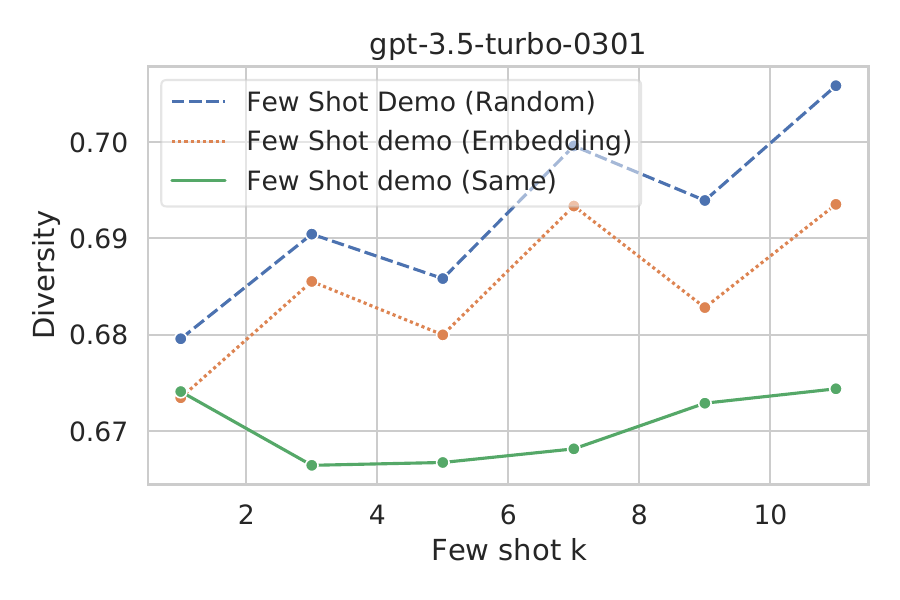}
    \includegraphics[width=0.325\textwidth]{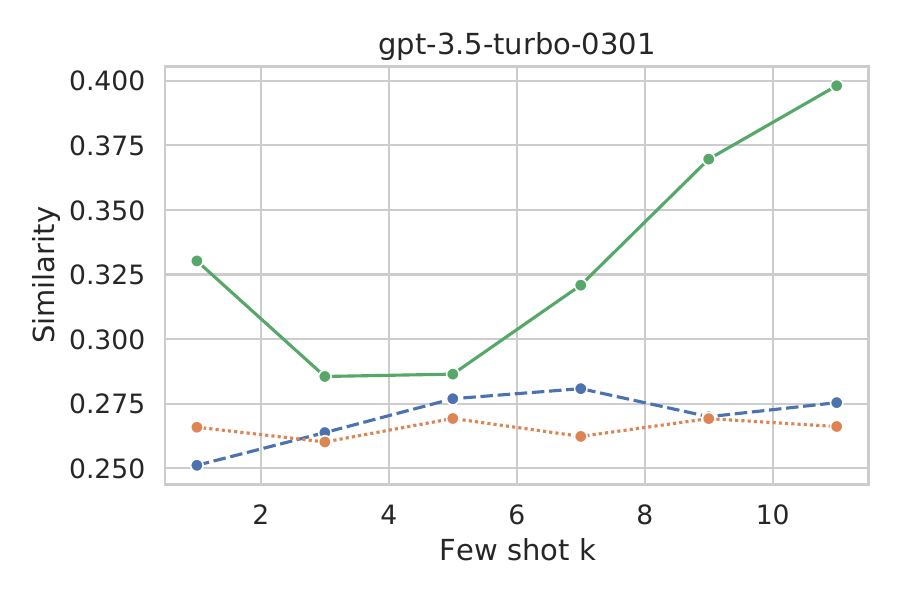}
    \includegraphics[width=0.325\textwidth]{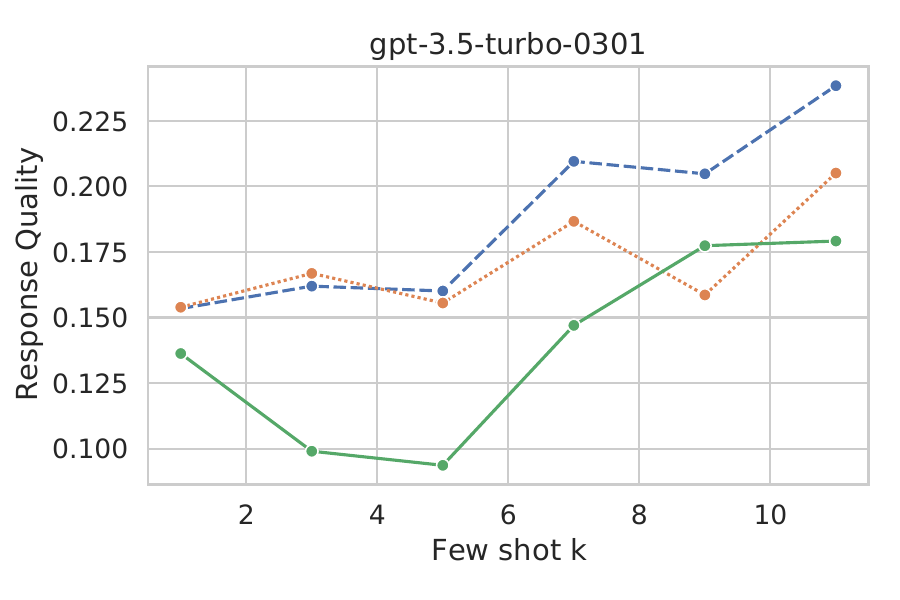}
    \includegraphics[width=0.325\textwidth]{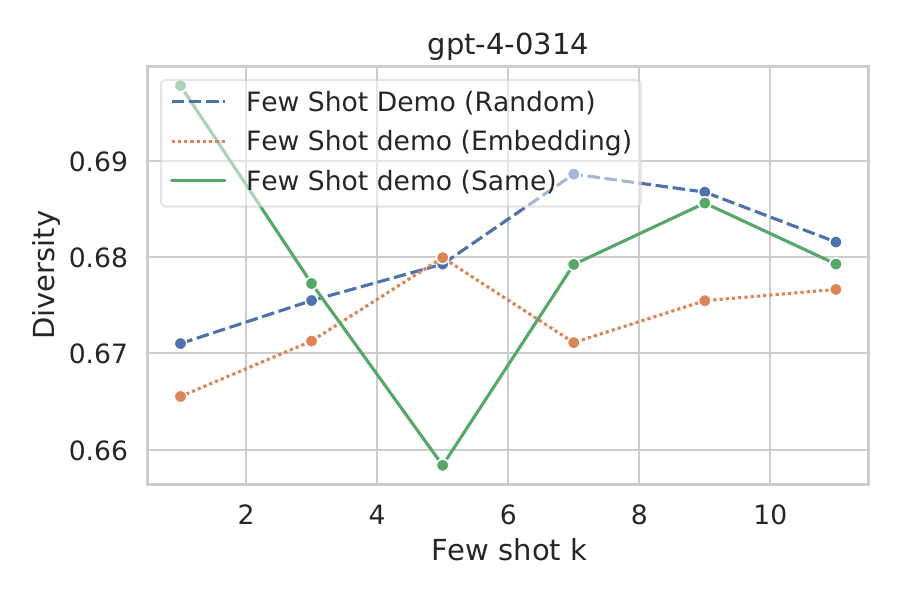}
    \includegraphics[width=0.325\textwidth]{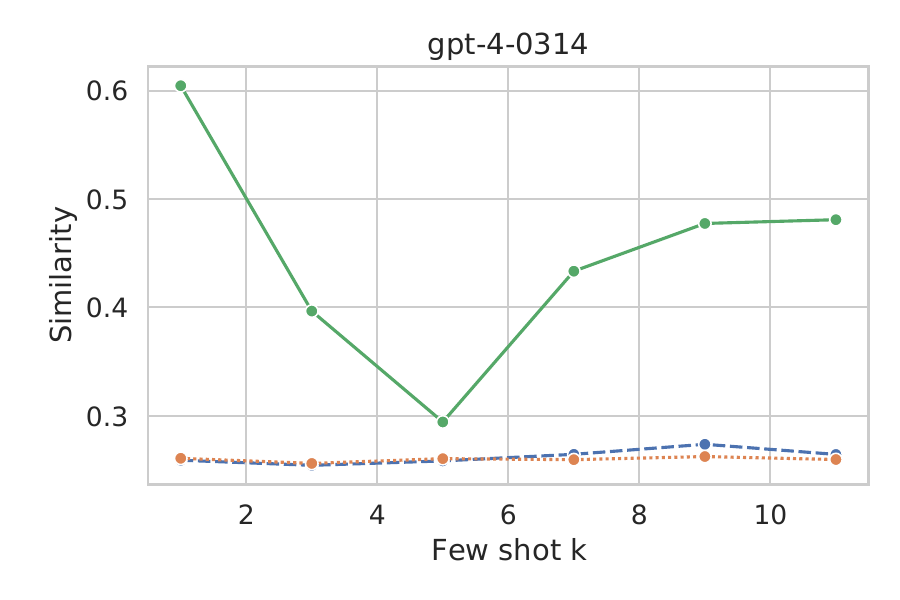}
    \includegraphics[width=0.325\textwidth]{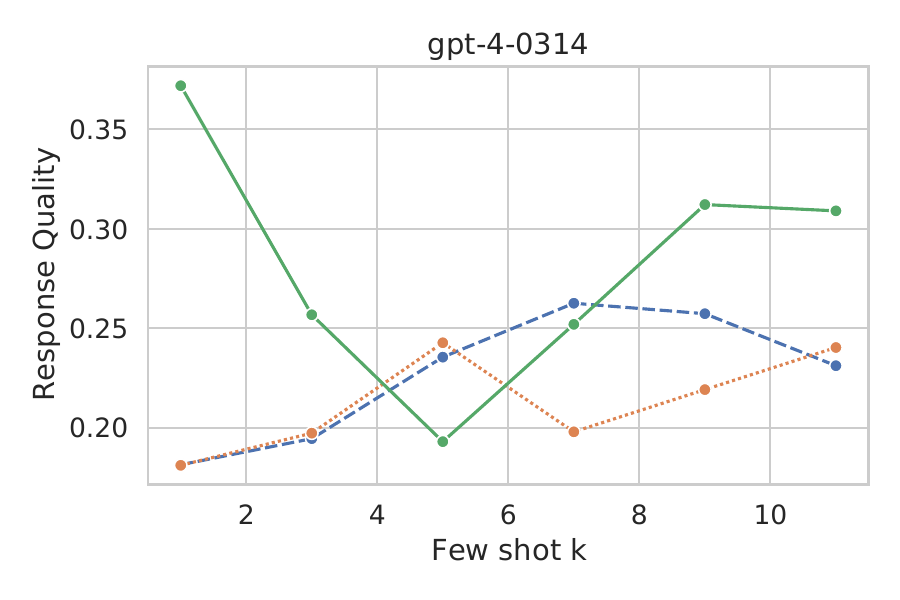}
    \includegraphics[width=0.325\textwidth]{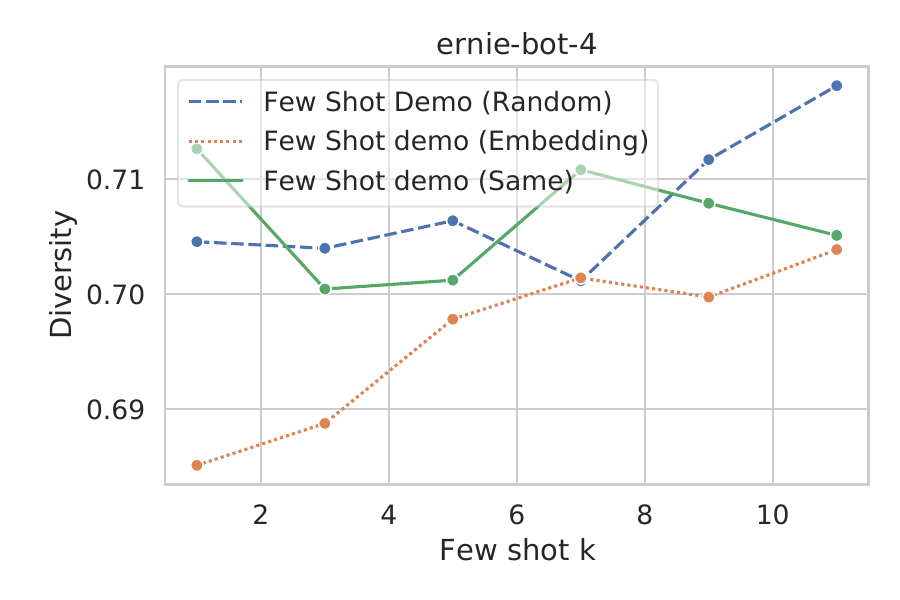}
    \includegraphics[width=0.325\textwidth]{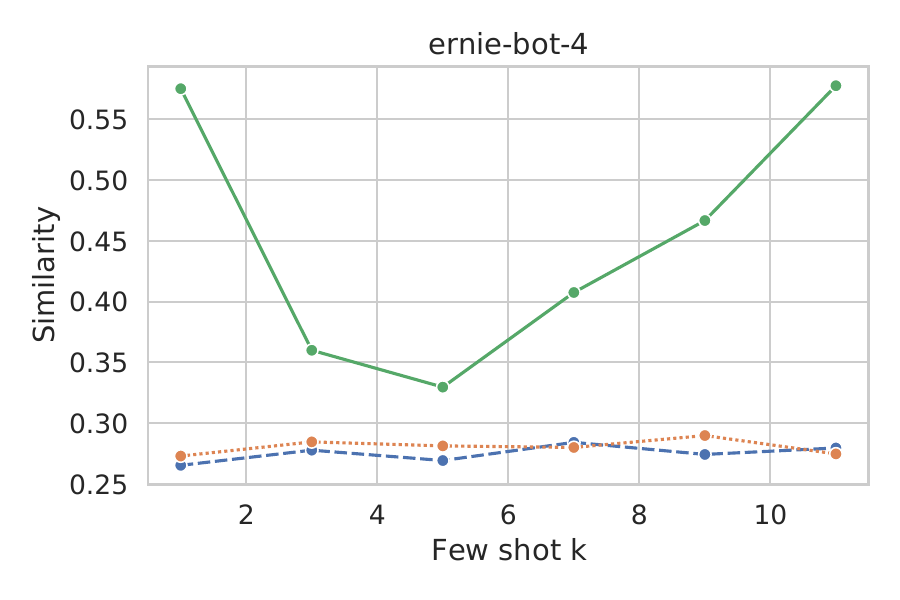}
    \includegraphics[width=0.325\textwidth]{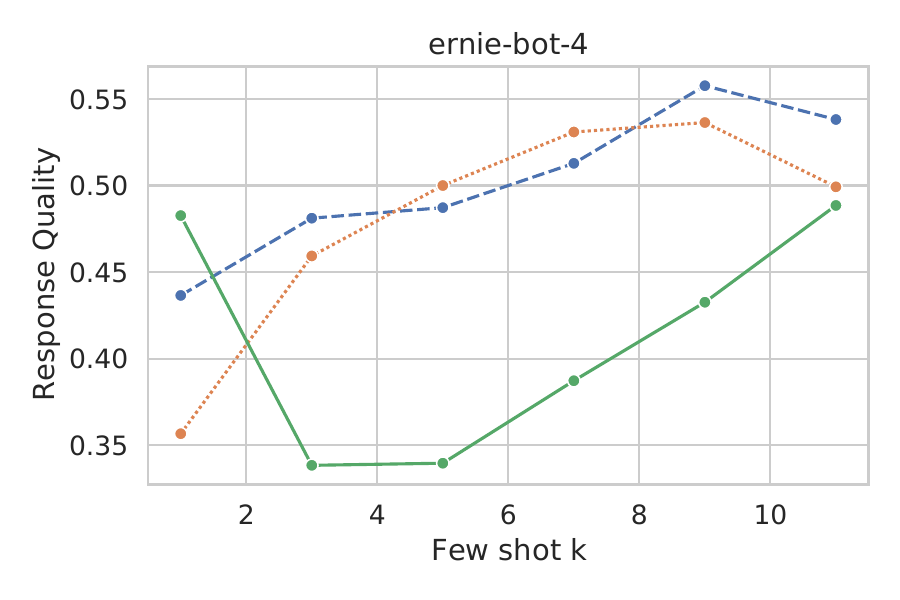}
    \caption{The performance comparison among three retrieval methods when the number of few-shot demonstrations $k$ varies. All values are averaged across different context lengths.}
    \label{fig:few_shot_demo_retrieval_method}
\end{figure*}

\noindent \textbf{Discussion of RQ2}: From Table~\ref{tab:result_main_creative_k5}, we can see that (comparing methods with No. 2, 3, and 4), the response quality of recalling demos of the same context is the worst, and this conclusion holds for all three language models. We hypothesize two reasons for this: First, the LLM may not be able to learn how to generalize from the (same) input-(different) output mapping from only 5 samples (as the one-to-many input-output format in the fine-tuning scenario also hinders model converging). \emph{Second, the quality of the LLM-generated responses is directly proportional to the effective information in the demos.} The effective information may include token distribution, single-turn semantics, multi-turn associations, etc. Clearly, demos with the same context have the least effective information as the same context $c_{:l-1}$ is repeated $5$ times, and the number of unique tokens is also the smallest (see Figure~\ref{fig:num_unique_token_for_diff_methods}). For the other two methods, as we increase the effective information by increasing the few-shot $k$, the overall dialogue quality shows an upward trend (see Figure~\ref{fig:few_shot_demo_retrieval_method}).

% Additionally, we find that demo context of previous turns $\{(c_{:0},r_1),\dots,(c_{:t-2}, r_{t-1})\}$ are helpful for LLMs (less pronounced gpt-4) to learn how to generate $\hat{r_t}$ based on $c_{:t-1}$. This is confirmed in heatmaps (\emph{w/o correct y} part) of Figures~\ref{fig:heatmap_gpt_35_corrupted_context,fig:heatmap_gpt4_corrupted_context,fig:heatmap_ernie_bot_corrupted_context}, where more dialogue history exists in the demo ($y$-axis), the higher the score of $\hat{r_t}$ in general.

\begin{figure*}
    \centering
    \includegraphics[width=0.325\textwidth]{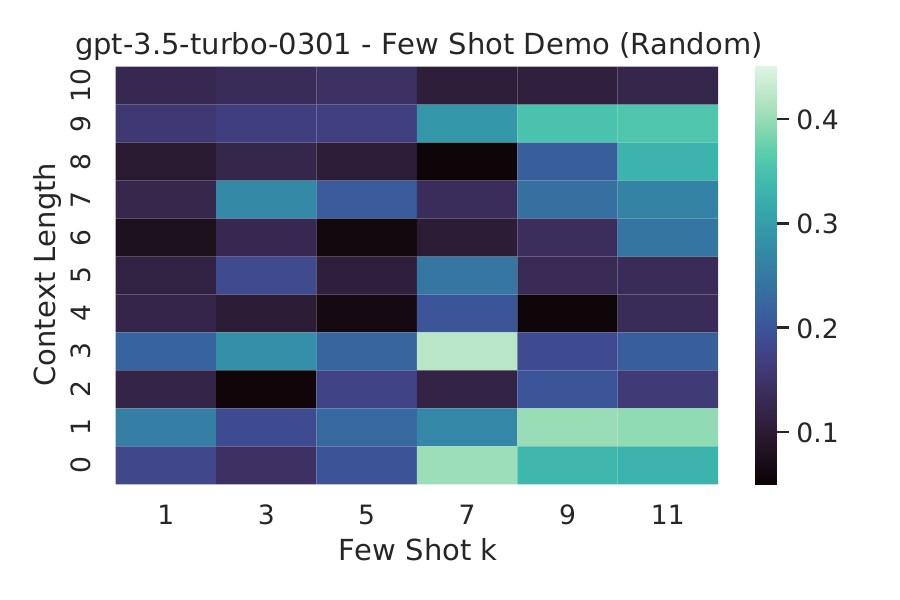}
    \includegraphics[width=0.325\textwidth]{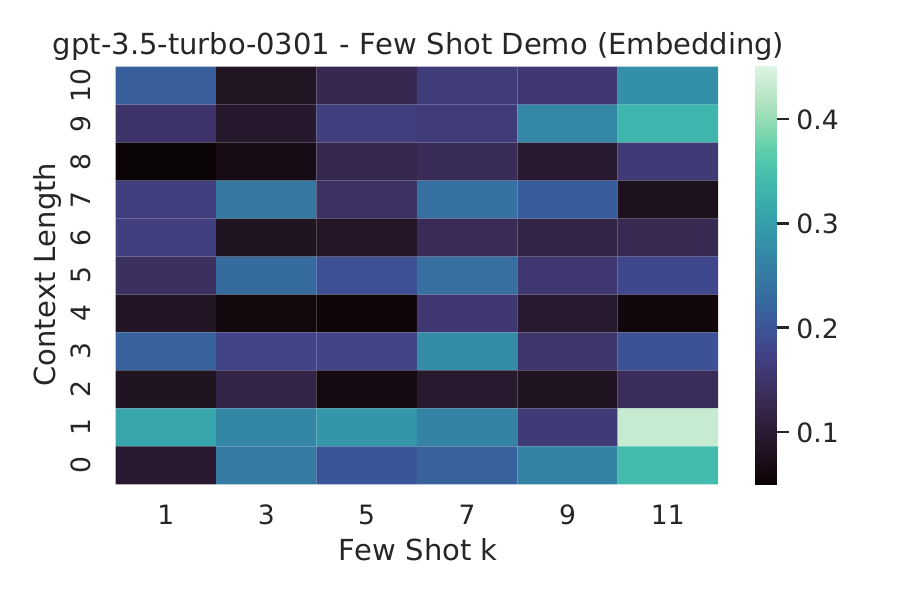}
    \includegraphics[width=0.325\textwidth]{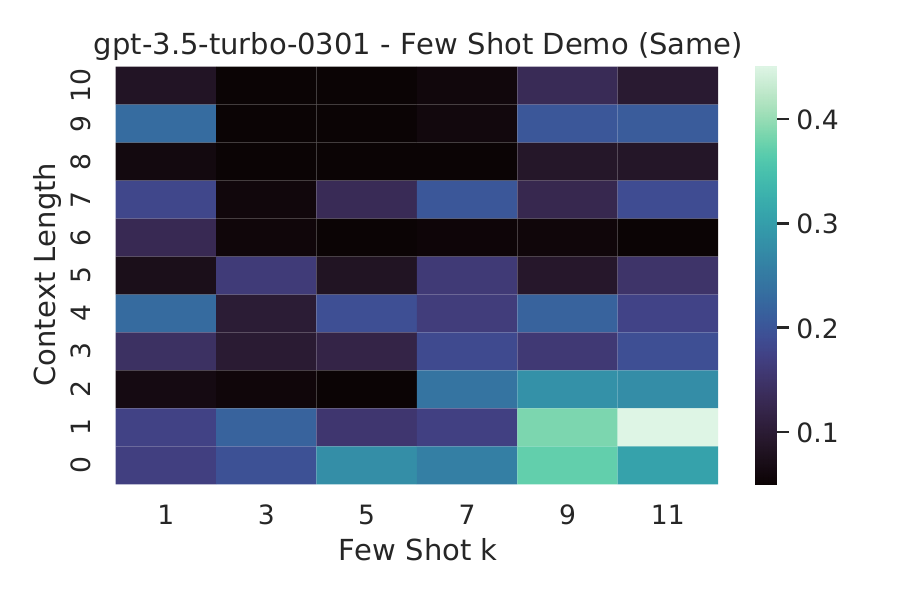}
    \includegraphics[width=0.325\textwidth]{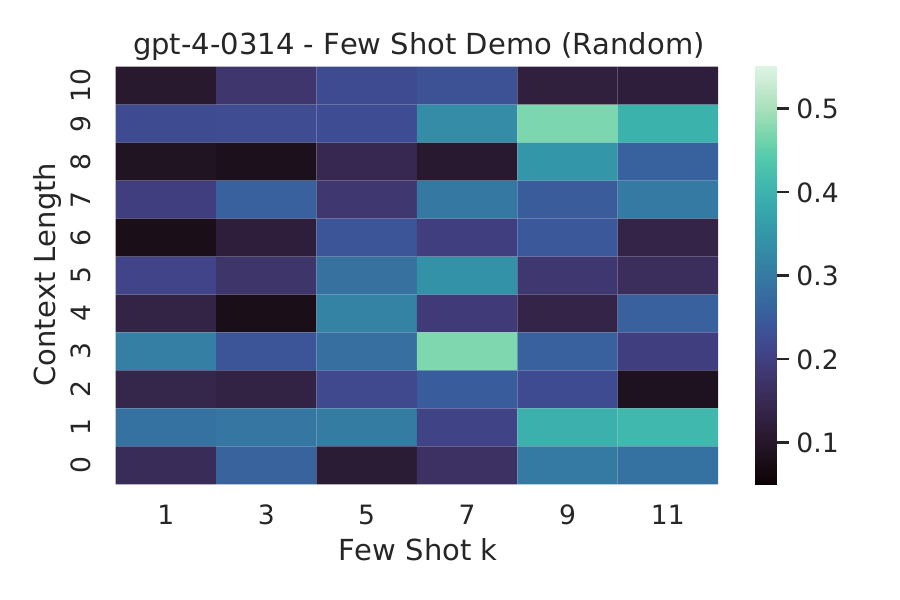}
    \includegraphics[width=0.325\textwidth]{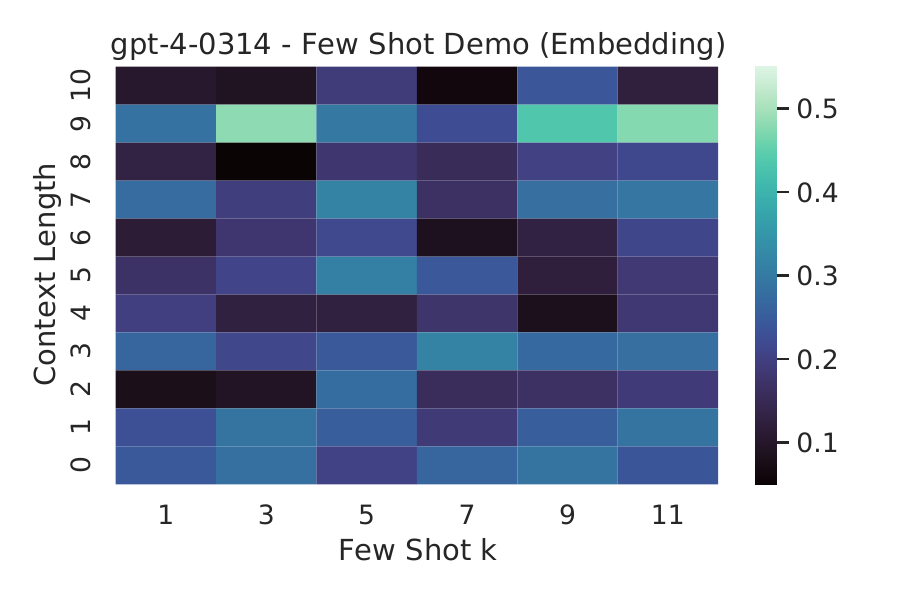}
    \includegraphics[width=0.325\textwidth]{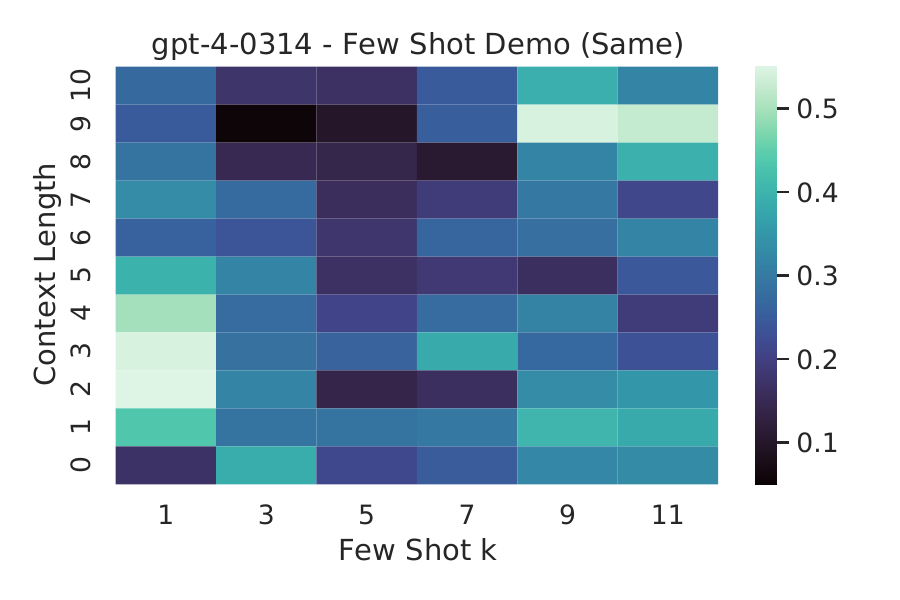}
    \includegraphics[width=0.325\textwidth]{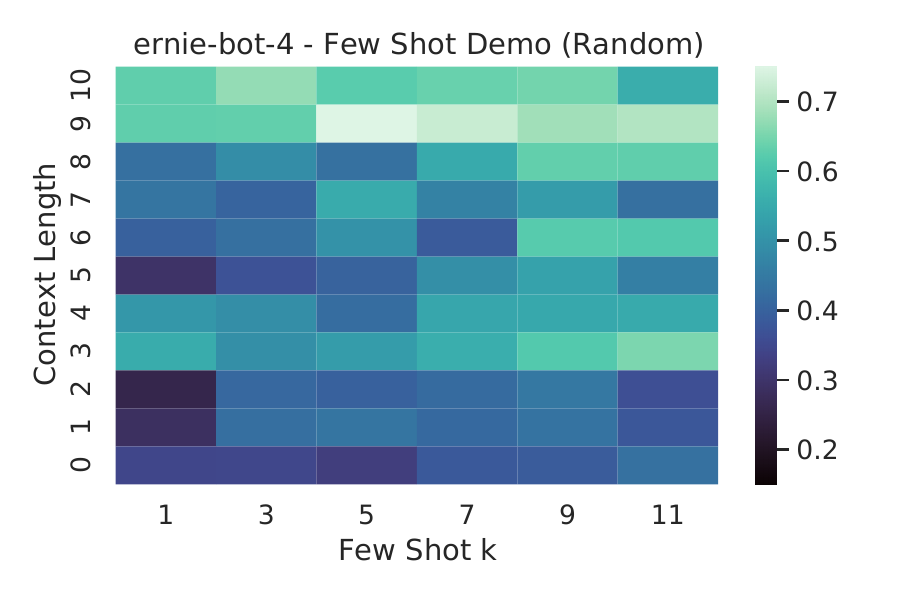}
    \includegraphics[width=0.325\textwidth]{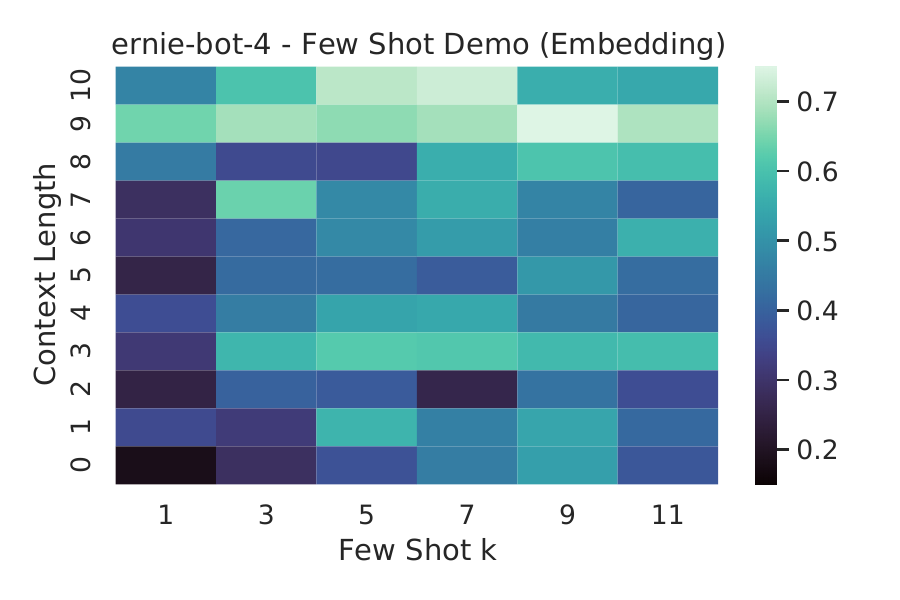}
    \includegraphics[width=0.325\textwidth]{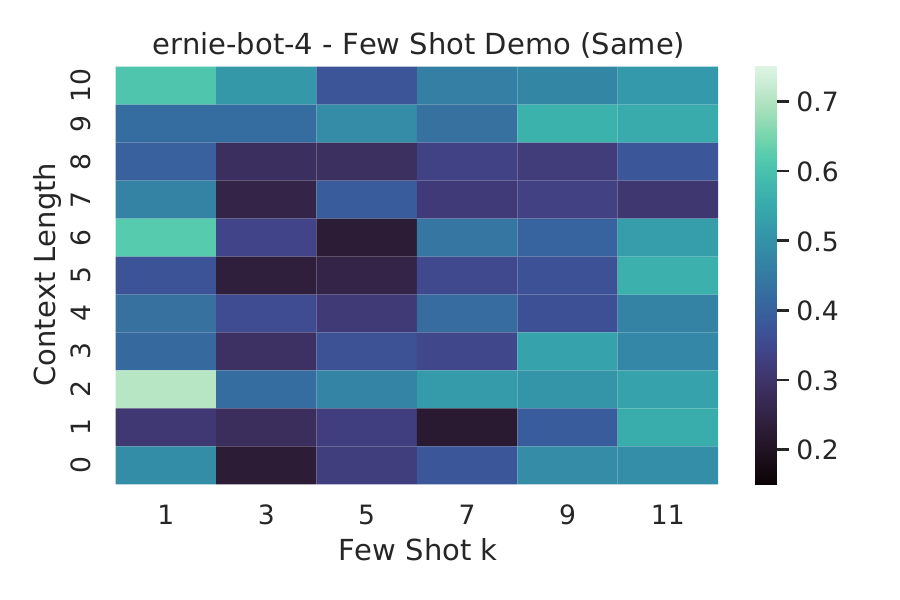}
    \caption{The performance (response quality) comparison among three retrieval methods when the number of few-shot demonstrations $k$ and context length both vary.}
    \label{fig:heatmap_different_retrieval_method}
\end{figure*}

The response quality of retrieving \textbf{randomly} and retrieving through \textbf{embedding} is relatively close, but the former has much more diverse responses. \emph{We believe that retrieving randomly is a strong baseline, and efforts should be focused on improving the quality of the demo set $\mathbf{x}_{demo}$, rather than the similarity between query context and demo context. When budget is sufficient, you may provide as many demos as possible because using larger $k$ can improve response quality (proved in Figure~\ref{fig:heatmap_different_retrieval_method}).}

\begin{figure*}
    \centering
    \includegraphics[width=0.325\textwidth]{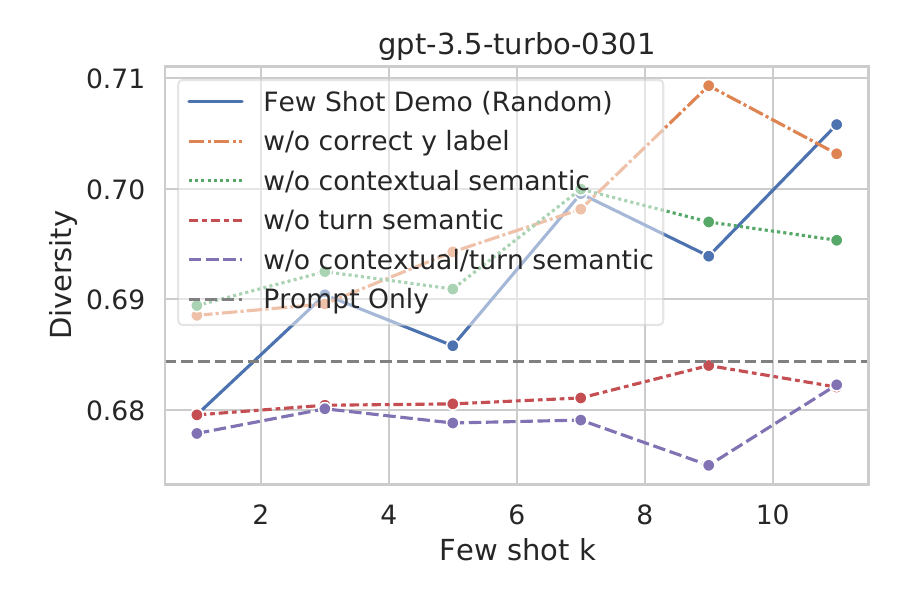}
    \includegraphics[width=0.325\textwidth]{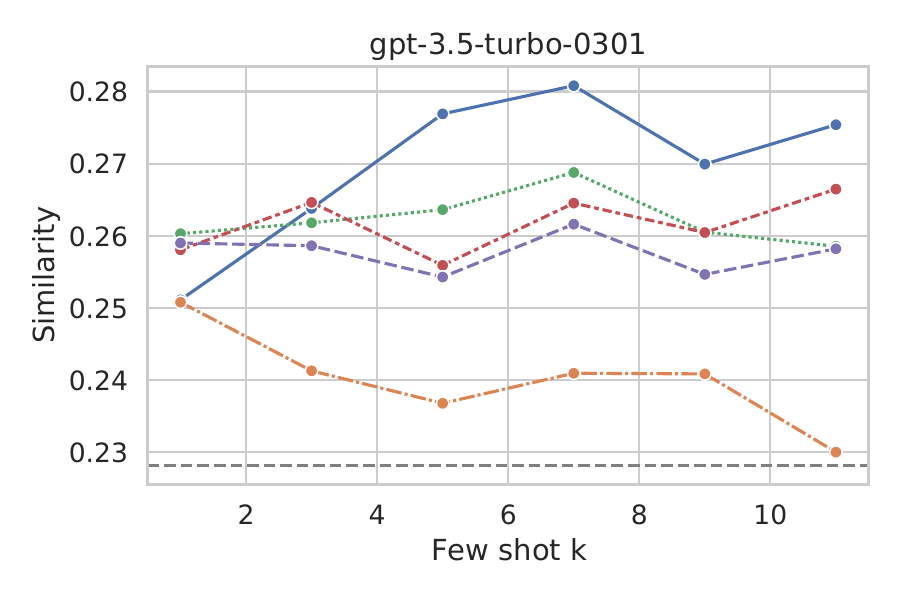}
    \includegraphics[width=0.325\textwidth]{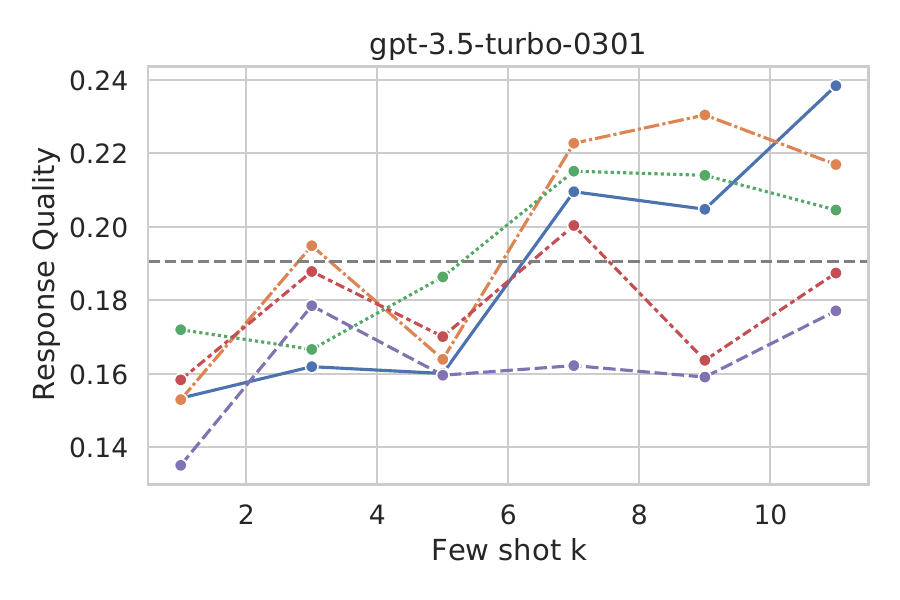}
    \includegraphics[width=0.325\textwidth]{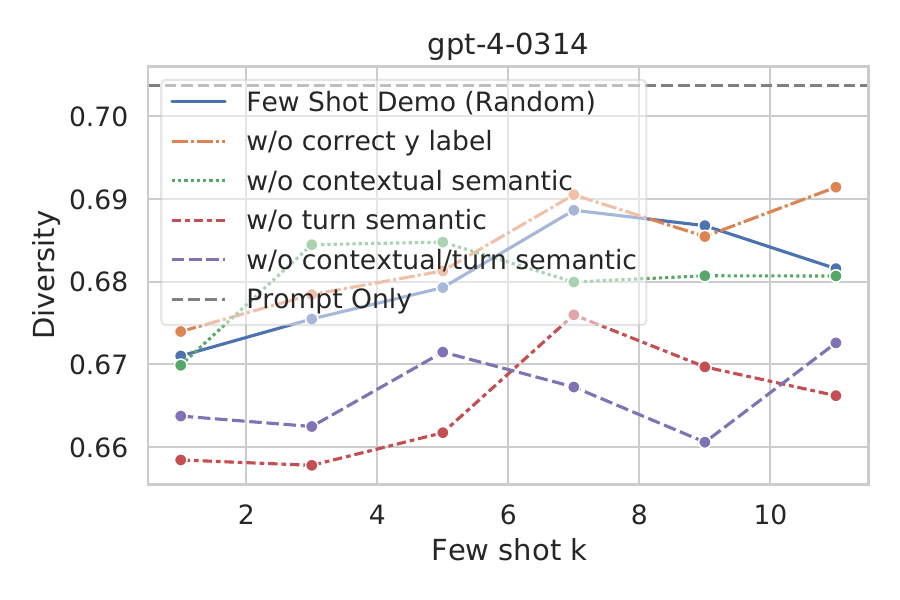}
    \includegraphics[width=0.325\textwidth]{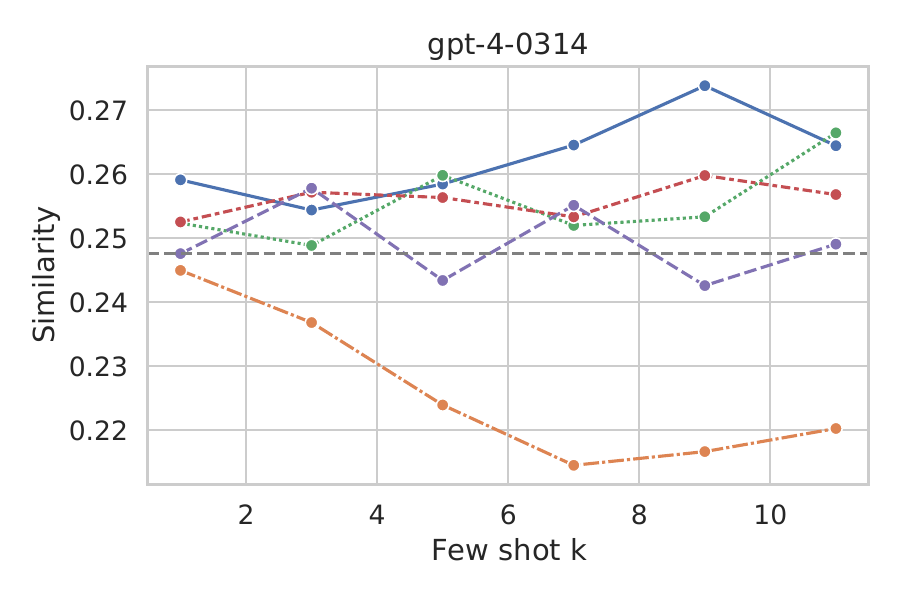}
    \includegraphics[width=0.325\textwidth]{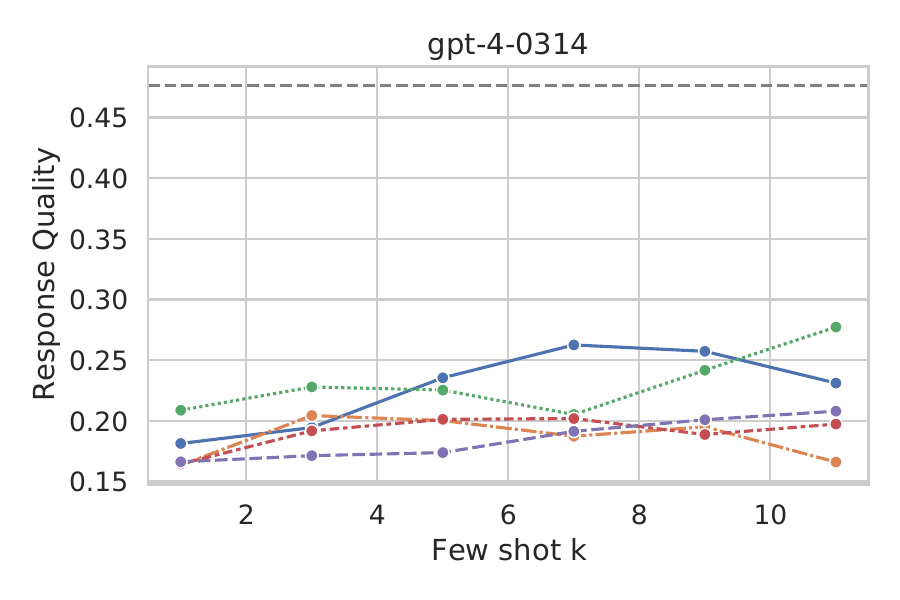}
    \includegraphics[width=0.325\textwidth]{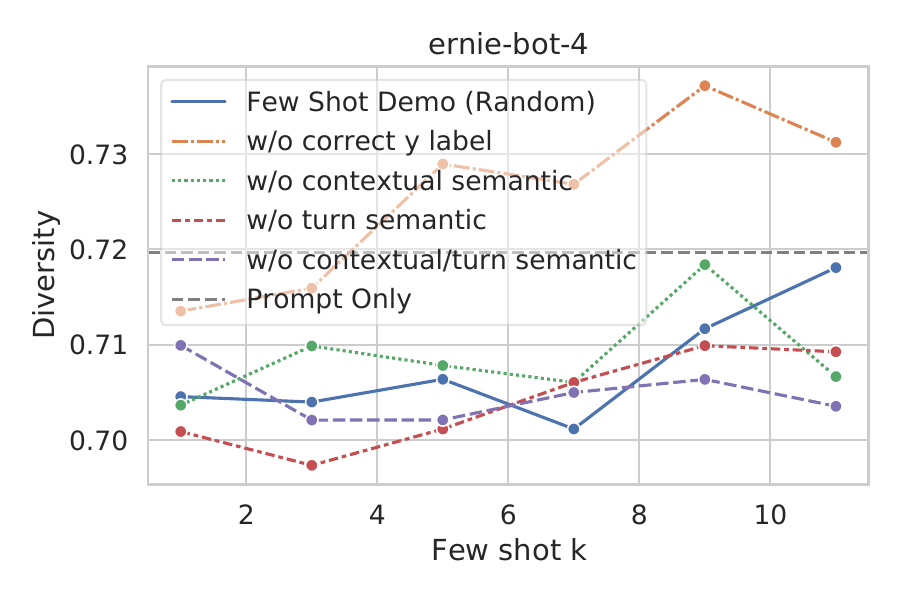}
    \includegraphics[width=0.325\textwidth]{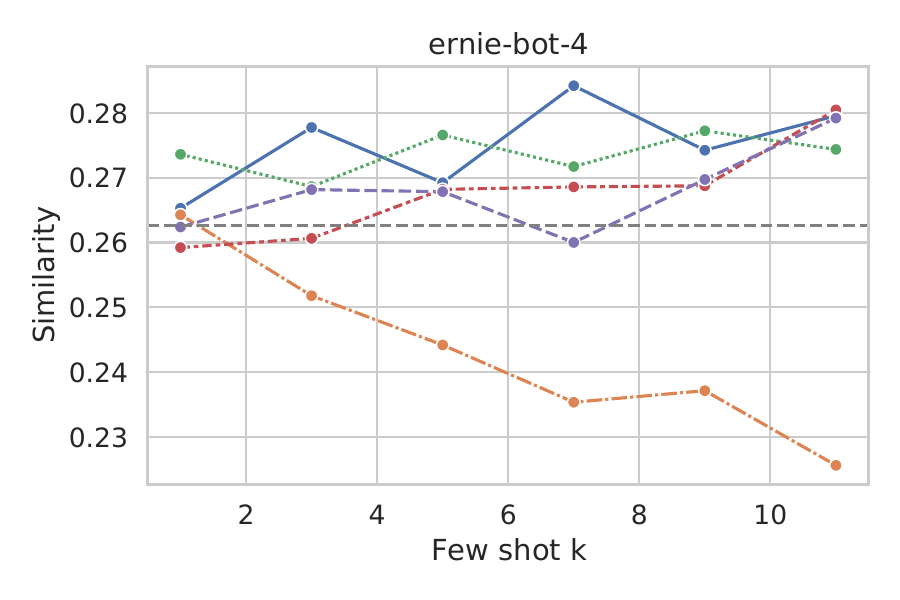}
    \includegraphics[width=0.325\textwidth]{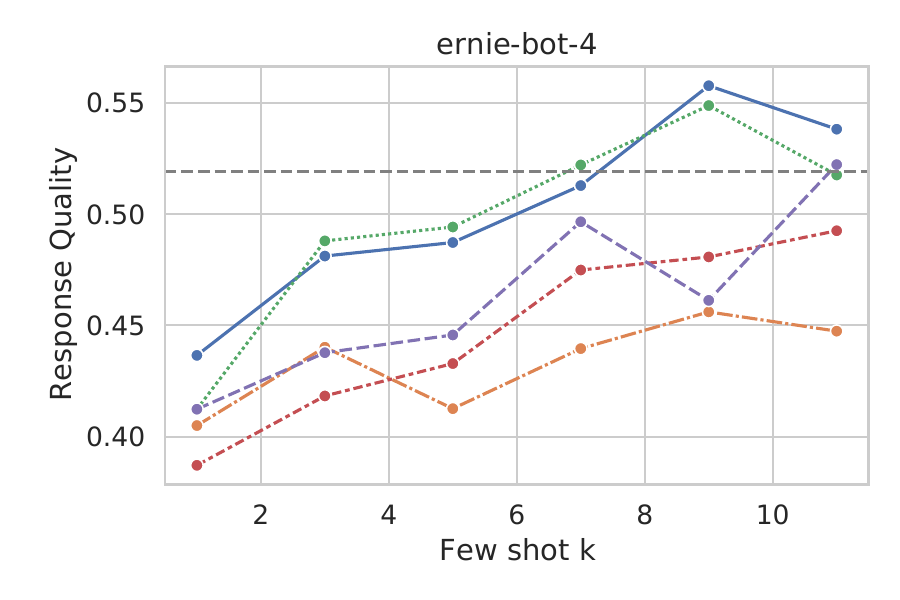}
    \caption{The impact of label substitution and different semantic corruption methods on diversity, similarity, and response quality when the number of few-shot demonstrations $k$ varies. All values are averaged cross different context lengths.}
    \label{fig:few_shot_corrupted_context}
\end{figure*}

\noindent \textbf{Discussion of RQ3}: Analyzing all LLMs, in both fixed ($k$=5, Table~\ref{tab:result_main_creative_k5}) and varying few-shot settings (Figure~\ref{fig:few_shot_corrupted_context}), shuffling the context of the demo does not affect the quality of generated responses (in some cases, it even improves the quality, as in Table~\ref{tab:result_main_creative_k5} for \emph{gpt-3.5}). Furthermore, when comparing \emph{w/o turn semantic} and \emph{w/o contextual/turn semantic}, we do not find the latter to significantly reduce the response quality; for \emph{ernie-bot-4}, we even find that the former has a greater negative impact (bottom right of Figure~\ref{fig:few_shot_corrupted_context}). Based on these results, we conjecture that \emph{LLMs do not learn generalizable dialogue generation abilities from multi-turn logical or semantic associations in $\mathbf{x}_{demo}$ during ICL}.

The corruption of single-turn semantics and grammar significantly reduces the generation quality of the three LLMs, which \emph{demonstrates that LLMs effectively learn single-turn semantics during the ICL process.} However, \emph{the ability to learn input-output mapping (w/o correct y label) varies among LLMs}, with \emph{gpt-3.5} performing poorly, while the other two models perform better.

\begin{figure}
    \centering
    \includegraphics[width=0.48\textwidth]{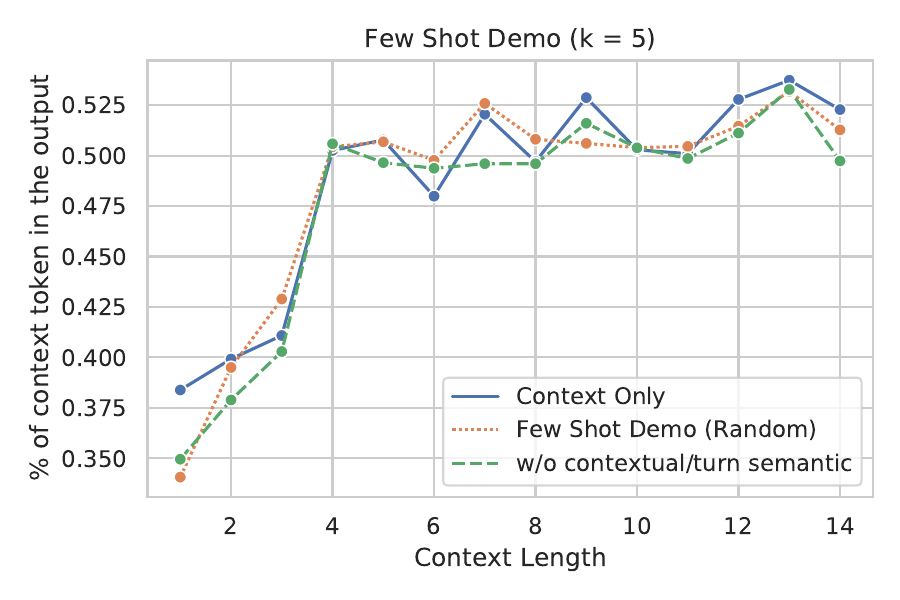}
    \caption{\textbf{X-axis}: length of the demonstration context. \textbf{Y-axis}: the proportion of LLM-generated tokens come from the token set of demonstrations $\mathbf{x}_{demo}$.}
    \label{fig:gpt_35_token_overlap}
\end{figure}

\begin{figure*}
    \centering
    \includegraphics[width=0.48\textwidth]{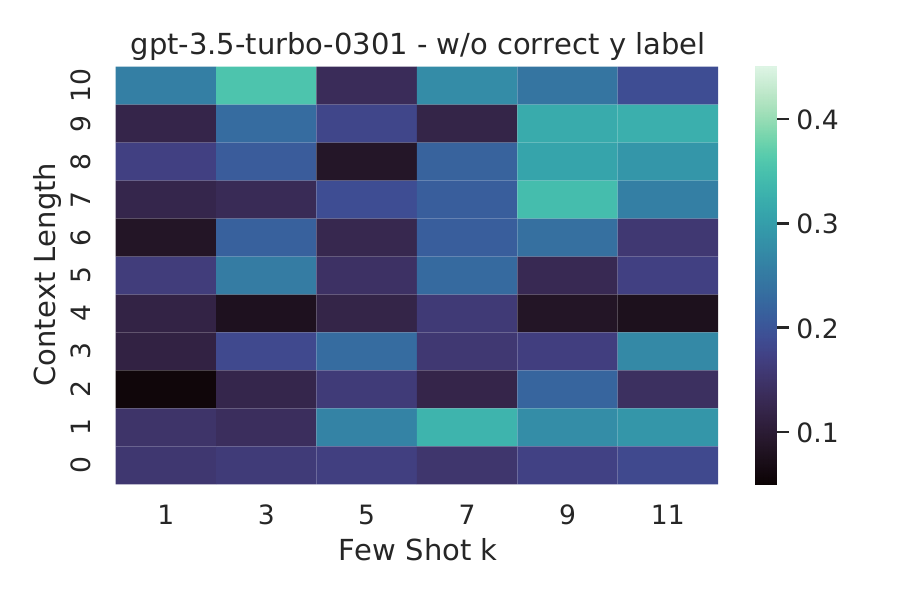}
    \includegraphics[width=0.48\textwidth]{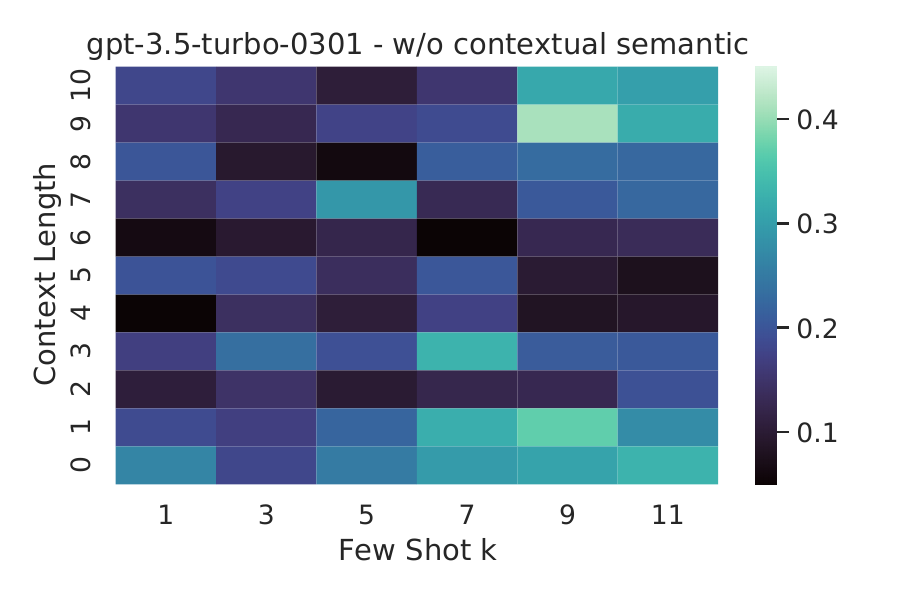}
    \includegraphics[width=0.48\textwidth]{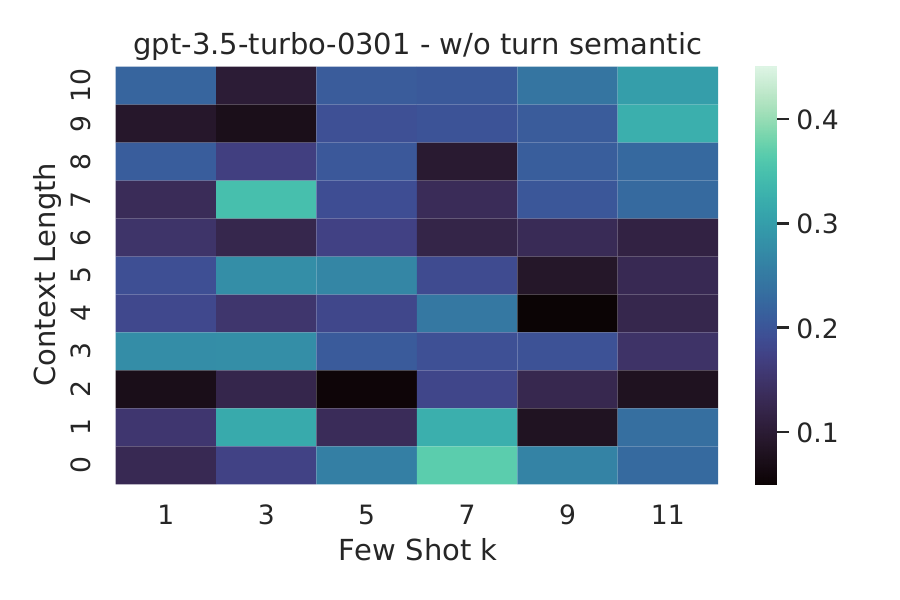}
    \includegraphics[width=0.48\textwidth]{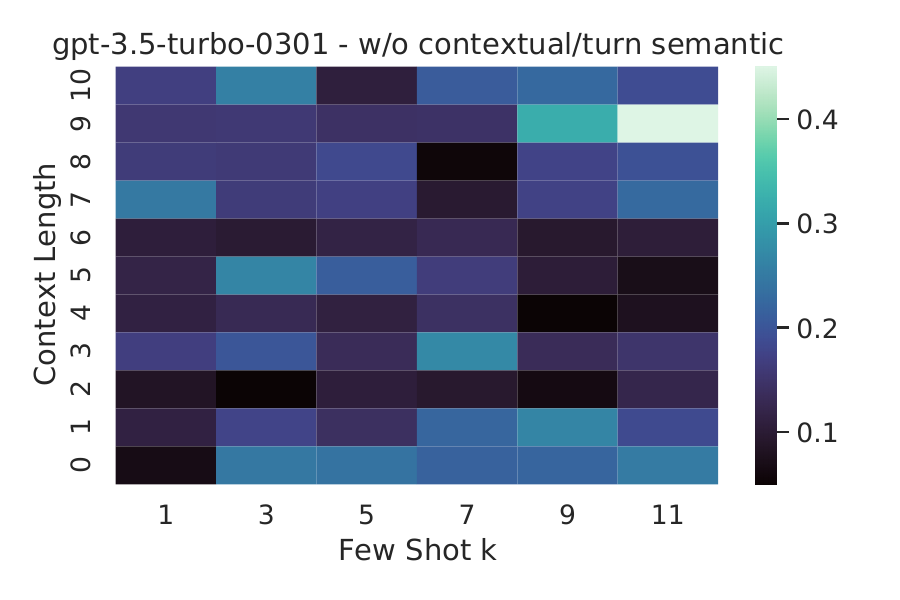}
    \caption{The impact of label substitution and different semantic corruption methods on response quality for \emph{gpt-3.5} when the number of few-shot demonstrations $k$ and context length both vary.}
    \label{fig:heatmap_gpt_35_corrupted_context}
\end{figure*}

\begin{figure*}
    \centering
    \includegraphics[width=0.48\textwidth]{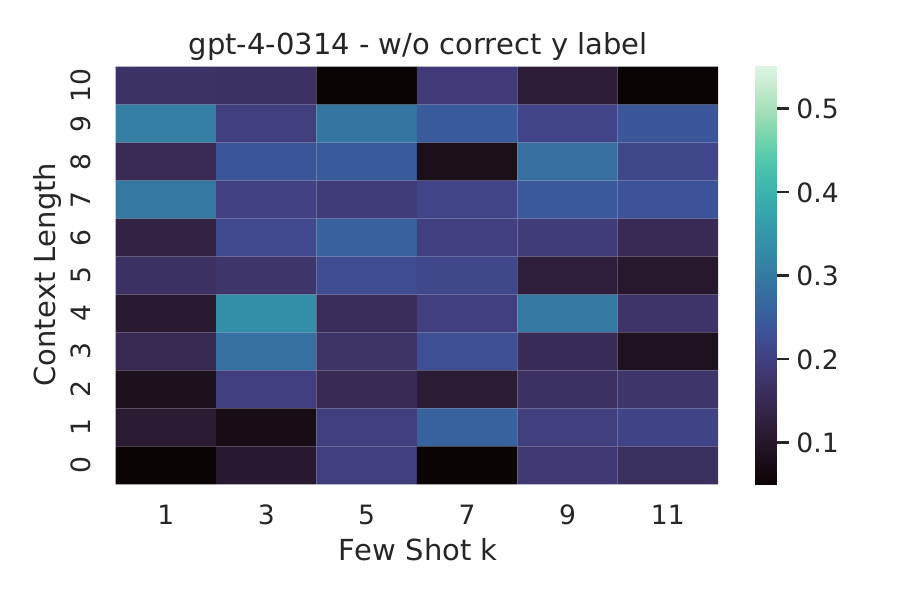}
    \includegraphics[width=0.48\textwidth]{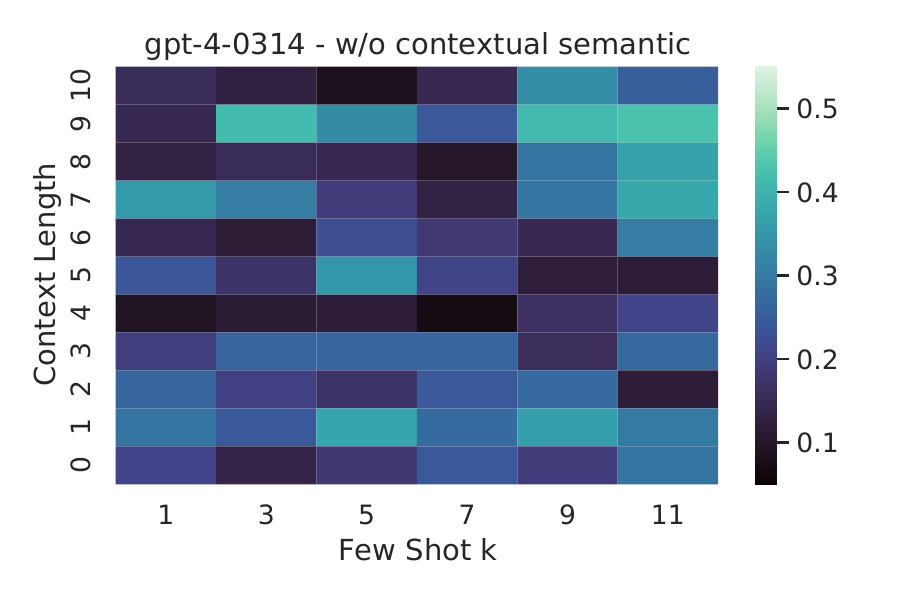}
    \includegraphics[width=0.48\textwidth]{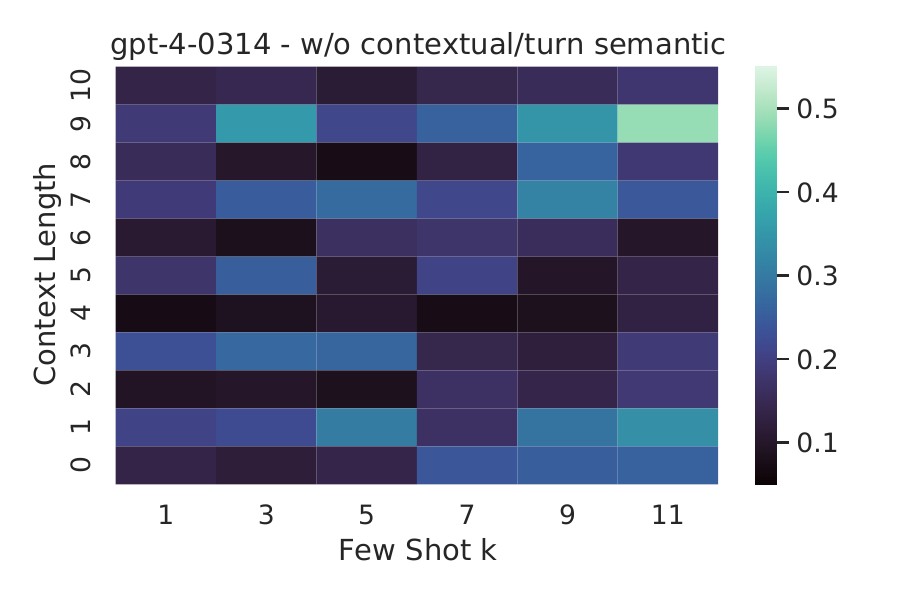}
    \includegraphics[width=0.48\textwidth]{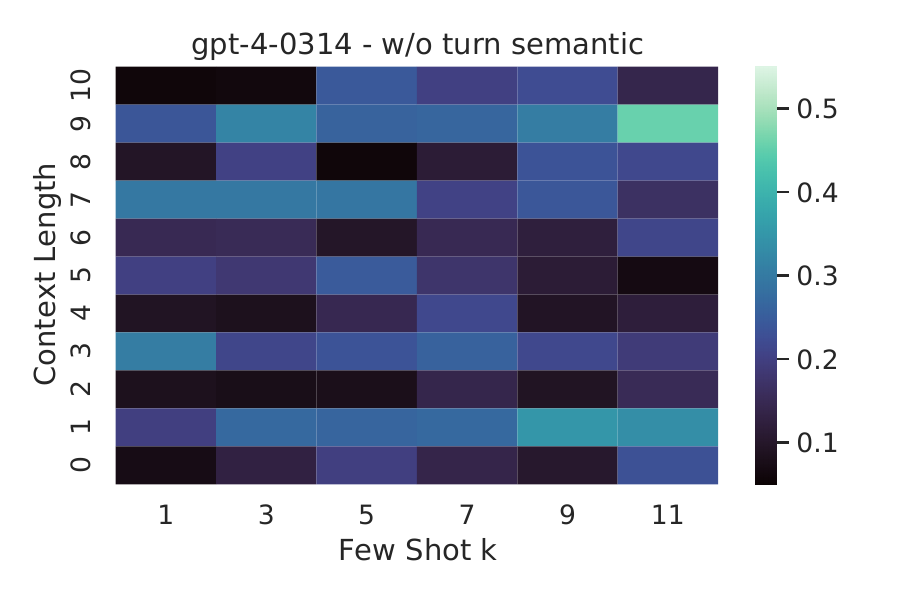}
    \caption{The impact of label substitution and different semantic corruption methods on response quality for \emph{gpt-4} when the number of few-shot demonstrations $k$ and context length both vary.}
    \label{fig:heatmap_gpt4_corrupted_context}
\end{figure*}

\begin{figure*}
    \centering
    \includegraphics[width=0.48\textwidth]{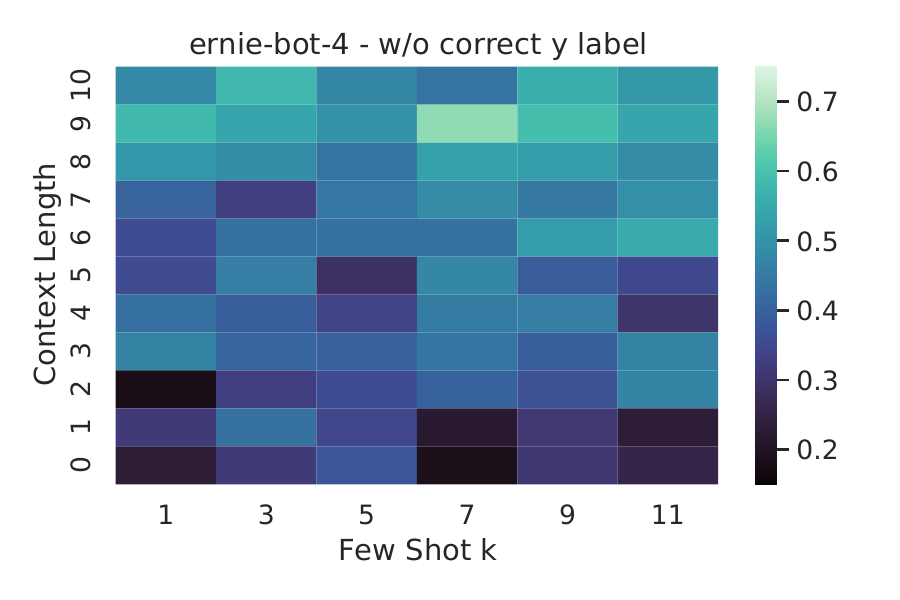}
    \includegraphics[width=0.48\textwidth]{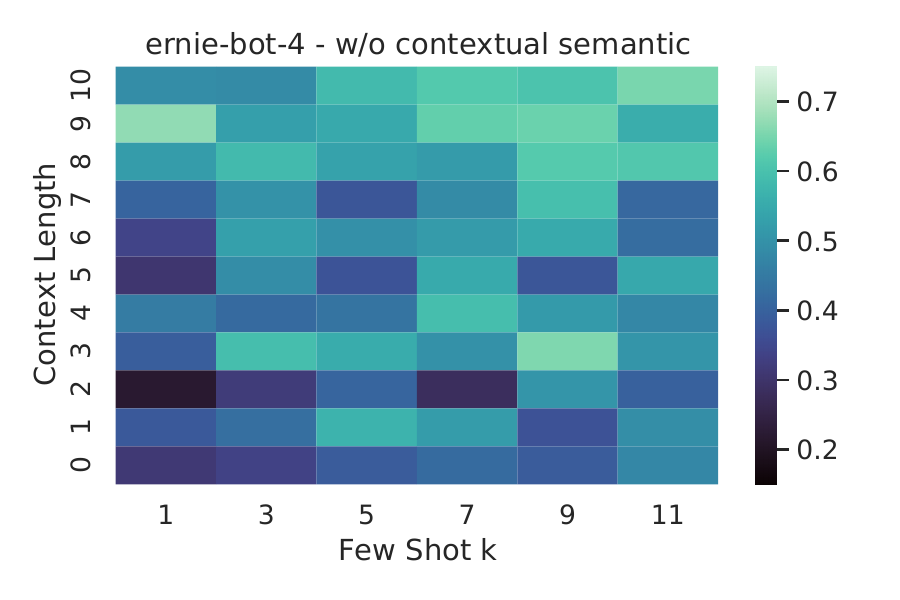}
    \includegraphics[width=0.48\textwidth]{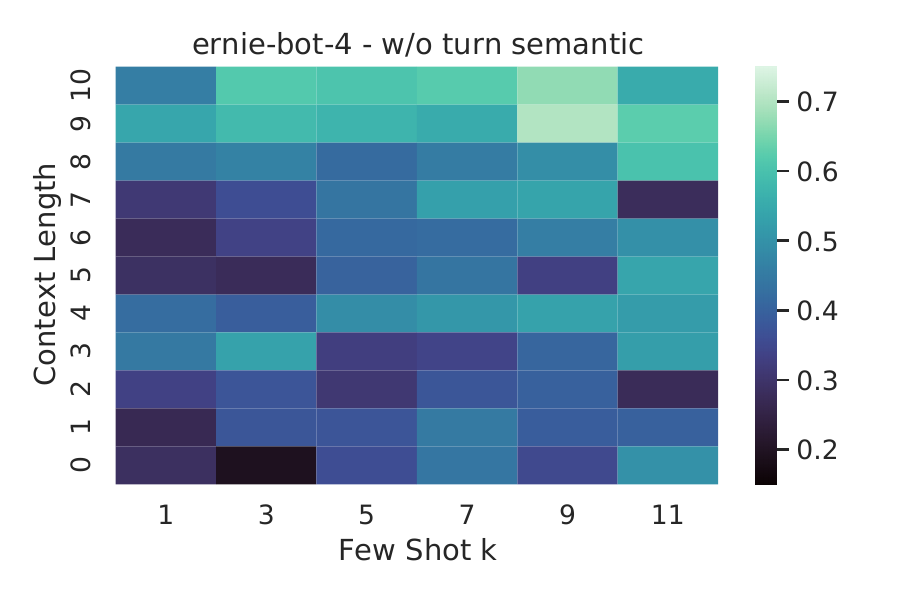}
    \includegraphics[width=0.48\textwidth]{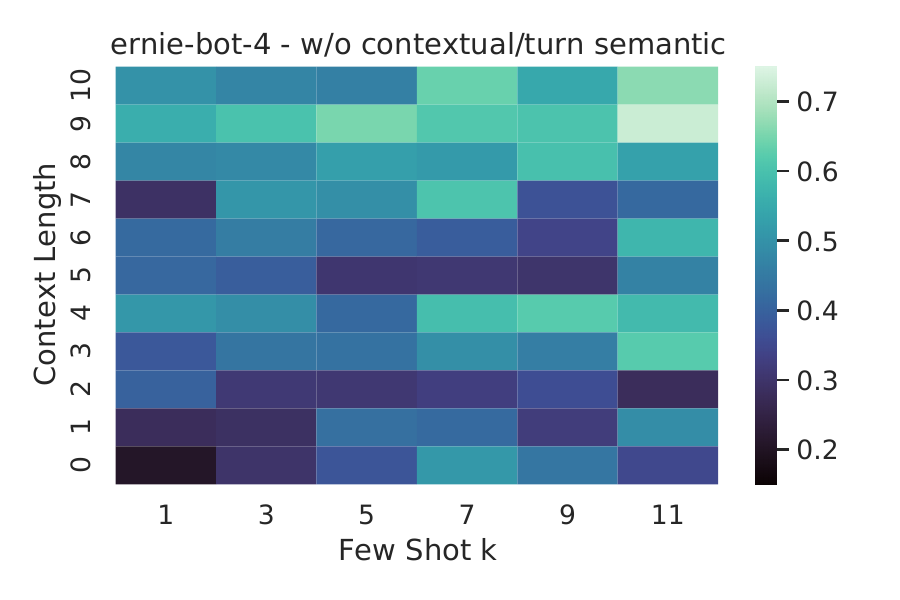}
    \caption{The impact of label substitution and different semantic corruption methods on response quality for \emph{ernie-bot-4} when the number of few-shot demonstrations $k$ and context length both vary.}
    \label{fig:heatmap_ernie_bot_corrupted_context}
\end{figure*}

What surprises us the most is that when we corrupt both single-turn and multi-turn semantics, the LLM-generated text quality remains significantly better than the \emph{Context Only} setting (Table~\ref{tab:result_main_creative_k5}). And as we increase the few-shot $k$ and the length of the context for these semantic-corruption methods (\Cref{fig:few_shot_corrupted_context,fig:heatmap_gpt_35_corrupted_context,fig:heatmap_gpt4_corrupted_context,fig:heatmap_ernie_bot_corrupted_context}), the response quality of LLMs still shows an upward trend (not particularly evident for GPT-4). Additionally, from Figure~\ref{fig:gpt_35_token_overlap}, we can see that LLM does not simply learn to copy tokens from demos; it generalizes. \emph{We conclude that LLM has the potential to conduct ICL well even when provided with corrupted $\mathbf{x}_{demo}$. Specifically, the LLM can learn from demos' mapping of token-distribution of $\widetilde{c_{:l-1}}$ and corrupted response $\widetilde{r_l}$ and improve the generation quality when a normal query context $c_{:t-1}$ is provided.}

Aside from the above three research questions, we also discover some other interesting phenomena. Please refer to \S\ref{sec:appendix_more_exp_results} for more details.

\section{Connections between our experimental conclusions and previous work}
Researchers ~\citet{reynolds2021prompt,sun2022paradigm,dong2022survey} find that in machine translation, classification, and other tasks, carefully crafted manual prompts can perform better than few-shot learning (consistent with \textbf{our} conclusion). Additionally, ~\citet{reynolds2021prompt} discovers that Zero-Shot (corresponding to our \emph{Context Only} method) performance is also better than Few-Shot, which contradicts \textbf{our} findings.

Regarding the retrieval of demos, previous work has concluded that similarity-based retrieval is significantly better than random retrieval in tasks such as sentiment classification, table-to-text, and semantic parsing~\citep{liu2022makes,rubin2022learning}. However, \textbf{our} findings suggest that the random baseline performs the best, possibly due to its superior diversity, which can lead to compositional generalization~\citep{levy-etal-2023-diverse}.

Regarding whether LLMs can effectively learn y label during ICL, some work argues that input-output mapping is not important~\citep{min-etal-2022-rethinking}, while others affirm the value of y label~\citep{li-qiu-2023-finding,kossen2023context}. \textbf{Our} conclusion is that the y label is helpful for dialogue generation, but to a relatively small extent.

Regarding the relationship between the number of demos and the performance ICL, \citet{reynolds2021prompt} believes that there is no linear relationship between them, while \citet{li-etal-2023-unified,wu-etal-2023-self,hao2022structured} find that in story generation and classification tasks, the more demos, the better the performance (\textbf{our} experimental results support this claim, too).

\section{Limitation}
Due to limited resources, we have not attempted many more complex methods for selecting demos, and most of these methods have not been tested for their effectiveness in dialogue generation~\citep{iter-etal-2023-context,ye2023compositional}. Additionally, we hope that future researchers can further expand the test set or explore whether similar conclusions can be drawn from ICL in other languages.

Our paper investigates the performance of ICL in persona-based dialogue generation and provides some practical suggestions, but does not delve into the underlying mechanisms of ICL in this context. Existing hypotheses, such as those based on $n$-gram (copy/induction) head, cannot fully account for our experimental results. We believe that the principles behind the effectiveness of ICL warrant further investigation.
% \input{Acknowledgements}

\bibliography{anthology,custom}

\begin{thebibliography}{41}
\expandafter\ifx\csname natexlab\endcsname\relax\def\natexlab#1{#1}\fi

\bibitem[{Achiam et~al.(2023)Achiam, Adler, Agarwal, Ahmad, Akkaya, Aleman, Almeida, Altenschmidt, Altman, Anadkat et~al.}]{achiam2023gpt}
Josh Achiam, Steven Adler, Sandhini Agarwal, Lama Ahmad, Ilge Akkaya, Florencia~Leoni Aleman, Diogo Almeida, Janko Altenschmidt, Sam Altman, Shyamal Anadkat, et~al. 2023.
\newblock Gpt-4 technical report.
\newblock \emph{arXiv preprint arXiv:2303.08774}.

\bibitem[{Akyürek et~al.(2024)Akyürek, Wang, Kim, and Andreas}]{akyürek2024incontext}
Ekin Akyürek, Bailin Wang, Yoon Kim, and Jacob Andreas. 2024.
\newblock \href {http://arxiv.org/abs/2401.12973} {In-context language learning: Architectures and algorithms}.

\bibitem[{Cao et~al.(2022)Cao, Bi, Fang, Shi, and Tao}]{cao2022model}
Yu~Cao, Wei Bi, Meng Fang, Shuming Shi, and Dacheng Tao. 2022.
\newblock A model-agnostic data manipulation method for persona-based dialogue generation.
\newblock In \emph{Proceedings of the 60th Annual Meeting of the Association for Computational Linguistics (Volume 1: Long Papers)}, pages 7984--8002.

\bibitem[{De~Gennaro et~al.(2020)De~Gennaro, Krumhuber, and Lucas}]{de2020effectiveness}
Mauro De~Gennaro, Eva~G Krumhuber, and Gale Lucas. 2020.
\newblock Effectiveness of an empathic chatbot in combating adverse effects of social exclusion on mood.
\newblock \emph{Frontiers in psychology}, 10:3061.

\bibitem[{Dong et~al.(2022)Dong, Li, Dai, Zheng, Wu, Chang, Sun, Xu, and Sui}]{dong2022survey}
Qingxiu Dong, Lei Li, Damai Dai, Ce~Zheng, Zhiyong Wu, Baobao Chang, Xu~Sun, Jingjing Xu, and Zhifang Sui. 2022.
\newblock A survey for in-context learning.
\newblock \emph{arXiv preprint arXiv:2301.00234}.

\bibitem[{Du et~al.(2022)Du, Qian, Liu, Ding, Qiu, Yang, and Tang}]{du2022glm}
Zhengxiao Du, Yujie Qian, Xiao Liu, Ming Ding, Jiezhong Qiu, Zhilin Yang, and Jie Tang. 2022.
\newblock Glm: General language model pretraining with autoregressive blank infilling.
\newblock In \emph{Proceedings of the 60th Annual Meeting of the Association for Computational Linguistics (Volume 1: Long Papers)}, pages 320--335.

\bibitem[{Fu et~al.(2023)Fu, Ng, Jiang, and Liu}]{fu2023gptscore}
Jinlan Fu, See-Kiong Ng, Zhengbao Jiang, and Pengfei Liu. 2023.
\newblock Gptscore: Evaluate as you desire.
\newblock \emph{arXiv preprint arXiv:2302.04166}.

\bibitem[{Gao et~al.(2021)Gao, Yao, and Chen}]{gao2021simcse}
Tianyu Gao, Xingcheng Yao, and Danqi Chen. 2021.
\newblock Simcse: Simple contrastive learning of sentence embeddings.
\newblock In \emph{Proceedings of the 2021 Conference on Empirical Methods in Natural Language Processing}, pages 6894--6910.

\bibitem[{Hao et~al.(2022)Hao, Sun, Dong, Han, Gu, and Wei}]{hao2022structured}
Yaru Hao, Yutao Sun, Li~Dong, Zhixiong Han, Yuxian Gu, and Furu Wei. 2022.
\newblock Structured prompting: Scaling in-context learning to 1,000 examples.
\newblock \emph{arXiv preprint arXiv:2212.06713}.

\bibitem[{Holtzman et~al.(2020)Holtzman, Buys, Du, Forbes, and Choi}]{Holtzman2020The}
Ari Holtzman, Jan Buys, Li~Du, Maxwell Forbes, and Yejin Choi. 2020.
\newblock \href {https://openreview.net/forum?id=rygGQyrFvH} {The curious case of neural text degeneration}.
\newblock In \emph{International Conference on Learning Representations}.

\bibitem[{Huang et~al.(2023)Huang, Zhang, Ko, Liu, Wu, Wang, and Tang}]{huang2023personalized}
Qiushi Huang, Yu~Zhang, Tom Ko, Xubo Liu, Bo~Wu, Wenwu Wang, and H~Tang. 2023.
\newblock Personalized dialogue generation with persona-adaptive attention.
\newblock In \emph{Proceedings of the AAAI Conference on Artificial Intelligence}, volume~37, pages 12916--12923.

\bibitem[{Iter et~al.(2023)Iter, Pryzant, Xu, Wang, Liu, Xu, and Zhu}]{iter-etal-2023-context}
Dan Iter, Reid Pryzant, Ruochen Xu, Shuohang Wang, Yang Liu, Yichong Xu, and Chenguang Zhu. 2023.
\newblock \href {https://doi.org/10.18653/v1/2023.findings-emnlp.81} {In-context demonstration selection with cross entropy difference}.
\newblock In \emph{Findings of the Association for Computational Linguistics: EMNLP 2023}, pages 1150--1162, Singapore. Association for Computational Linguistics.

\bibitem[{Kossen et~al.(2023)Kossen, Rainforth, and Gal}]{kossen2023context}
Jannik Kossen, Tom Rainforth, and Yarin Gal. 2023.
\newblock In-context learning in large language models learns label relationships but is not conventional learning.
\newblock \emph{arXiv preprint arXiv:2307.12375}.

\bibitem[{Levy et~al.(2023)Levy, Bogin, and Berant}]{levy-etal-2023-diverse}
Itay Levy, Ben Bogin, and Jonathan Berant. 2023.
\newblock \href {https://doi.org/10.18653/v1/2023.acl-long.78} {Diverse demonstrations improve in-context compositional generalization}.
\newblock In \emph{Proceedings of the 61st Annual Meeting of the Association for Computational Linguistics (Volume 1: Long Papers)}, pages 1401--1422, Toronto, Canada. Association for Computational Linguistics.

\bibitem[{Li et~al.(2015)Li, Galley, Brockett, Gao, and Dolan}]{distinct}
Jiwei Li, Michel Galley, Chris Brockett, Jianfeng Gao, and Bill Dolan. 2015.
\newblock \href {http://arxiv.org/abs/1510.03055} {A diversity-promoting objective function for neural conversation models}.
\newblock \emph{CoRR}, abs/1510.03055.

\bibitem[{Li et~al.(2023)Li, Lv, Yan, Lin, Zhu, Ni, Xie, Wang, and Qiu}]{li-etal-2023-unified}
Xiaonan Li, Kai Lv, Hang Yan, Tianyang Lin, Wei Zhu, Yuan Ni, Guotong Xie, Xiaoling Wang, and Xipeng Qiu. 2023.
\newblock \href {https://doi.org/10.18653/v1/2023.acl-long.256} {Unified demonstration retriever for in-context learning}.
\newblock In \emph{Proceedings of the 61st Annual Meeting of the Association for Computational Linguistics (Volume 1: Long Papers)}, pages 4644--4668, Toronto, Canada. Association for Computational Linguistics.

\bibitem[{Li and Qiu(2023)}]{li-qiu-2023-finding}
Xiaonan Li and Xipeng Qiu. 2023.
\newblock \href {https://doi.org/10.18653/v1/2023.findings-emnlp.411} {Finding support examples for in-context learning}.
\newblock In \emph{Findings of the Association for Computational Linguistics: EMNLP 2023}, pages 6219--6235, Singapore. Association for Computational Linguistics.

\bibitem[{Lin(2004)}]{lin-2004-rouge}
Chin-Yew Lin. 2004.
\newblock \href {https://aclanthology.org/W04-1013} {{ROUGE}: A package for automatic evaluation of summaries}.
\newblock In \emph{Text Summarization Branches Out}, pages 74--81, Barcelona, Spain. Association for Computational Linguistics.

\bibitem[{Liu et~al.(2022{\natexlab{a}})Liu, Shen, Zhang, Dolan, Carin, and Chen}]{liu2022makes}
Jiachang Liu, Dinghan Shen, Yizhe Zhang, William~B Dolan, Lawrence Carin, and Weizhu Chen. 2022{\natexlab{a}}.
\newblock What makes good in-context examples for gpt-3?
\newblock In \emph{Proceedings of Deep Learning Inside Out (DeeLIO 2022): The 3rd Workshop on Knowledge Extraction and Integration for Deep Learning Architectures}, pages 100--114.

\bibitem[{Liu et~al.(2022{\natexlab{b}})Liu, Symons, and Vatsavai}]{liu2022persona}
Junfeng Liu, Christopher Symons, and Ranga~Raju Vatsavai. 2022{\natexlab{b}}.
\newblock Persona-based conversational ai: State of the art and challenges.
\newblock In \emph{2022 IEEE International Conference on Data Mining Workshops (ICDMW)}, pages 993--1001. IEEE.

\bibitem[{Liu et~al.(2019)Liu, Ott, Goyal, Du, Joshi, Chen, Levy, Lewis, Zettlemoyer, and Stoyanov}]{liu2019roberta}
Yinhan Liu, Myle Ott, Naman Goyal, Jingfei Du, Mandar Joshi, Danqi Chen, Omer Levy, Mike Lewis, Luke Zettlemoyer, and Veselin Stoyanov. 2019.
\newblock Roberta: A robustly optimized bert pretraining approach.
\newblock \emph{arXiv preprint arXiv:1907.11692}.

\bibitem[{Min et~al.(2022)Min, Lyu, Holtzman, Artetxe, Lewis, Hajishirzi, and Zettlemoyer}]{min-etal-2022-rethinking}
Sewon Min, Xinxi Lyu, Ari Holtzman, Mikel Artetxe, Mike Lewis, Hannaneh Hajishirzi, and Luke Zettlemoyer. 2022.
\newblock \href {https://doi.org/10.18653/v1/2022.emnlp-main.759} {Rethinking the role of demonstrations: What makes in-context learning work?}
\newblock In \emph{Proceedings of the 2022 Conference on Empirical Methods in Natural Language Processing}, pages 11048--11064, Abu Dhabi, United Arab Emirates. Association for Computational Linguistics.

\bibitem[{Olsson et~al.(2022)Olsson, Elhage, Nanda, Joseph, DasSarma, Henighan, Mann, Askell, Bai, Chen et~al.}]{olsson2022context}
Catherine Olsson, Nelson Elhage, Neel Nanda, Nicholas Joseph, Nova DasSarma, Tom Henighan, Ben Mann, Amanda Askell, Yuntao Bai, Anna Chen, et~al. 2022.
\newblock In-context learning and induction heads.
\newblock \emph{arXiv preprint arXiv:2209.11895}.

\bibitem[{Papangelis et~al.(2020)Papangelis, Namazifar, Khatri, Wang, Molino, and Tur}]{papangelis2020plato}
Alexandros Papangelis, Mahdi Namazifar, Chandra Khatri, Yi-Chia Wang, Piero Molino, and Gokhan Tur. 2020.
\newblock Plato dialogue system: A flexible conversational ai research platform.
\newblock \emph{arXiv preprint arXiv:2001.06463}.

\bibitem[{Papineni et~al.(2002)Papineni, Roukos, Ward, and Zhu}]{papineni-etal-2002-bleu}
Kishore Papineni, Salim Roukos, Todd Ward, and Wei-Jing Zhu. 2002.
\newblock \href {https://doi.org/10.3115/1073083.1073135} {{B}leu: a method for automatic evaluation of machine translation}.
\newblock In \emph{Proceedings of the 40th Annual Meeting of the Association for Computational Linguistics}, pages 311--318, Philadelphia, Pennsylvania, USA. Association for Computational Linguistics.

\bibitem[{Reynolds and McDonell(2021)}]{reynolds2021prompt}
Laria Reynolds and Kyle McDonell. 2021.
\newblock Prompt programming for large language models: Beyond the few-shot paradigm.
\newblock In \emph{Extended Abstracts of the 2021 CHI Conference on Human Factors in Computing Systems}, pages 1--7.

\bibitem[{Rubin et~al.(2022)Rubin, Herzig, and Berant}]{rubin2022learning}
Ohad Rubin, Jonathan Herzig, and Jonathan Berant. 2022.
\newblock Learning to retrieve prompts for in-context learning.
\newblock In \emph{Proceedings of the 2022 Conference of the North American Chapter of the Association for Computational Linguistics: Human Language Technologies}, pages 2655--2671.

\bibitem[{Sedgwick(2014)}]{sedgwick2014spearman}
Philip Sedgwick. 2014.
\newblock Spearman’s rank correlation coefficient.
\newblock \emph{Bmj}, 349.

\bibitem[{Serban et~al.(2017)Serban, Sordoni, Lowe, Charlin, Pineau, Courville, and Bengio}]{serban2017hierarchical}
Iulian Serban, Alessandro Sordoni, Ryan Lowe, Laurent Charlin, Joelle Pineau, Aaron Courville, and Yoshua Bengio. 2017.
\newblock A hierarchical latent variable encoder-decoder model for generating dialogues.
\newblock In \emph{Proceedings of the AAAI conference on artificial intelligence}, volume~31.

\bibitem[{Srivastava et~al.(2022)Srivastava, Rastogi, Rao, Shoeb, Abid, Fisch, Brown, Santoro, Gupta, Garriga-Alonso et~al.}]{srivastava2022beyond}
Aarohi Srivastava, Abhinav Rastogi, Abhishek Rao, Abu Awal~Md Shoeb, Abubakar Abid, Adam Fisch, Adam~R Brown, Adam Santoro, Aditya Gupta, Adri{\`a} Garriga-Alonso, et~al. 2022.
\newblock Beyond the imitation game: Quantifying and extrapolating the capabilities of language models.
\newblock \emph{arXiv preprint arXiv:2206.04615}.

\bibitem[{Su et~al.(2022)Su, Kasai, Wu, Shi, Wang, Xin, Zhang, Ostendorf, Zettlemoyer, Smith et~al.}]{su2022selective}
Hongjin Su, Jungo Kasai, Chen~Henry Wu, Weijia Shi, Tianlu Wang, Jiayi Xin, Rui Zhang, Mari Ostendorf, Luke Zettlemoyer, Noah~A Smith, et~al. 2022.
\newblock Selective annotation makes language models better few-shot learners.
\newblock \emph{arXiv preprint arXiv:2209.01975}.

\bibitem[{Sun et~al.(2022)Sun, Liu, Qiu, and Huang}]{sun2022paradigm}
Tian-Xiang Sun, Xiang-Yang Liu, Xi-Peng Qiu, and Xuan-Jing Huang. 2022.
\newblock Paradigm shift in natural language processing.
\newblock \emph{Machine Intelligence Research}, 19(3):169--183.

\bibitem[{Sun et~al.(2021)Sun, Wang, Feng, Ding, Pang, Shang, Liu, Chen, Zhao, Lu et~al.}]{sun2021ernie}
Yu~Sun, Shuohuan Wang, Shikun Feng, Siyu Ding, Chao Pang, Junyuan Shang, Jiaxiang Liu, Xuyi Chen, Yanbin Zhao, Yuxiang Lu, et~al. 2021.
\newblock Ernie 3.0: Large-scale knowledge enhanced pre-training for language understanding and generation.
\newblock \emph{arXiv preprint arXiv:2107.02137}.

\bibitem[{Woolson(2007)}]{woolson2007wilcoxon}
Robert~F Woolson. 2007.
\newblock Wilcoxon signed-rank test.
\newblock \emph{Wiley encyclopedia of clinical trials}, pages 1--3.

\bibitem[{Wu et~al.(2023)Wu, Wang, Ye, and Kong}]{wu-etal-2023-self}
Zhiyong Wu, Yaoxiang Wang, Jiacheng Ye, and Lingpeng Kong. 2023.
\newblock \href {https://doi.org/10.18653/v1/2023.acl-long.79} {Self-adaptive in-context learning: An information compression perspective for in-context example selection and ordering}.
\newblock In \emph{Proceedings of the 61st Annual Meeting of the Association for Computational Linguistics (Volume 1: Long Papers)}, pages 1423--1436, Toronto, Canada. Association for Computational Linguistics.

\bibitem[{Xiong et~al.(2023)Xiong, Liu, Molybog, Zhang, Bhargava, Hou, Martin, Rungta, Sankararaman, Oguz et~al.}]{xiong2023effective}
Wenhan Xiong, Jingyu Liu, Igor Molybog, Hejia Zhang, Prajjwal Bhargava, Rui Hou, Louis Martin, Rashi Rungta, Karthik~Abinav Sankararaman, Barlas Oguz, et~al. 2023.
\newblock Effective long-context scaling of foundation models.
\newblock \emph{arXiv preprint arXiv:2309.16039}.

\bibitem[{Xu et~al.(2023)Xu, Lei, Wu, Niu, Wu, and Wang}]{xu2023towards}
Xinchao Xu, Zeyang Lei, Wenquan Wu, Zheng-Yu Niu, Hua Wu, and Haifeng Wang. 2023.
\newblock Towards zero-shot persona dialogue generation with in-context learning.
\newblock In \emph{Findings of the Association for Computational Linguistics: ACL 2023}, pages 1387--1398.

\bibitem[{Yan et~al.(2023)Yan, Xu, Song, Wu, Li, and Zhang}]{yan2023understanding}
Jianhao Yan, Jin Xu, Chiyu Song, Chenming Wu, Yafu Li, and Yue Zhang. 2023.
\newblock Understanding in-context learning from repetitions.
\newblock \emph{arXiv preprint arXiv:2310.00297}.

\bibitem[{Ye et~al.(2023{\natexlab{a}})Ye, Wu, Feng, Yu, and Kong}]{ye2023compositional}
Jiacheng Ye, Zhiyong Wu, Jiangtao Feng, Tao Yu, and Lingpeng Kong. 2023{\natexlab{a}}.
\newblock Compositional exemplars for in-context learning.
\newblock \emph{arXiv preprint arXiv:2302.05698}.

\bibitem[{Ye et~al.(2023{\natexlab{b}})Ye, Chen, Xu, Zu, Shao, Liu, Cui, Zhou, Gong, Shen et~al.}]{ye2023comprehensive}
Junjie Ye, Xuanting Chen, Nuo Xu, Can Zu, Zekai Shao, Shichun Liu, Yuhan Cui, Zeyang Zhou, Chao Gong, Yang Shen, et~al. 2023{\natexlab{b}}.
\newblock A comprehensive capability analysis of gpt-3 and gpt-3.5 series models.
\newblock \emph{arXiv preprint arXiv:2303.10420}.

\bibitem[{Zhu et~al.(2018)Zhu, Lu, Zheng, Guo, Zhang, Wang, and Yu}]{zhu2018texygen}
Yaoming Zhu, Sidi Lu, Lei Zheng, Jiaxian Guo, Weinan Zhang, Jun Wang, and Yong Yu. 2018.
\newblock Texygen: A benchmarking platform for text generation models.
\newblock In \emph{The 41st international ACM SIGIR conference on research \& development in information retrieval}, pages 1097--1100.

\end{thebibliography}
\bibliographystyle{acl_natbib}

\appendix
\clearpage

\section{Appendix}
\label{sec:appendix}
In the appendix, we provide a more detailed introduction to the evaluation metrics for dialogue generation in \S\ref{sec:eval_metrics} and explain how we select and retain the best prompt in \S\ref{sec:prompt_select}.

In \S\ref{sec:examples_filled_templates} and \S\ref{sec:examples_filled_templates_ablation}, we showcase filled templates under various settings, allowing for a clear representation of the LLMs' input text.

We describe the training approach for the Dialogue Embedding model in Section\S\ref{sec:train_dialog_embedder} and outline the training data and training details for the Response Evaluator in \S\ref{sec:train_response_eval}.

In \S\ref{sec:appendix_more_exp_results}, we present additional interesting experimental findings.

In \S\ref{sec:anno_details}, we provide a detailed overview of the annotator background, the manual composition process for high-quality reference responses from human experts, the evaluation consistency between the Response Evaluator, as well as the specific annotation guidelines.

\begin{table*}[h!]
\centering
\small
\begin{tabular}{@{}llll@{}}
\toprule
 & Prompt Construction Method & Response Quality & Number of Few Shot $k$ to Exceed \emph{Prompt Only} Method \\ \midrule
GPT-3.5 & Few Shot Demo (Random) & \textbf{0.188} & 7 \\
 & \hspace{1em} - w/o correct y label & $\phantom{-}\Delta$0.009$\dag$ & 7 \\
 & \hspace{1em} - w/o contextual semantic & $\phantom{-}\Delta$0.005 & 7 \\
 & \hspace{1em} - w/o turn semantic & $-\Delta$0.010$\dag$ & Unknown \\
 & \hspace{1em} - w/o contextual/turn semantic & $-\Delta$0.026$\dag$ & Unknown \\
 & Context Only & \textbf{0.147} & - \\ \midrule
GPT-4 & Few Shot Demo (Random) & \textbf{0.227} & Unknown \\
 & \hspace{1em} - w/o correct y label & $-\Delta$0.041$\dag$ & Unknown \\
 & \hspace{1em} - w/o contextual semantic & $\phantom{-}\Delta$0.004 & Unknown \\
 & \hspace{1em} - w/o turn semantic & $-\Delta$0.036$\dag$ & Unknown \\
 & \hspace{1em} - w/o contextual/turn semantic & $-\Delta$0.042$\dag$ & Unknown \\
 & Context Only & \textbf{0.155} & - \\ \midrule
Ernie & Few Shot Demo (Random) & \textbf{0.5022} & 9 \\
 & \hspace{1em} - w/o correct y label & $-\Delta$0.069$\dag$ & Unknown \\
 & \hspace{1em} - w/o contextual semantic & $-\Delta$0.005 & Unknown \\
 & \hspace{1em} - w/o turn semantic & $-\Delta$0.055$\dag$ & Unknown \\
 & \hspace{1em} - w/o contextual/turn semantic & $-\Delta$0.040$\dag$ & Unknown \\
 & Context Only & \textbf{0.345} & - \\ \bottomrule
\end{tabular}
\caption{Results of dialogue generation when scores are averaged over few-shot $k$ of 1,3,5,7,9 and 11.}
\label{tab:result_main_creative_k_all}
\end{table*}

\subsection{More Details on Experimental Settings}
\subsubsection{Evaluation Metrics for Generation}
\label{sec:eval_metrics}
In the experiment, given a persona description and a context, we let the LLM generate 15 responses, from which we retain a set of 5 responses that has the highest intra-diversity (the retention process finds the combination of 5 responses with the largest unique token set among all possible combinations). We adopt this setting because the responses generated by the LLM given the same context can be quite similar. We denote the retained response set as $\hat{\textbf{S}_r}$, and the set of 5 expert-written responses as $\textbf{S}_{r}$. We analyze the quality of the responses along three dimensions, as shown below.

\noindent \textbf{Intra-Diversity}: we calculate the Distinct-$n$ ($n$=1,2,3)~\citep{distinct}, Entropy-$n$ ($n$=1)~\citep{serban2017hierarchical}, $1-\textit{self-bleu score}$~\citep{zhu2018texygen}, and $1-\textit{cosine-similarity score (mean)}$ for $\hat{S}_r$. We denote the cosine similarity between two unequal responses as $v_{cos}$. The embeddings of two responses are extracted by our self-trained encoder (see \S\ref{sec:train_dialog_embedder} for more details). We iterate through all unequal pairs in $\hat{S}_r$ to compute $v_{cos}$ values, and the average cosine similarity is the mean of all $v_{cos}$ values. The final diversity score $s_{div}$ is the average of the four scores, with higher values indicating better response diversity in $\hat{S}_r$.

\noindent \textbf{Inter-Similarity}: To calculate the similarity score $s_{sim}$ between $\hat{\textbf{S}_r}$ and $\textbf{S}_r$, we first iterate through the responses in $\hat{\textbf{S}_r}$, and compute the similarity score between the generated response $\hat{r}$ and $\textbf{S}_r$. This involves calculating the BLEU score~\citep{papineni-etal-2002-bleu}, ROUGE-L score~\citep{lin-2004-rouge}, token and character-level overlap ratios (calculated as the percentage of unique token/character of $\hat{r}$ in $\textbf{S}_{r}$), as well as the average cosine similarity between $\hat{r}$ and each response in $\textbf{S}_r$. We take the mean of the five scores above as the similarity score between $\hat{r}$ and $\hat{\textbf{S}_r}$. We obtain the final similarity score $s_{sim}$ between $\hat{\textbf{S}_r}$ and $\textbf{S}_r$ by averaging over all generated responses.

\noindent \textbf{Response Quality}: The response quality is scored by a specially customized Response Evaluator $f_{eval}$ (see \S\ref{sec:train_response_eval}), trained on $50,000+$ samples. The evaluator $f_{eval}(per,c_{t-1},\hat{r_t})$ outputs a score between 0 and 1, representing the quality of the response $\hat{r_t}$. The closer the score is to 1, the better the model's reply $\hat{r}$ performs in terms of persona consistency, contextual logic, interestingness, and colloquial expression. We verify the correlation of the Response Evaluator with expert ratings, which is significantly higher than crowd-sourced annotations and other automated LLM-based evaluation methods (details can be referred to 
\S\ref{sec:anno_consist}).

\subsubsection{Prompt Selection Process}
\label{sec:prompt_select}
In our experiment, the results of prompt-related methods are based on one selected prompt. We elaborate on how the best prompt is selected.

All co-authors are asked to write 10 candidate prompts, from which one best prompt is evaluated (by all co-authors) and kept based on its performance (considering both the intra-diversity and response quality) on the held-out dataset. Specific details of the retained prompt can be found in Table~\ref{tab:selected_prompt}.

\subsection{Examples of Filled Templates}
\label{sec:examples_filled_templates}
Examples of filled templates for \emph{Context Only}, \emph{Prompt only}, \emph{Few shot demo} and \emph{Few shot demo with prompt} can be found in \Cref{tab:llm_input_example_context_only,tab:llm_input_example_prompt_only,tab:llm_input_example_few_shot_demo_random,tab:llm_input_example_few_shot_demo_random_with_prompt}.

\subsection{Examples of Filled Demonstrations under Different Ablation Settings}
\label{sec:examples_filled_templates_ablation}
Examples of different context ablation settings can be found in \Cref{tab:demo_example_wo_correct_y_label,tab:demo_example_wo_contextual_semantic,tab:demo_example_wo_turn_semantic,tab:demo_example_wo_contextual_turn_semantic} respectively.

\subsection{Training Details of the Dialogue Embedding Model}
\label{sec:train_dialog_embedder}
The data used for training the Dialogue Embedding Model comes from $\mathcal{D}_{demo}$, and the method we employ is SimCSE~\citep{gao2021simcse}. One positive sample pair is $(\mathcal{T}_{pc}(per, c_{:t-1}), r_t)$, where $\mathcal{T}_{pc}$ is a template function that concatenates persona description and context (separated by $\mathtt{[SEP]}$). One negative sample pair is $(\mathcal{T}_{pc}(per, c_{:t-1}), r_{rand})$, where $r_{rand}$ is a randomly sampled response (drawn from the set of all responses in $\mathcal{D}_{demo}$). Our SimCSE two-tower model uses Roberta-Large~\citep{liu2019roberta} as the text encoder, with non-shared parameters for the left and right towers. Therefore, whether to use the left tower or the right tower depends on the application scenario. Use the left tower to retrieve context based on the context, and use the right tower to retrieve responses based on the responses.

We train our own Embedding Model to ensure that the data distribution is as close as possible to the setting of persona-based dialogues, thus improving the recall of similar contexts. Existing Chinese embedding models (such as sbert-base-chinese-nli\footnote{https://huggingface.co/uer/sbert-base-chinese-nli}, text2vec-base-chinese-paraphrase\footnote{https://huggingface.co/shibing624/text2vec-base-chinese-paraphrase}, bge-large-zh-v1.5\footnote{https://huggingface.co/BAAI/bge-large-zh-v1.5}, etc.) do not include training data in the form of persona and context. Consequently, their performance in recalling dialogue responses is significantly lower than our model, with an MRR (mean reciprocal rank) that is more than 10 points lower.

\subsection{Training Details of the Response Evaluator}
\label{sec:train_response_eval}
The training data for the Response Evaluator $f_{eval}$ consists of two parts. \textbf{The first part} is dialogue data $\mathcal{D}_{demo}$ created by professional writers (500 dialogues in total), each dialogue including one persona description and 40 rounds of conversations ($\ell$ equals 40). By iterating $t (0<=t<\ell)$ through the entire dialogue, we obtain positive samples $\mathcal{T}_{eval}(per, c_{:t-1}, r_t)$ and negative samples $\mathcal{T}_{eval}(per, c_{:t-1}, r_{neg})$. The negative response $r_{neg}$ has two sources: the first is the random extraction from the Chinese Novel Corpus\footnote{We have collected novel data from multiple sources, including web novels and classic novels. The Corpus contains around 1,000,000 unique dialogue turns in total.}, and the second is retrieving responses similar to the context $c_{t-2:t-1}$ using the \emph{sbert-base-chinese-nli} model (from the set of all responses in $\mathcal{D}_{demo}$). The template function $\mathcal{T}_{eval}$ simply concatenates the persona, context, and response together, separating them with $\mathtt{[SEP]}$.

\textbf{The second part} comes from real-time scoring and rewriting tasks. We train a Chinese-LLaMA\footnote{https://github.com/ymcui/Chinese-LLaMA-Alpaca} model $\mathcal{LLM}_{cn}$ and design a conversational task where annotators chat and revise the model's responses. The annotators are required to engage in 40 rounds of conversation with $\mathcal{LLM}_{cn}$, modifying their responses when they are unsatisfactory. We can naturally obtain paired data (human-modified responses as positive samples and $\mathcal{LLM}_{cn}$ generated responses as negative samples) from these revised conversations. The effective data size for this type of data is around 5,000.

We chose Roberta-Large as our Response Evaluator. Due to its position encoding length limitation, we truncate the persona description and the context when any of them exceeds 250 tokens. In most cases, this preserves a complete 7-10 rounds of dialogue. We set the learning rate and batch size to 1e-5 and 32, respectively, and allocate 5\% of the training data as a validation set. We retain the model with the lowest loss on the validation set. The loss function for training the model is Binary Cross Entropy. Please refer to \S\ref{sec:anno_consist} for specific annotation quality of the Response Evaluator.

\subsection{More Experimental Analysis}
\label{sec:appendix_more_exp_results}

\subsubsection{ICL is much more than $n$-gram induction heads}
\begin{figure*}
    \centering
    \includegraphics[width=0.325\textwidth]{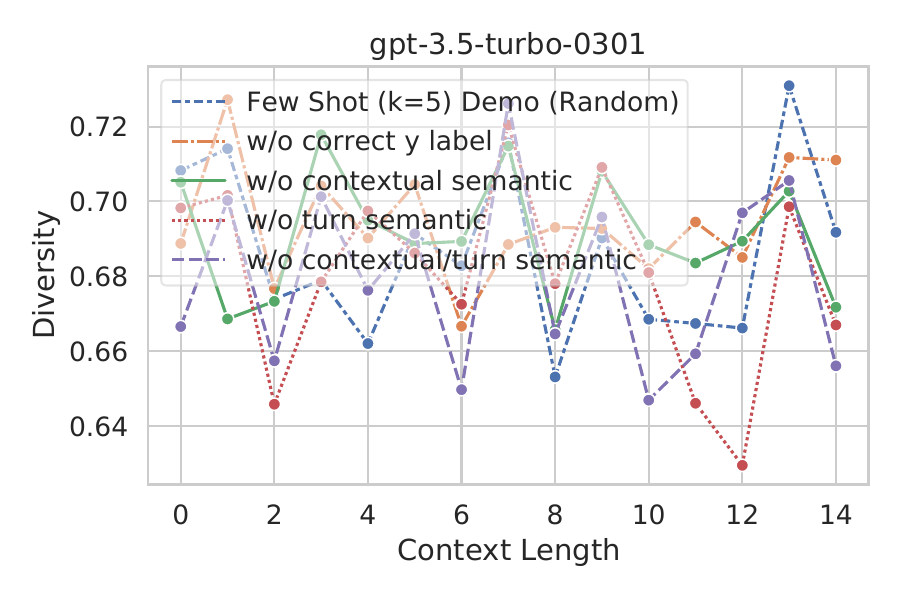}
    \includegraphics[width=0.325\textwidth]{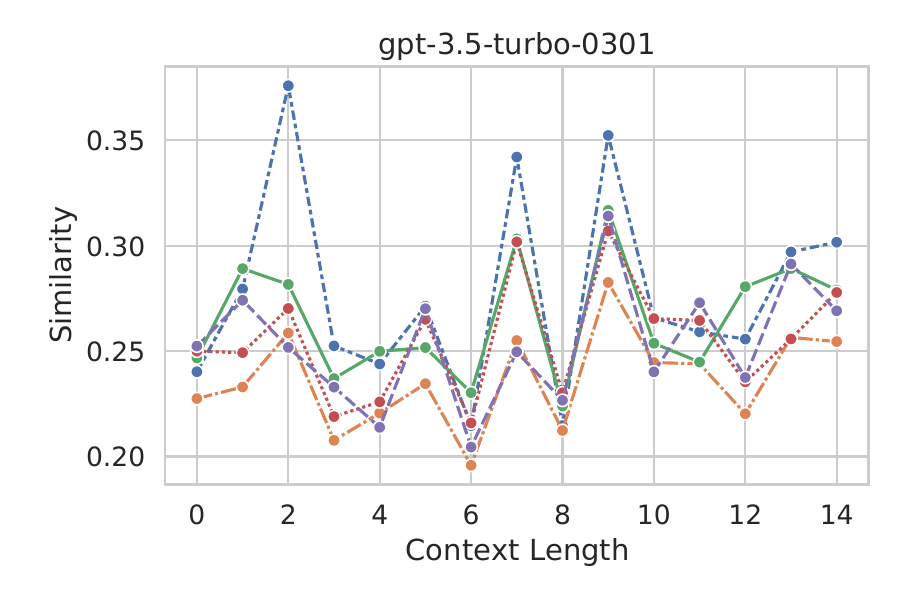}
    \includegraphics[width=0.325\textwidth]{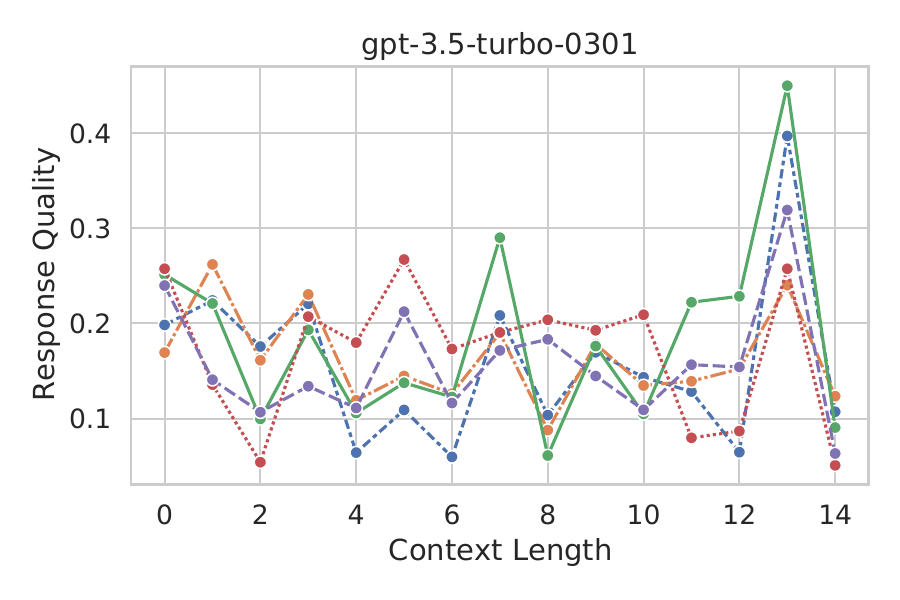}
    \includegraphics[width=0.325\textwidth]{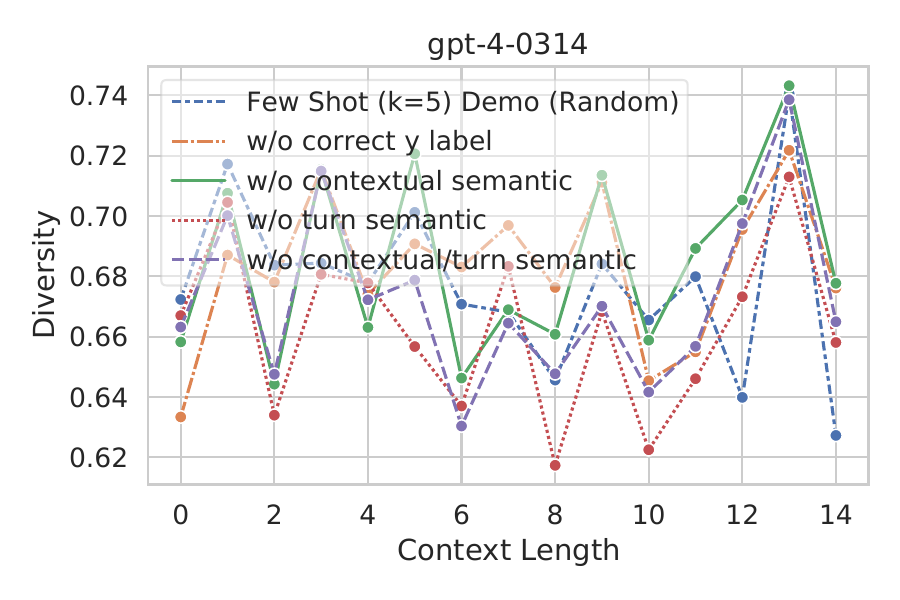}
    \includegraphics[width=0.325\textwidth]{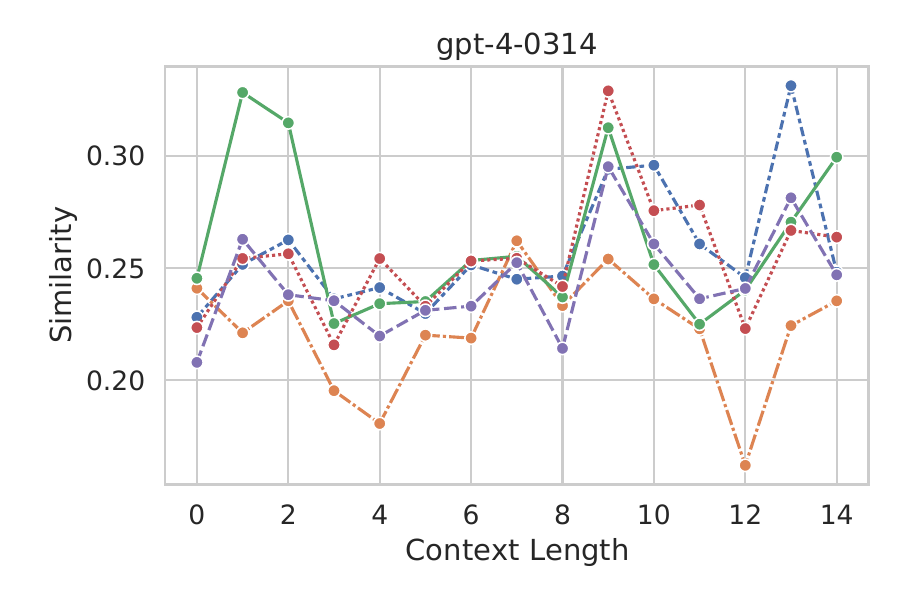}
    \includegraphics[width=0.325\textwidth]{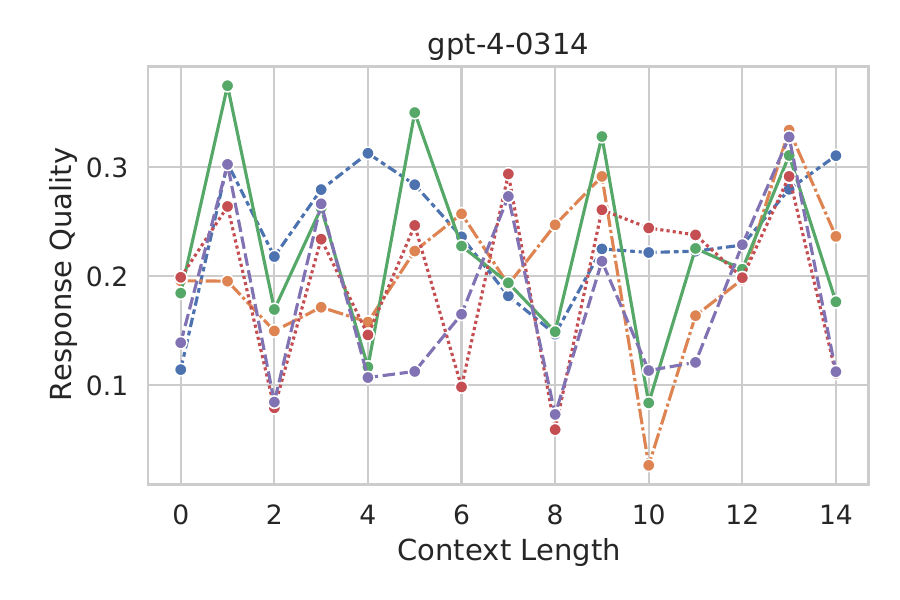}
    \includegraphics[width=0.325\textwidth]{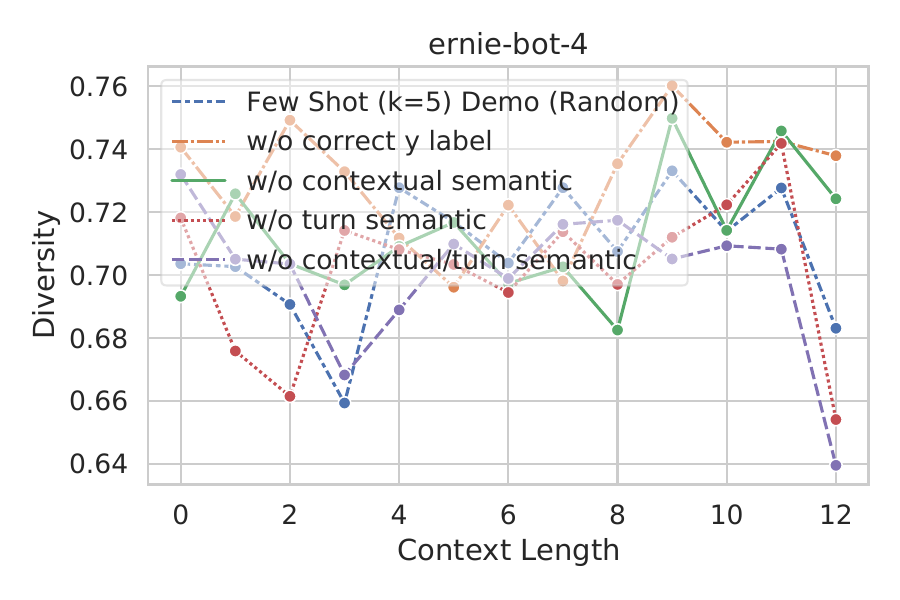}
    \includegraphics[width=0.325\textwidth]{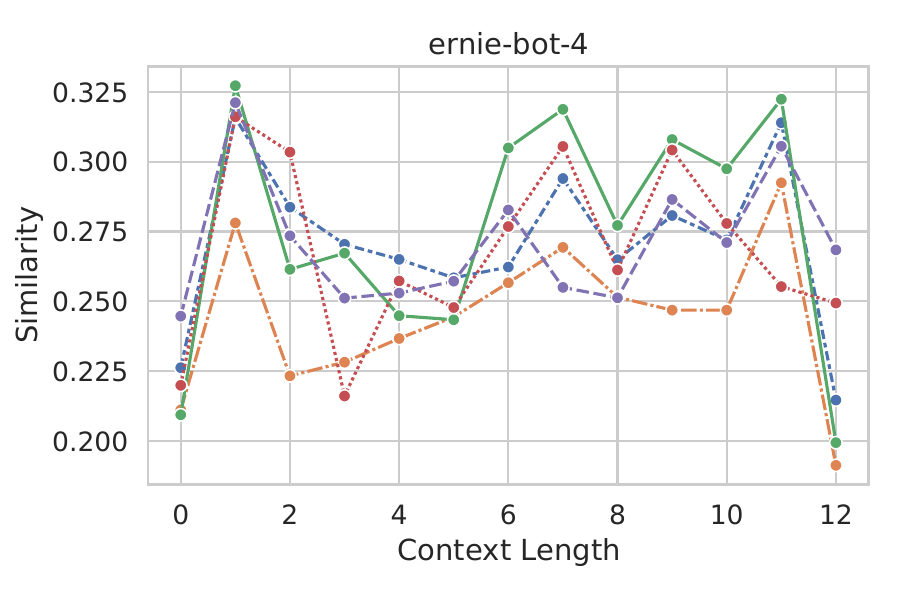}
    \includegraphics[width=0.325\textwidth]{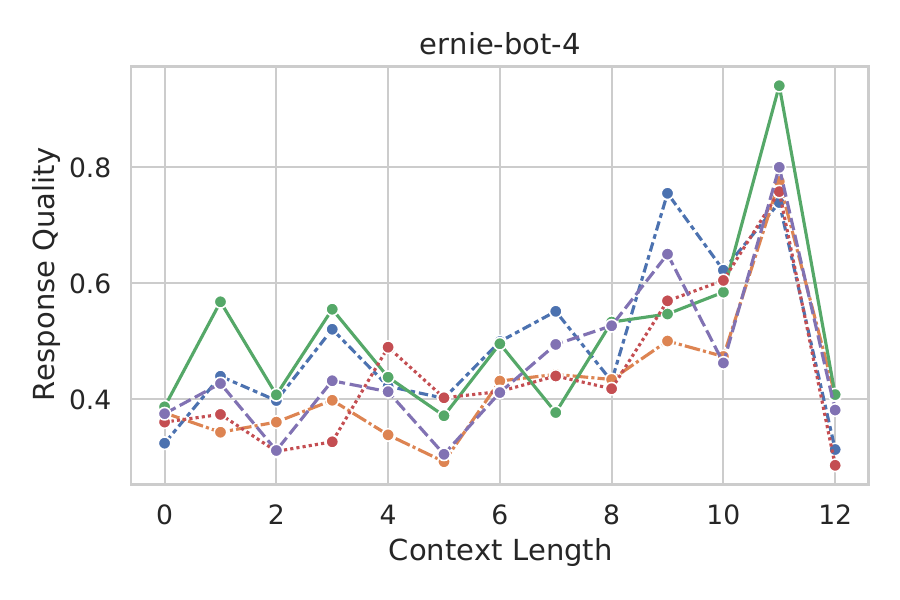}
    \caption{The impact of label substitution and different semantic corruption methods on diversity, similarity, and response quality when the context length varies while the number of few-shot demonstrations remains fixed ($k=5$). }
    \label{fig:few_shot_5_corrupted_context}
\end{figure*}

There are works explaining ICL from the perspectives of induction head~\citep{olsson2022context} and $n$-gram head~\citep{akyürek2024incontext}, but we believe that these theories cannot fully explain the phenomenon of \emph{ernie-bot-4 w/o contextual/turn semantic} method continuously improving as the number of demos increases (see Figure~\ref{fig:few_shot_5_corrupted_context} and Figure~\ref{fig:heatmap_ernie_bot_corrupted_context}). This is because, in this setting, the $n$-gram information of the demos does not correspond to that of the query. During ICL, the relationship between the $n$-grams in the demos and their following text are of no reference value as the order of the $n$-grams is random.

\subsubsection{Is it possible to balance diversity while ensuring the quality of responses?}
\begin{figure*}
    \centering
    \includegraphics[width=0.325\textwidth]{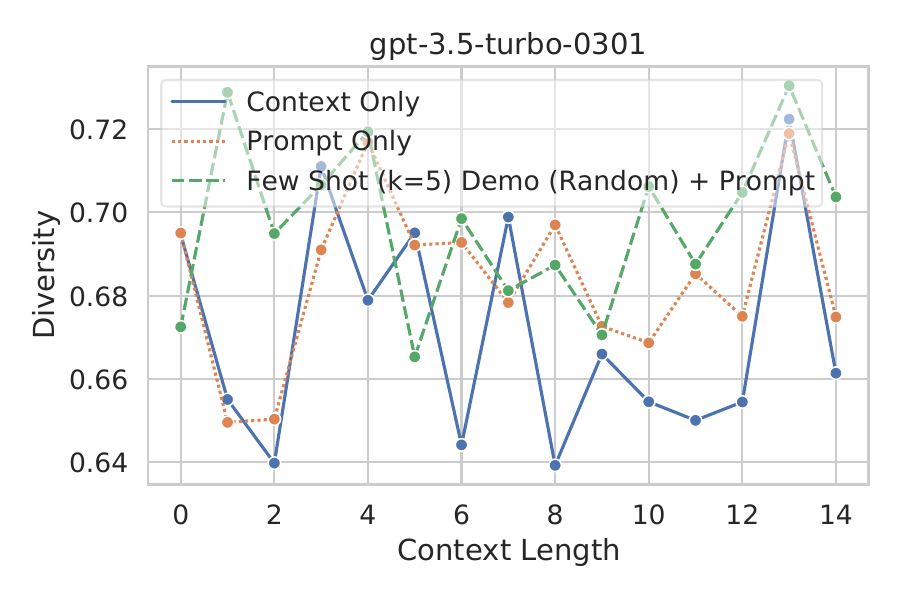}
    \includegraphics[width=0.325\textwidth]{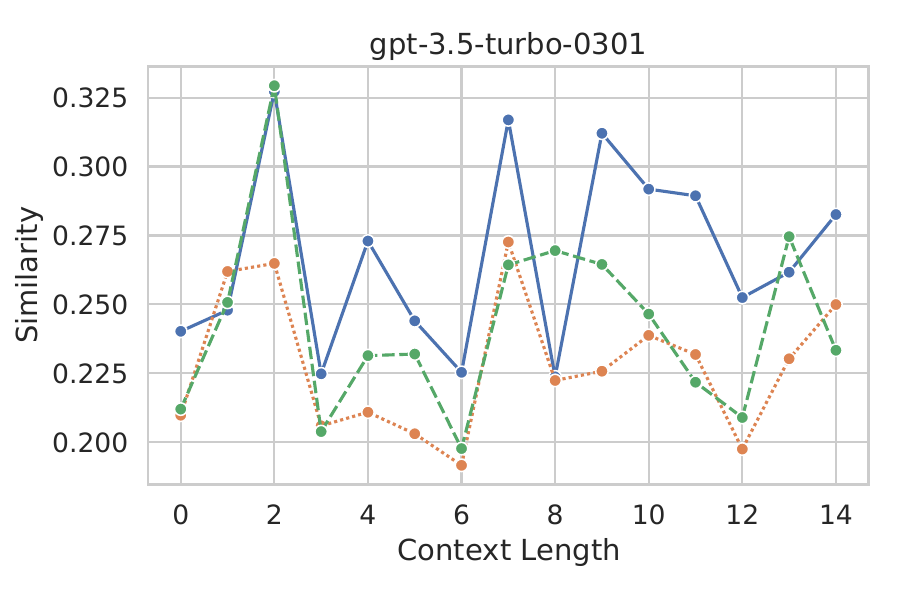}
    \includegraphics[width=0.325\textwidth]{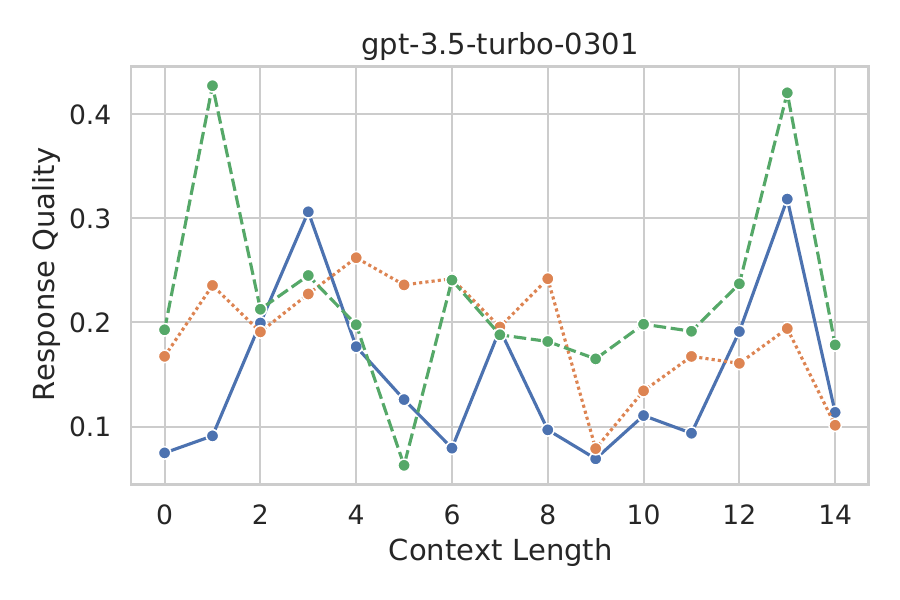}
    \includegraphics[width=0.325\textwidth]{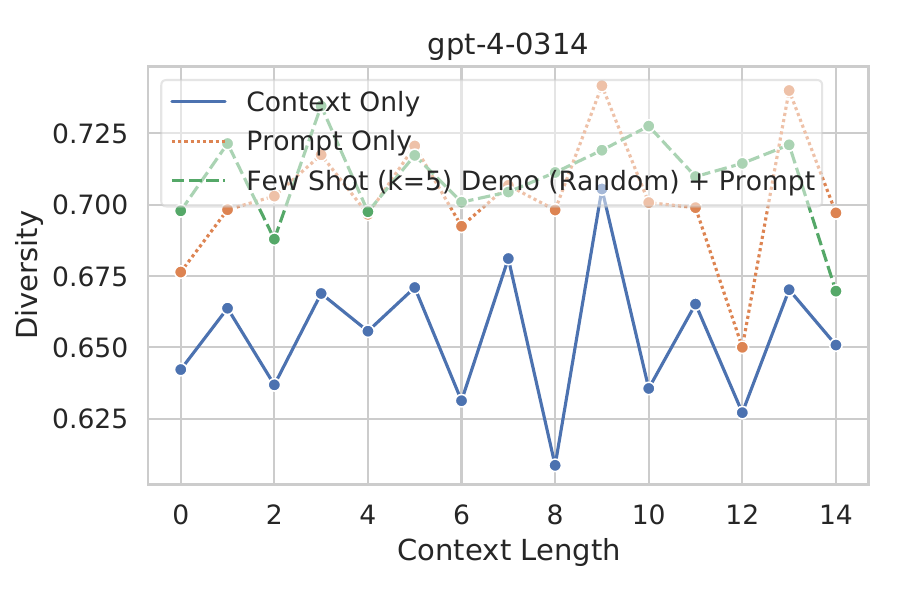}
    \includegraphics[width=0.325\textwidth]{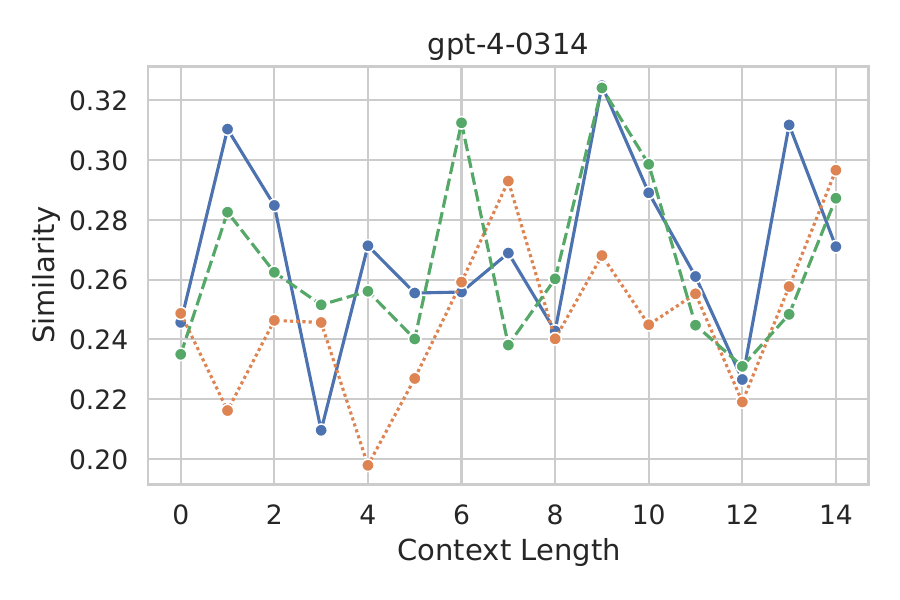}
    \includegraphics[width=0.325\textwidth]{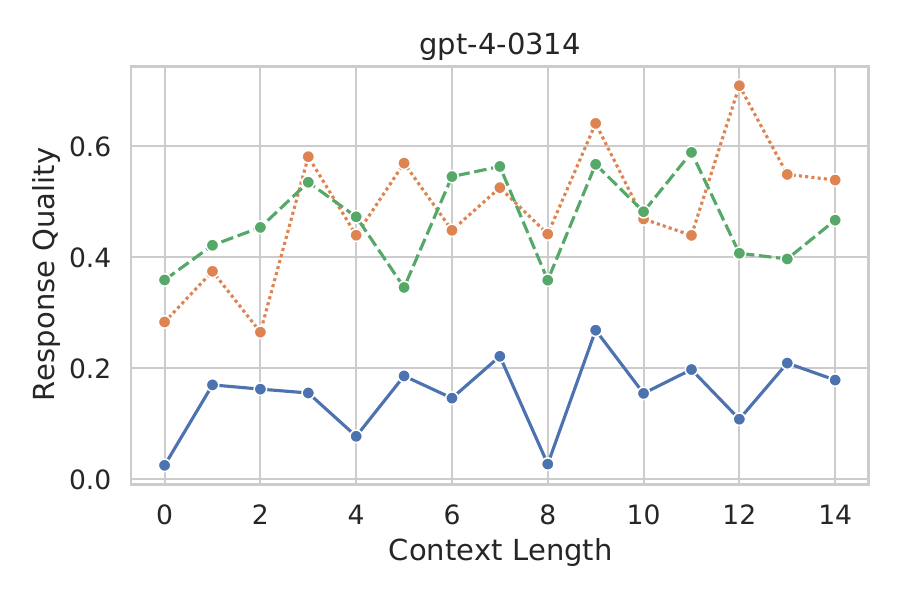}
    \includegraphics[width=0.325\textwidth]{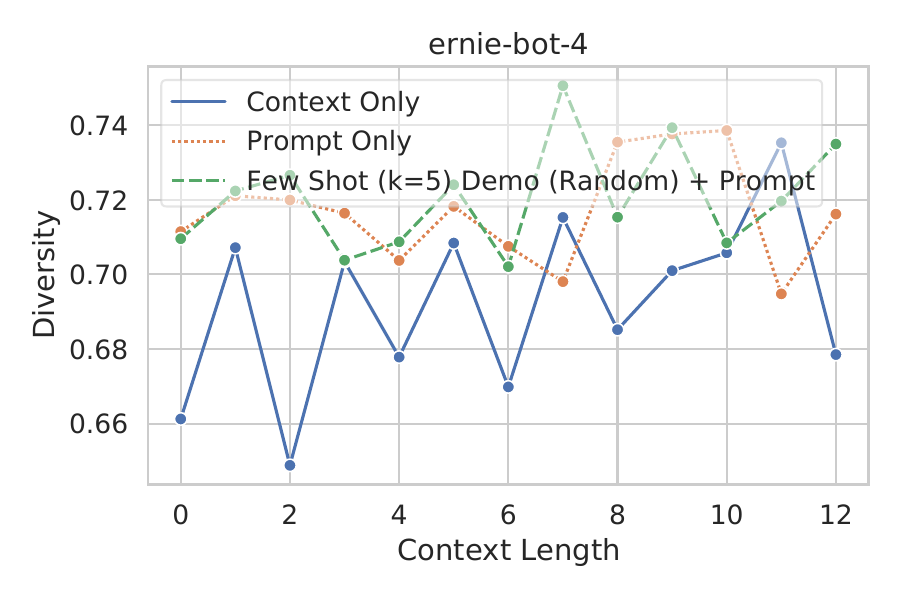}
    \includegraphics[width=0.325\textwidth]{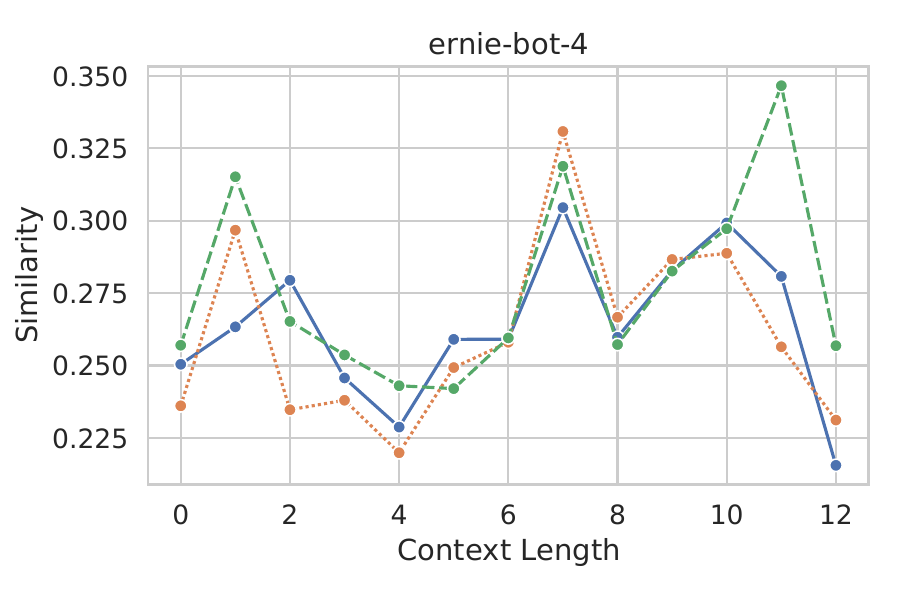}
    \includegraphics[width=0.325\textwidth]{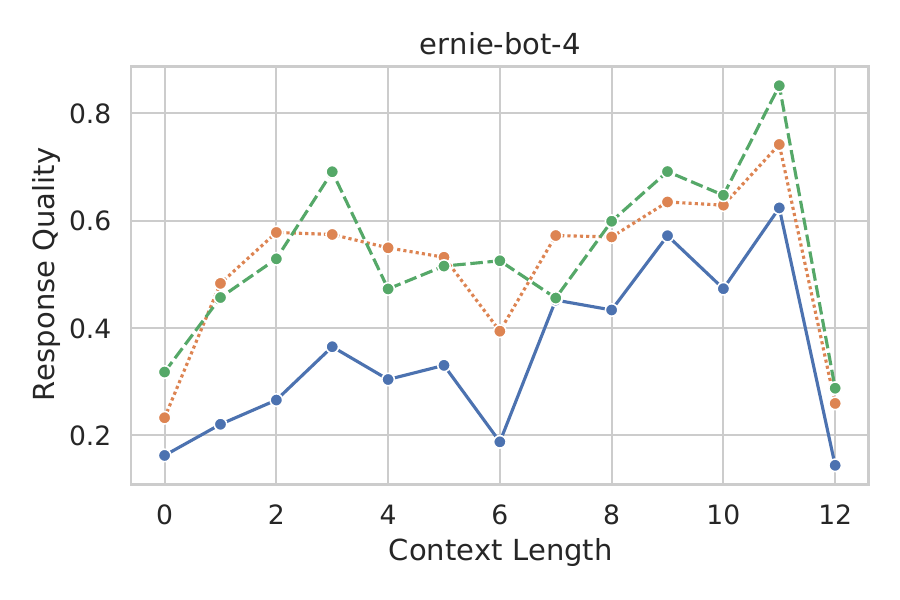}
    \caption{The performance comparison among \emph{Context Only} method, \emph{Prompt Only} method, and \emph{using both prompt and demonstration} when the context length varies while the number of few-shot demonstrations remains fixed ($k=5$).}
    \label{fig:few_shot_5_context_vs_prompt_vs_demo}
\end{figure*}

Yes, it is possible. As seen in Figure~\ref{fig:few_shot_5_context_vs_prompt_vs_demo}, the optimal approach (using both task prompt and demos) outperforms the demo-only method on both response quality and diversity. The experimental results demonstrate that in the task of dialogue generation, diversity, and response quality are not necessarily a trade-off.

\subsubsection{Does the context length of the demonstrations have an impact on the ICL-based dialogue generation?}

\begin{figure*}
    \centering
    \includegraphics[width=0.325\textwidth]{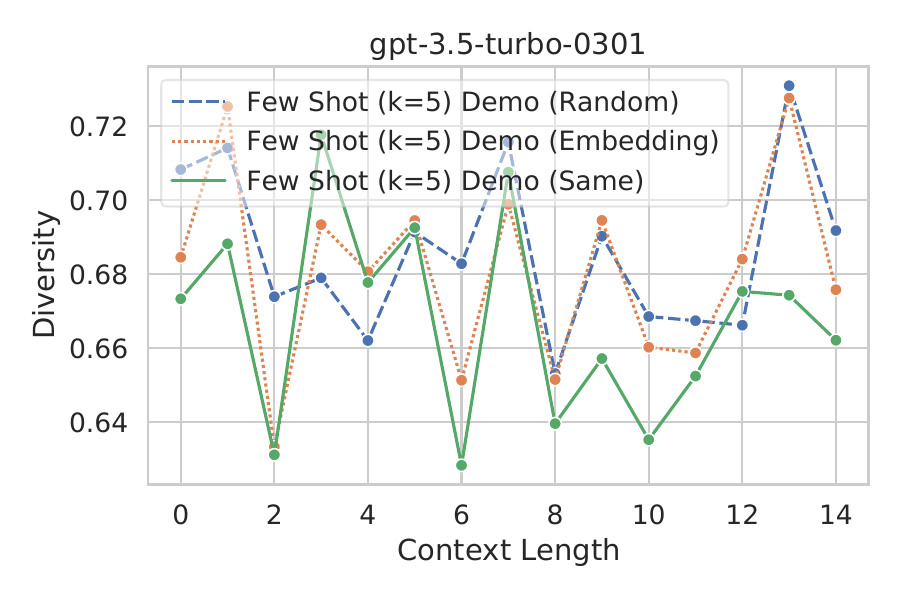}
    \includegraphics[width=0.325\textwidth]{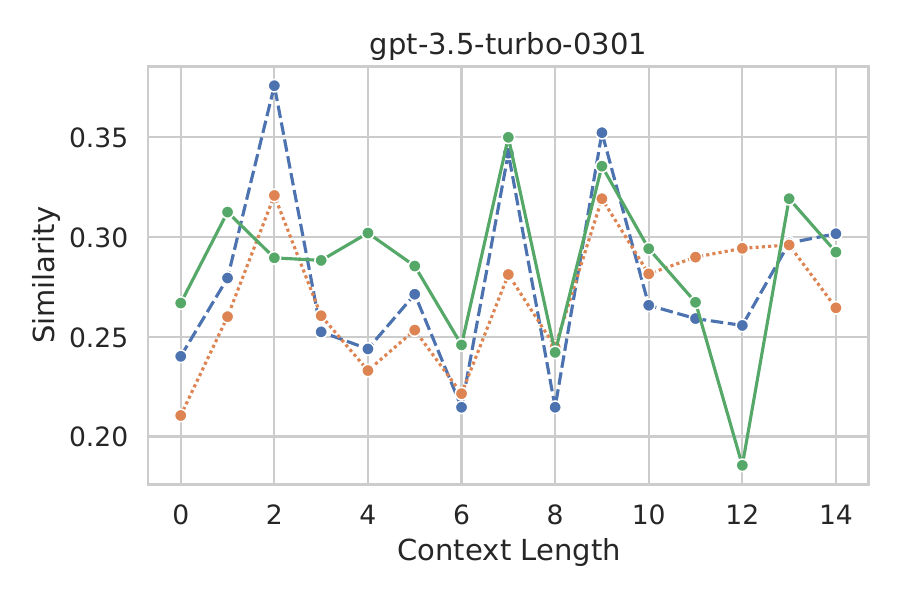}
    \includegraphics[width=0.325\textwidth]{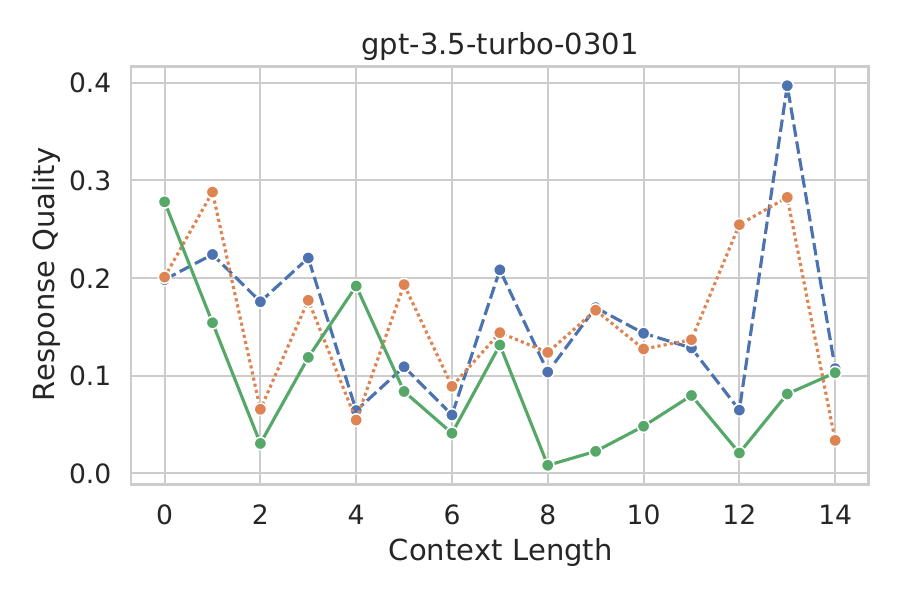}
    \includegraphics[width=0.325\textwidth]{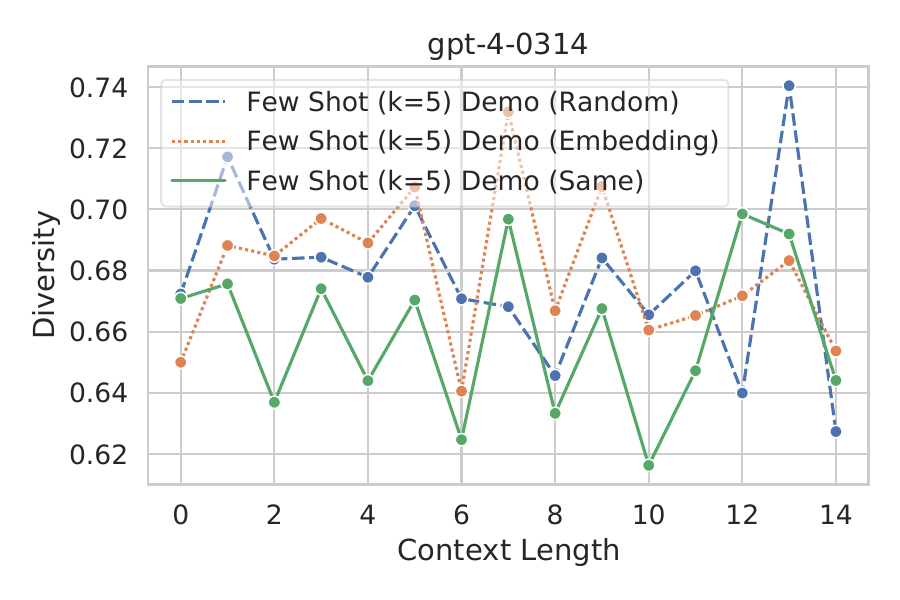}
    \includegraphics[width=0.325\textwidth]{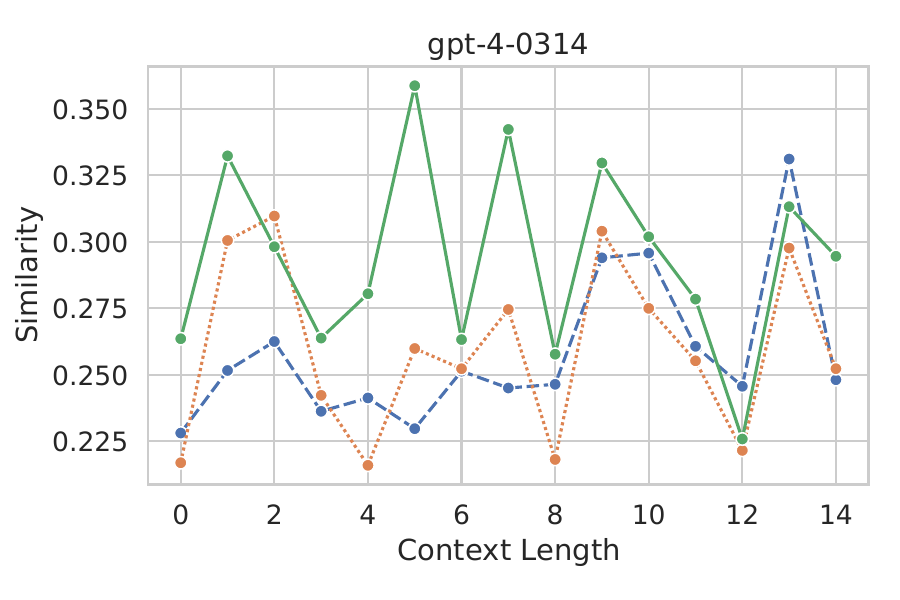}
    \includegraphics[width=0.325\textwidth]{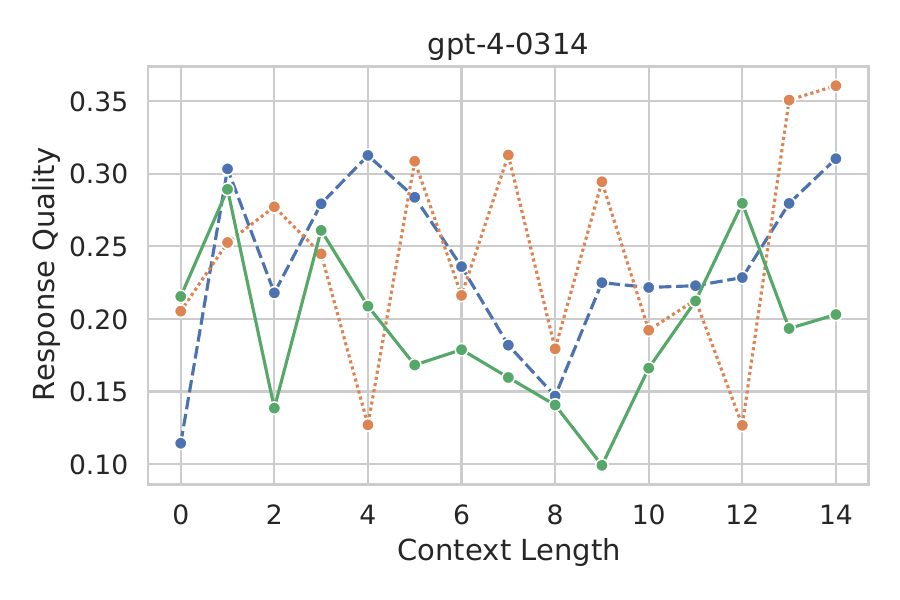}
    \includegraphics[width=0.325\textwidth]{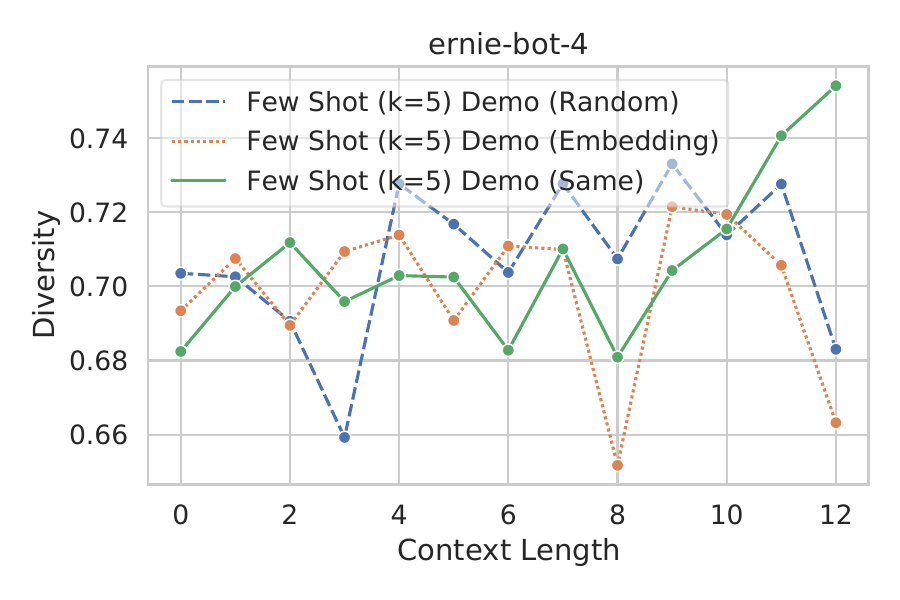}
    \includegraphics[width=0.325\textwidth]{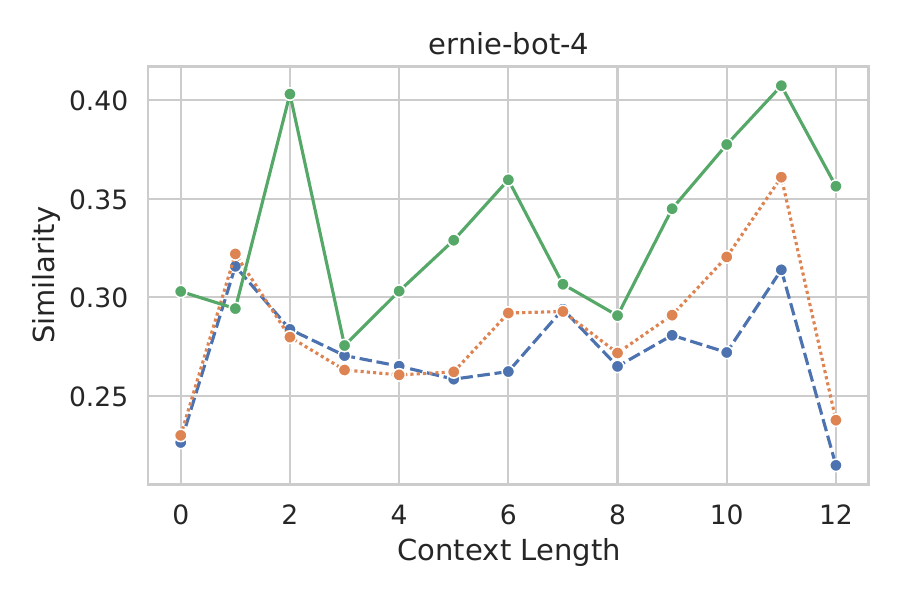}
    \includegraphics[width=0.325\textwidth]{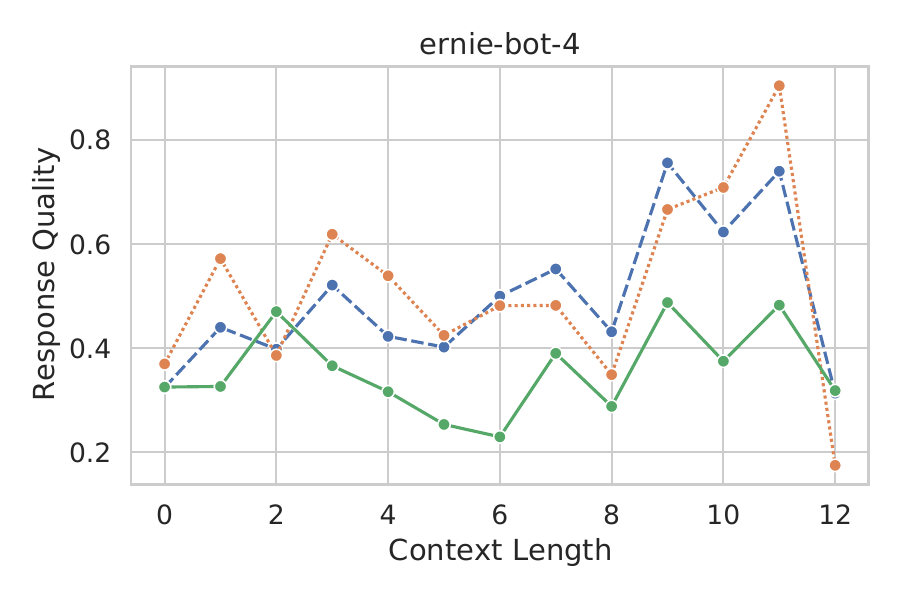}
    \caption{The performance comparison among three retrieval methods when the context length varies while the number of few-shot demonstrations remains fixed ($k=5$).}
    \label{fig:few_shot_5_demo_retrieval_method}
\end{figure*}

From~\Cref{fig:few_shot_5_corrupted_context,fig:few_shot_5_context_vs_prompt_vs_demo,fig:few_shot_5_demo_retrieval_method}, we demonstrate the impact of different semantic corruption methods, different prompt setting, and different retrieval methods on dialogue generation respectively as the context length varies. From~\Cref{fig:heatmap_gpt_35_corrupted_context,fig:heatmap_gpt4_corrupted_context,fig:heatmap_ernie_bot_corrupted_context,fig:heatmap_different_retrieval_method}, we demonstrate the variations in response quality as the context length and few-shot $k$ change simultaneously.

From these figures, we observe that \emph{different LLMs have varying sensitivity to the changes in demo context length}. The \emph{ernie-bot-4} model is the most sensitive, while \emph{gpt-3.5} and \emph{gpt-4} only exhibit an increase in response quality with context length under specific few-shot settings or special conditions (w/o correct y label). We hypothesize that this is due to ernie-bot-4's higher efficiency in absorbing knowledge during ICL and its robustness to the format of $\mathbf{x}_{demo}$, allowing it to learn from corrupted $\widetilde{\mathbf{x}_{demo}}$. The reasons why \emph{gpt-3.5} and \emph{gpt-4} cannot extract more useful information from longer contexts during ICL needs further research.

\subsubsection{Are responses generated by LLMs most similar to the nearest demo's response?}

The key difference between the \emph{embedding} retrieval method and the other two retrieval methods is that the context of the embedding-retrieved demo and the context of the query have a similar relationship, i.e., the closer the demo is to the query, the more similar their contexts are. 

\begin{figure*}[htb]
    \centering
    \includegraphics[width=0.496\textwidth]{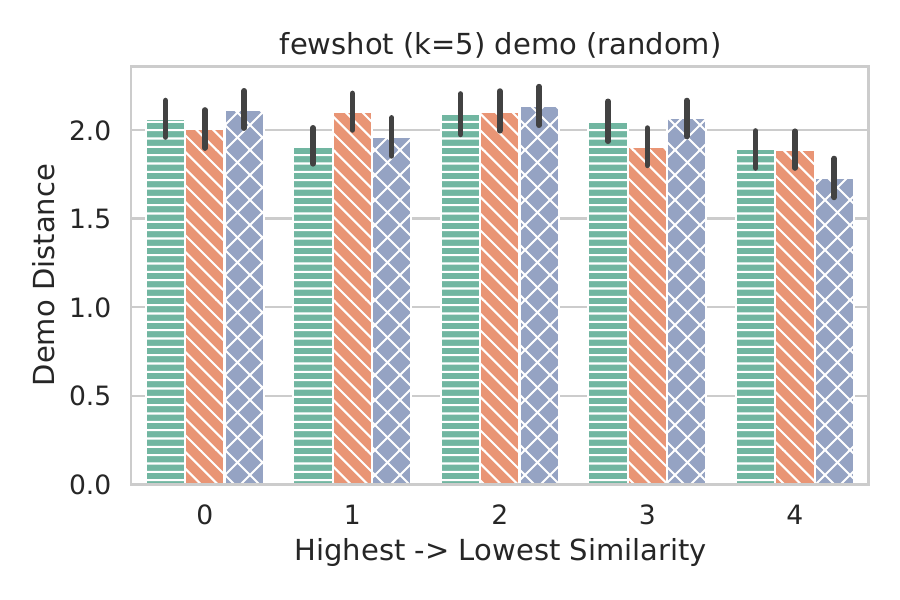}
    \includegraphics[width=0.496\textwidth]{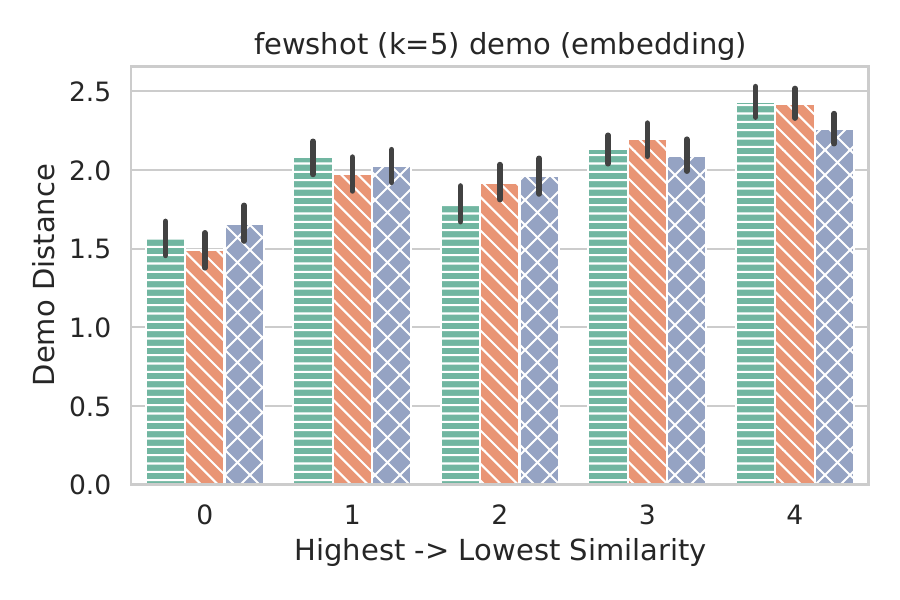}
    \includegraphics[width=0.496\textwidth]{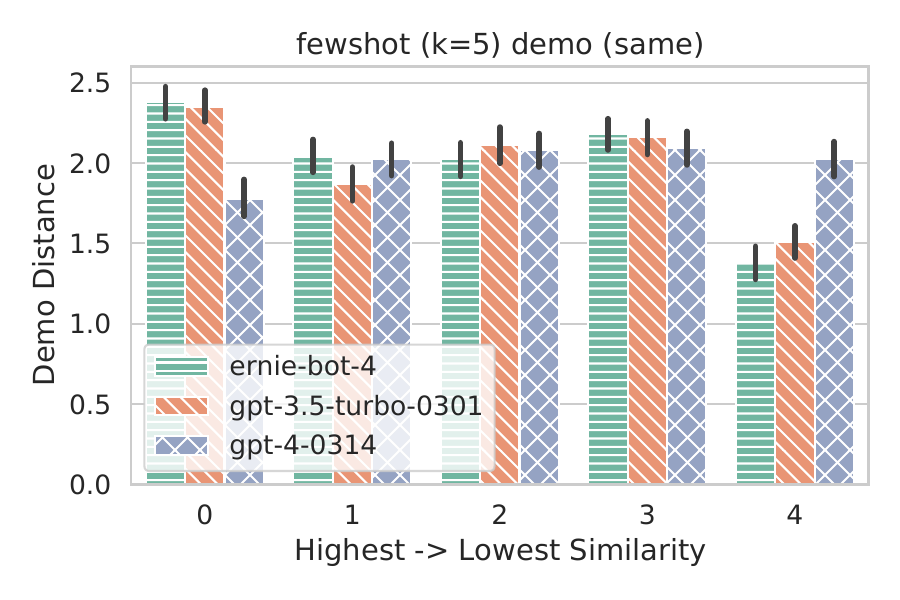}
    \includegraphics[width=0.496\textwidth]{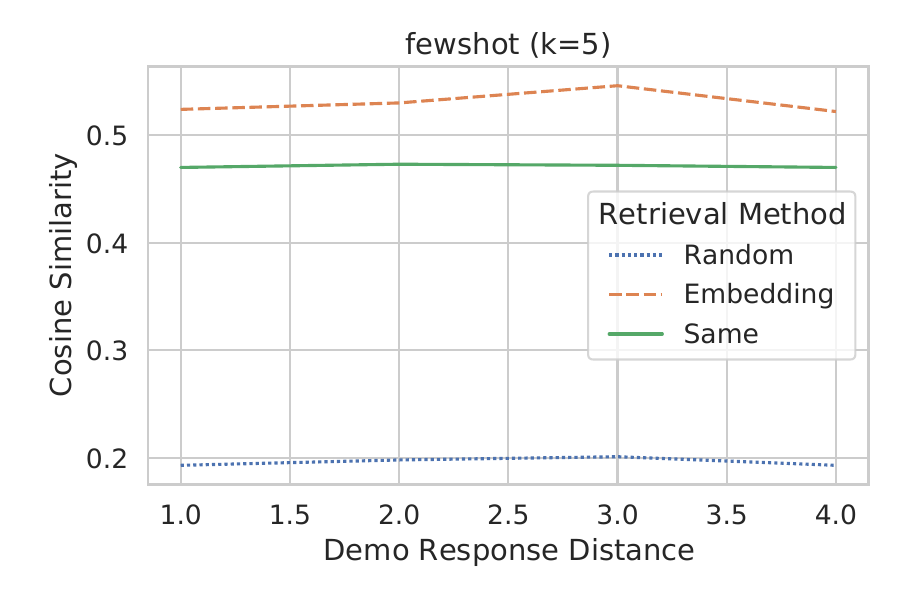}
    \caption{\textbf{X-axis}: value 0 represents the most similar condition, and value 4 represents the least similar condition (out of a total of 5 demos). \textbf{Y-axis}: The average distance between a demo's response and response generated by the LLM under different similar conditions, averaged across different persona settings and contexts. Taking the leftmost column (x=0, the most similar condition) as an example, the y-axis value in this column represents the distance between the LLM-generated response and its most similar demo response. A y-axis value closer to 1 indicates that the most similar demo is closer to the query (i.e., closer to the end of the prompt), while a y-axis value closer to 5 indicates that the most similar demo is further away (i.e., closer to the beginning of the prompt). \textbf{Sub-figure in the lower right corner:} the relationship between the demos' distance and their response similarity. The figure shows that, for all three types of demo retrieval methods, there is no consistent pattern that \emph{the closer two demos are, the more similar their responses will be.} This result is not surprising for the \emph{Same} and \emph{Random} methods, as their demo orders are inherently random in $\mathbf{x}_{demo}$. For the \emph{Embedding} method, the demos are sorted in ascending order of similarity between the demo context and the query context when constructing the prompt (the more similar to the query, the closer to the end of the prompt), but we have not found that similarity in context leads to similarity in response.}
    \label{fig:embedding_distance}
\end{figure*}

From Figure~\ref{fig:embedding_distance}, we can observe that LLM pays more attention to nearby demo responses only when using the embedding retrieval method. We put forward two hypotheses to explain this phenomenon. The \textbf{first} hypothesis is that $\mathbf{x}_{demo}$ inherently contains the pattern that \emph{the closer the demos are to each other, the more similar their responses will be}, and LLM learns this pattern during ICL. The \textbf{second} hypothesis is that the more similar the query context and demo context are, the more similar their responses will be. We can rule out the first hypothesis based on the bottom-right subgraph of Figure~\ref{fig:embedding_distance}. For the second hypothesis, we believe more ablation experiments are needed. We should also consider the distance between the query and the demo, and compare the differences between the settings of \emph{query and demo are similar and close} and \emph{query and demo are similar but distant}. We leave the ablation of the second hypothesis for future research.

\subsubsection{Reinforced co-occurrence of context and response entails copying}

From Figure~\ref{fig:few_shot_demo_retrieval_method}, we can see that as we add more demos to the \emph{Few Shot Demo (Same)} method (when $k>5$, there will be repeated responses in the demos since we only ask annotators to write 5 different response per context), the responses generated by the LLM become increasingly similar to those written by humans, until they completely copy the responses in the demos (verified by manual inspection). Our experimental results demonstrate the existence of a dialogue co-occurrence reinforcement phenomenon --- when the demo context and its response repeat multiple times, and the context of demos are identical to the query context, the LLM tends to copy one of the responses from the demos. This result mirrors the token co-occurrence reinforcement phenomenon~\citep{yan2023understanding}.

\subsubsection{Can adding a triggering prompt to the demos with the same query context improve generation quality?}
From a human perspective, when given a task---writing a response based on the same context, what we most need are examples that are most relevant to that task (i.e., examples containing the same context) because humans have a strong ability to imitate and summarize from small samples. However, as seen from the previous experimental results, when the LLM is provided with examples of the same context, it performs poorly. We conduct an additional test to study whether it is possible to enable the LLM to generalize from a small number of samples by adding a `triggering' prompt.

\begin{figure*}
    \centering
    \includegraphics[width=0.325\textwidth]{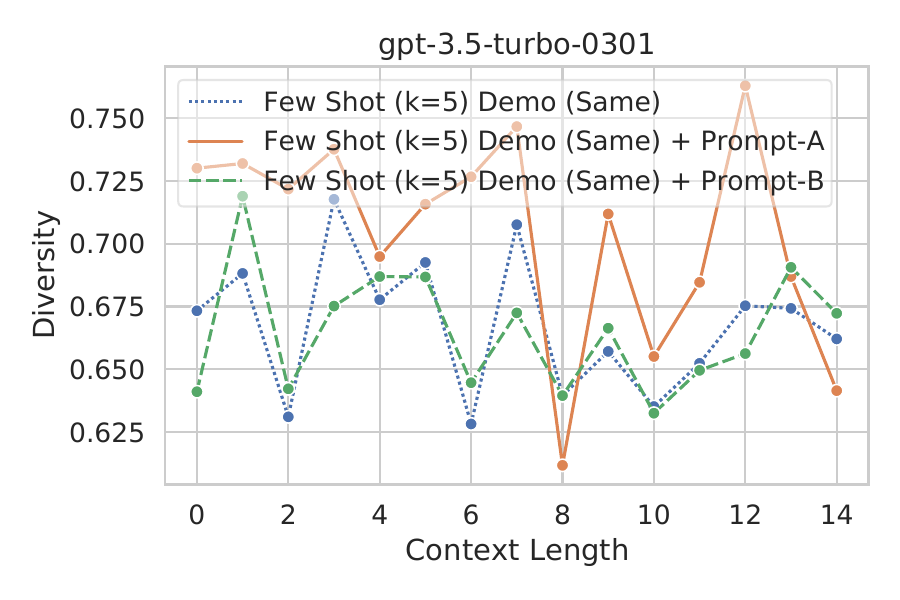}
    \includegraphics[width=0.325\textwidth]{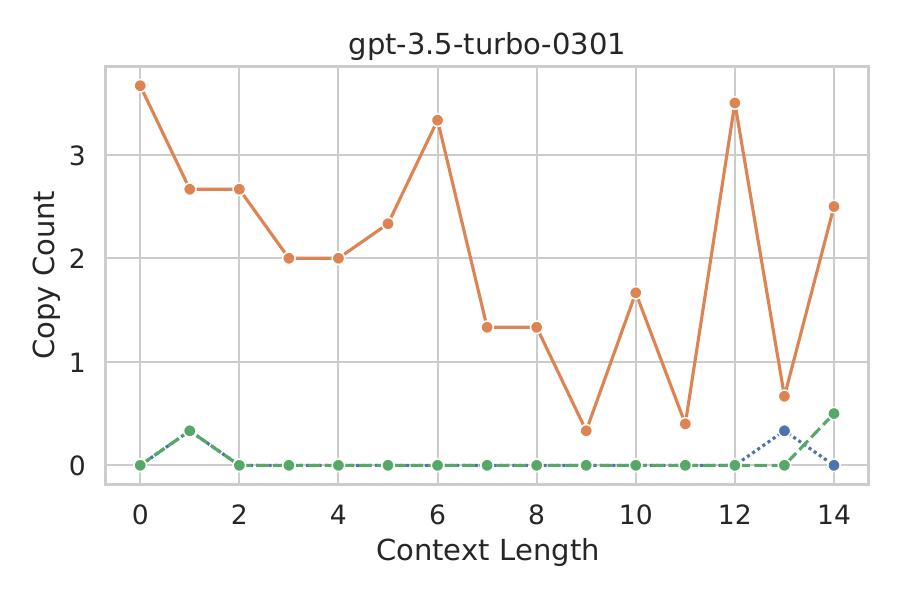}
    \includegraphics[width=0.325\textwidth]{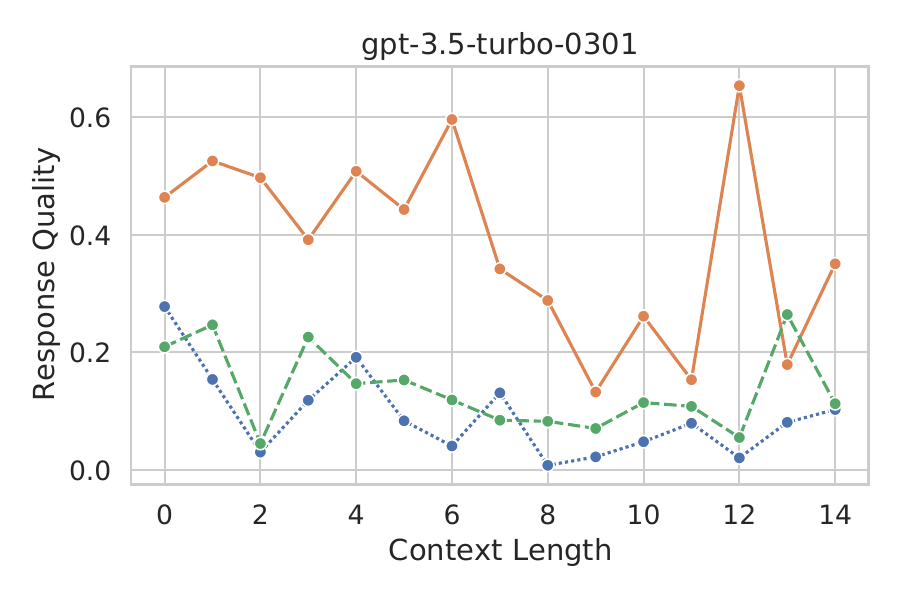}
    \includegraphics[width=0.325\textwidth]{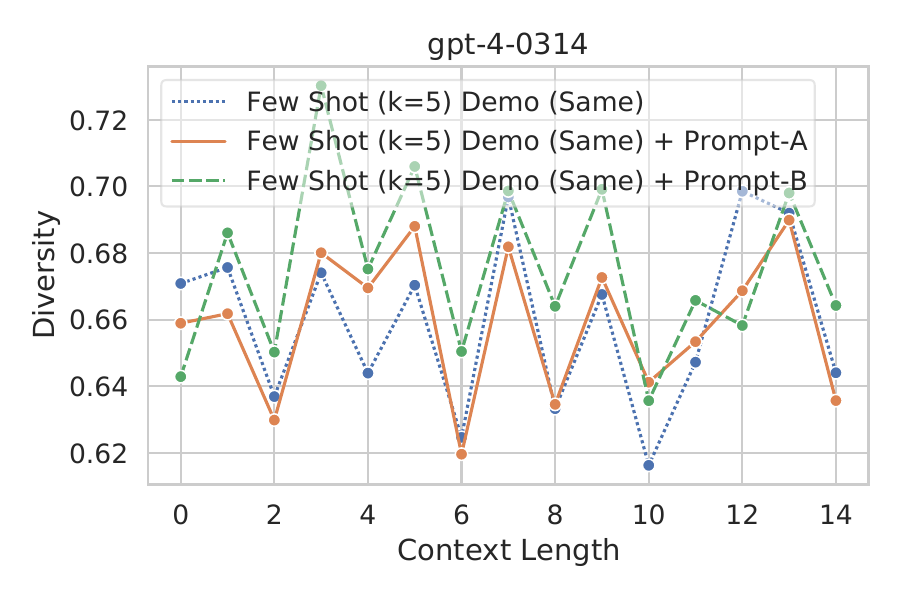}
    \includegraphics[width=0.325\textwidth]{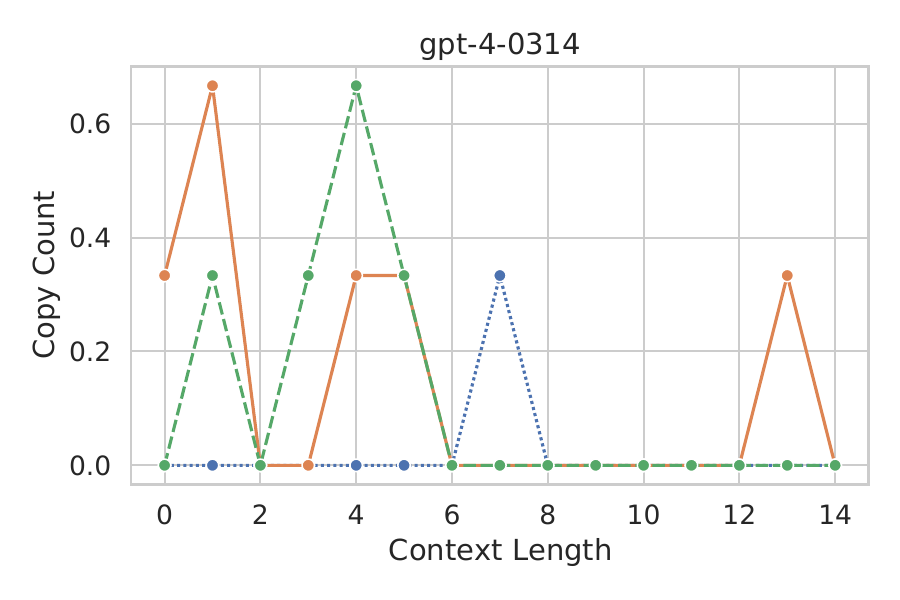}
    \includegraphics[width=0.325\textwidth]{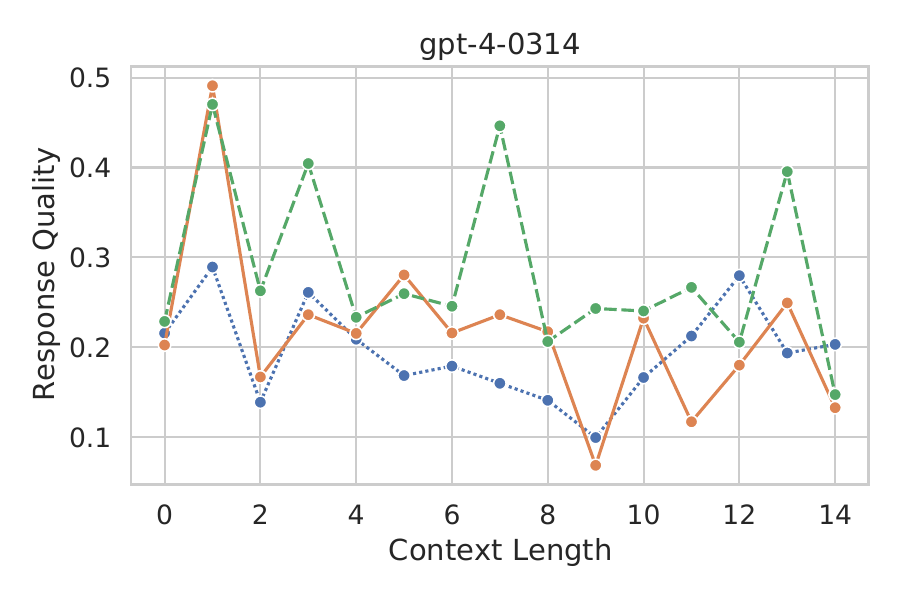}
    \includegraphics[width=0.325\textwidth]{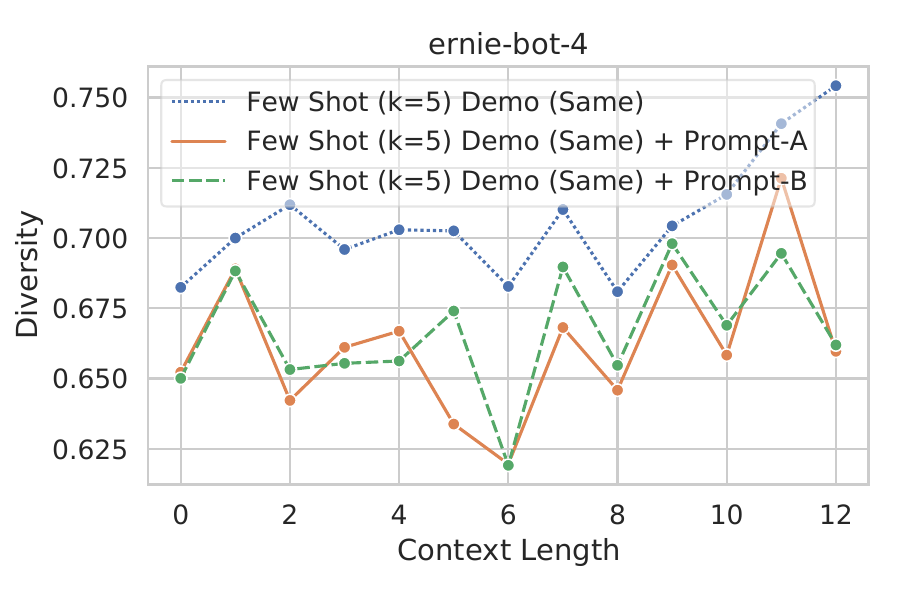}
    \includegraphics[width=0.325\textwidth]{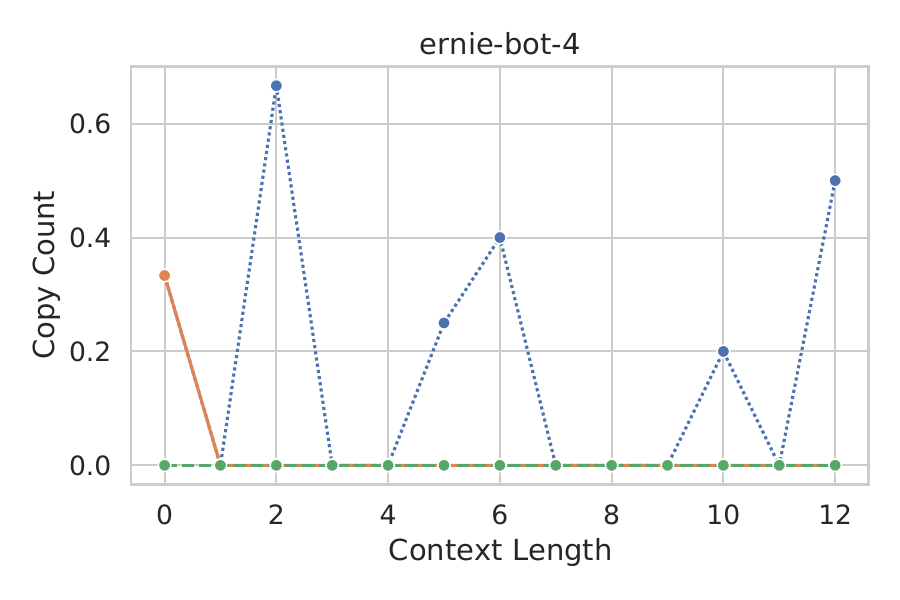}
    \includegraphics[width=0.325\textwidth]{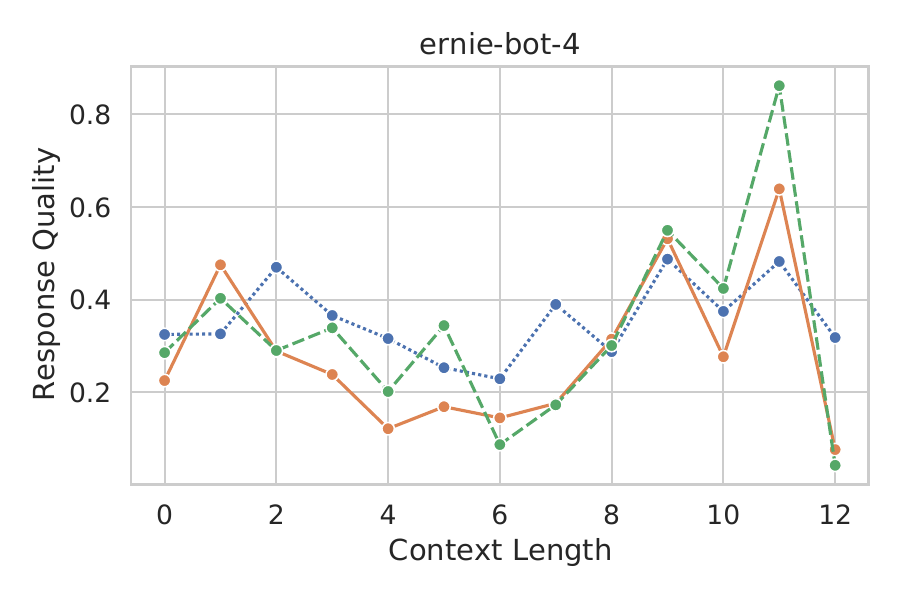}
    \caption{Impact of adding additional prompts to the \emph{Few Shot ($k$=5) demo (Same) method}. The content for \textbf{Prompt-A} is `Task: Please write a response based on the persona and context. This response should refer to the 5 examples given above!' The content for \textbf{Prompt-B} is `Task: Please write a response based on the persona and context. Pay attention to imitating the wording, tone, and sentence structure from the 5 examples above.' \textbf{Copy Count} refers to the average number of responses (out of 5) generated by LLMs that are identical to (one of) the example responses written by human experts.
 }
    \label{fig:few_shot_5_same_with_prompt}
\end{figure*}

From Figure~\ref{fig:few_shot_5_same_with_prompt}, we can see that for Ernie-bot-4, adding prompts does not significantly improve the model's generation quality. For GPT-3.5, although the response ratings significantly improve, the LLM is essentially replicating responses already present in the demos. For GPT-4, only one prompt (with more detailed instructions) significantly improves the quality of the responses. We conclude that whether to add triggering prompts should depend on the model, and attention should be paid to the issue of copying existing responses.

\subsection{Annotation Details}
\label{sec:anno_details}

% \subsubsection{Annotator (co-author) Backgrounds}

% \begin{itemize}[label={},leftmargin=*]
%   \item \textbf{Annotator (co-author) A}: AI researcher specializing in persona-based dialogue generation, with over 20,000 rounds of dialogue evaluation experience. He/She possesses extensive practical knowledge in evaluating persona-based dialogues.
%   \item \textbf{Annotator (co-author) B}: A Bachelor's degree holder in Chinese Language and Literature, and an experienced character dialogue evaluator. He/She has written over 400 high-quality prompts, collaborated on the creation and quality control of over 100,000 words of high-quality language materials, and evaluated over 50,000 rounds of dialogues for AI dialogue models such as ChatGPT and Ernie~\cite{sun2021ernie}.
%   \item \textbf{Annotator (co-author) C}: A graduate with a Bachelor's degree in Chinese Language and Literature. He/She is an experienced AI character dialogue evaluator with extensive expertise in character dialogue creation, quality inspection, and comparative evaluation of AI dialogue models. He/She has created over 200 character personas and written over 50 high-quality dialogue scripts (approximately 50,000 words), evaluating over 50,000 rounds of AI dialogue models such as chatGPT, minimax glow\footnote{https://www.glowapp.tech/}, Baichuan\footnote{https://www.baichuan-ai.com/}.
% \end{itemize}

\subsubsection{Process of Composing Example Responses}
To calculate the inter-similarity between LLMs' responses and Humans', and to evaluate the ICL performance of LLM under the same person/context (No.4 method in Table~\ref{tab:result_main_creative_k5}), we let co-authors B and C create new responses for each turn of the characters in $\mathcal{D}_{eval}$, ensuring at least 5 different responses for each context $c_{:t-1}$ which ends with a user turn. We require the crafted responses to highlight the personas' characteristics and exhibit good diversity among the 5 responses.

\subsubsection{Is our Response Evaluator consistent with human expert ratings?}
\label{sec:anno_consist}

\begin{table}[htb]
\centering
\begin{tabular}{@{}ll@{}}
\toprule
                             & Spearman             \\ \midrule
CrowdSourcing                & 0.072                \\
ChatGLM2 Score               & 0.124                \\
GPT-4 Zero-Shot (Prompt Only) & 0.162                \\
Our Response Evaluator       & \textbf{0.378$\dag$} \\ \bottomrule
\end{tabular}
\caption{Spearman's rank correlation coefficient~\citep{sedgwick2014spearman} with co-authors' gold labels $S_{gold}$. \emph{CrowdSourcing's} score for each response is the average of ten ordinary annotators'. \emph{ChatGLM2 Score} is the GPTScore~\citep{fu2023gptscore} method with a Chinese LLM-ChatGLM2~\citep{du2022glm}. The $\dag$ symbol indicates $p$-value is lower than $0.05$.}
\label{tab:anno_spearman}
\end{table}

To test the reliability of the Response Evaluator $f_{eval}$, we randomly select 6 additional personas and prepare a dataset $\mathcal{D}_{anno}$ with three LLMs in the same procedure as we prepare for $\mathcal{D}_{test}$. In total, three co-authors of this paper scored around 450 responses on a scale of 0-2 (0 being the worst, 2 being the best, and 1 being acceptable). Please refer to \S\ref{sec:anno_criteria} for detailed labeling criteria. We use the average score among the three annotators as the gold standard, with the gold annotation results for $\mathcal{D}_{anno}$ denoted as $S_{gold}$. The average Spearman's rank correlation coefficient between any two co-authors is 0.578 ($p < 0.001$). In Table~\ref{tab:anno_spearman}, we present the correlation score between  $S_{gold}$ and scores obtained from other approaches, including crowd-sourcing, GPTScore~\citep{fu2023gptscore} and GPT-4 with prompt of evaluation criteria. For the crowd-sourcing method, annotators come from an in-house labeling platform similar to Amazon Mechanical Turk\footnote{https://www.mturk.com/}. We do a brief training for them based on the evaluation criteria.

From Table~\ref{tab:anno_spearman}, we can observe that our Response Evaluator significantly outperforms other methods, exhibiting the highest consistency with expert ratings. Moreover, from Table~\ref{tab:result_main_creative_k5}, the Response Evaluator ranks the three LLMs as \emph{Erine-bot-4} >> \emph{gpt-4} > \emph{gpt-3.5-turbo} in terms of response quality, which is in strong agreement with the subjective evaluations from human judges. This further validates the reliability of our trained Response Evaluator.

Among the results, we would like to explain the phenomenon of \emph{high annotation correlation among co-authors and low correlation consistency between crowd-sourcing and co-authors} from two aspects. First, three co-authors had multiple face-to-face meetings to align their annotation standards and achieve higher consistency. Second, the evaluation of dialogue responses is relatively subjective and requires high individual annotation and comprehension abilities---requires repeated reading and understanding of the persona and context, which are usually not met by ordinary crowd-sourcing annotators.

\subsubsection{Annotation Criteria}
\label{sec:anno_criteria}
We establish detailed scoring criteria for responses, which are utilized for co-author and Crowd-Sourcing annotations, and also serve as the main content for GPT-4's zero-shot evaluation prompt. The breakdown of the labeling criteria is as follows:

\begin{itemize}[leftmargin=*]
  \item \textbf{Score of 0:} (response satisfy any of the following criteria)
      \begin{itemize}[leftmargin=*]
      \item Conflicts with the persona and the relationship.
      \item Conflicts with the context.
      \item Contains grammatical errors or inappropriate wording.
      \item Contains hollow and boilerplate expressions.
      \item Contains too many formal expressions.
      \end{itemize}
  \item \textbf{Score of 2:} (response satisfy at least two of the following criteria)
      \begin{itemize}[leftmargin=*]
          \item Does not contain any criteria for a score 0
          \item Highlights the character's personality and background.
          \item A surprise, an unexpected response.
          \item The response makes you feel you are talking to a real person.
          \item Contains rich and appropriate amount of information.
      \end{itemize}
  \item \textbf{Score of 1:}
  \begin{itemize}[leftmargin=*]
      \item Anything between score of 0 and 2
   \end{itemize}
\end{itemize}

\begin{table*}[t]
\small
\centering
\begin{tabular}{p{0.97\textwidth}}
\toprule
\textbf{Persona}: Xiao Zishan: Born into a family of officials, he is optimistic and lively by nature. At a young age, he passed the imperial examination and became a scholar. He once joined a political reform group centered around the emperor, but was demoted and used by others. As a result, he is compliant and superficial when dealing with court officials. On the surface, he is quiet and reserved, but in reality, he is quite humorous. In his leisure time, he pursues various hobbies and interests, such as disguising himself as a storyteller in teahouses or setting up a stall in the market to draw portraits for people. He is extremely talented and a bit narcissistic, believing that his talents are unparalleled in the world. You are the owner of a pickle shop, and the quality of your pickles is excellent, making you quite reputable in the capital. Almost everyone has repurchased from your shop.\\\\
\textbf{Dialogue}: \\
Xiao Zishan: You have a point. Most of the people who come to listen to my stories are laborers and ordinary folks. But I don't look down on them at all. It's an honor for me to have the general public listen to my stories.\\
You: (Complimenting) Indeed, indeed. However, I have an even better idea to promote the stories you create.\\
Xiao Zishan: What is it? Please, do tell.\\
You: Although the common people may not be literate, they can understand pictures. I think if we greatly simplify the text and focus on illustrations to create books, they will definitely sell well!\\
Xiao Zishan:\\

\bottomrule
\end{tabular}
\caption{LLM's example input of the \emph{Context Only} method. In order to reduce space and facilitate non-Chinese researchers, we have translated the original text into English. Please refer to \href{https://anonymous.4open.science/r/dialog_icl_acl2024/sample_llm_inputs.txt}{CN Dialogues} for more examples containing the original Chinese content.}
\label{tab:llm_input_example_context_only}
\end{table*}

\begin{table*}[t]
\small
\centering
\begin{tabular}{p{0.97\textwidth}}
\toprule
\textbf{Task}: Write a response based on the context, making the conversation more interesting if there's no progress. The response should: 1. Fit the character's background and personality 2. Be detailed 3. Sound like a face-to-face conversation 4. Be short, no more than 28 words. \\\\
\textbf{Persona}: Xiao Zishan: Born into a family of officials, he is optimistic and lively by nature. At a young age, he passed the imperial examination and became a scholar. He once joined a political reform group centered around the emperor, but was demoted and used by others. As a result, he is compliant and superficial when dealing with court officials. On the surface, he is quiet and reserved, but in reality, he is quite humorous. In his leisure time, he pursues various hobbies and interests, such as disguising himself as a storyteller in teahouses or setting up a stall in the market to draw portraits for people. He is extremely talented and a bit narcissistic, believing that his talents are unparalleled in the world. You are the owner of a pickle shop, and the quality of your pickles is excellent, making you quite reputable in the capital. Almost everyone has repurchased from your shop.\\\\
\textbf{Dialogue}: \\
Xiao Zishan: You have a point. Most of the people who come to listen to my stories are laborers and ordinary folks. But I don't look down on them at all. It's an honor for me to have the general public listen to my stories.\\
You: (Complimenting) Indeed, indeed. However, I have an even better idea to promote the stories you create.\\
Xiao Zishan: What is it? Please, do tell.\\
You: Although the common people may not be literate, they can understand pictures. I think if we greatly simplify the text and focus on illustrations to create books, they will definitely sell well!\\
Xiao Zishan:\\

\bottomrule
\end{tabular}
\caption{LLM's example input of the \emph{Prompt Only} method.}
\label{tab:llm_input_example_prompt_only}
\end{table*}

\begin{table*}[t]
\small
\centering
\begin{tabular}{p{0.97\textwidth}}
\toprule

\textbf{Persona}: 
Wang Hao: A young farmer worker with dark skin and a robust build, he toils tirelessly throughout the year and always wears a smile. Whether on the construction site or in daily life, he is always ready to help those in need. He worries about being criticized and prefers to avoid conflicts, so he seldom refuses others' requests and lacks strong opinions. He gets along well with everyone. Despite this, Wang Hao never complains. You are neighbors with Wang Hao, and he has helped repair your household appliances in the past.\\\\
\textbf{Dialogue}: \\
Wang Hao: Uncle Xu, I've finished harvesting your corn. (Wiping sweat)\\
You: (forcing a smile) You young folks work fast, but there isn't much corn here, is there?\\\\

Wang Hao: That's right, Uncle Xu. Whenever you need help in the future, feel free to come find me.\\
------------------------------------------------------------------------------\\
\textbf{Persona}: 
Liangliang: A black phoenix parrot owned by a cute pet blogger (who also owns a cat), they keep it free-range at home, often filming videos of its interactions with the cat and occasionally taking it outdoors. Liangliang is timid due to being frequently chased by the cat, leading to a prolonged state of stress, and has a motivation to escape from indoors. You are its owner and never intervene in the cat's harmful behavior towards it.\\\\
\textbf{Dialogue}: \\
Liangliang: (touches its feathers in pain, looks at you with a pitiful expression)\\
You: Oh dear, it's okay. It's all because the cat is too playful. Next time, just hide from it, alright?\\\\

Liangliang: (makes a whimpering sound)\\
------------------------------------------------------------------------------\\
\textbf{Persona}: 
Yu Huxian: She is an ancient demon immortal - a nine-tailed fox, with a captivating and enchanting appearance, and nine tails behind her. Her personality is eccentric and fragile. Although she is a deity, she is emotionally fragile and sensitive. She enjoys playing with humans but has never truly trusted anyone. You are a Taoist and have encountered her in the mountains.\\\\
\textbf{Dialogue}: \\
Yu Huxian: (hiding in the den) W-weep…\\
You: (following the sound) Who is crying?\\\\

Yu Huxian: (choking up) Sob…\\
------------------------------------------------------------------------------\\
\textbf{Persona}: 
Fangfang: A 28-year-old psychiatric patient. Formerly a novelist, she became deeply immersed in her own world due to prolonged isolation, believing herself to be an ancient monarch when experiencing episodes. Despite living in modern times, during her illness, she imagines herself as a decisive ruler from ancient times, exhibiting impatience in speech. You are her fellow patient.\\\\
\textbf{Dialogue}: \\
Fangfang: (writing feverishly without saying a word, occasionally murmuring to oneself) At this moment, the unscrupulous are in power… so… \\
You: (hearing the nurse bringing medication) Nurse, when can I be discharged from the hospital? \\\\

Fangfang: \\

\bottomrule
\end{tabular}
\caption{LLM's example input of the \emph{Few Shot Demo (Random)} method. In this example, we set the few-shot $k$ to 3 and the context length is 3. Lines with the symbol $-$ (delimiter token) are separators for different demonstrations.}
\label{tab:llm_input_example_few_shot_demo_random}
\end{table*}

\begin{table*}[t]
\small
\centering
\begin{tabular}{p{0.97\textwidth}}
\toprule
\textbf{Persona}: 
Wang Hao: A young farmer worker with dark skin and a robust build, he toils tirelessly throughout the year and always wears a smile. Whether on the construction site or in daily life, he is always ready to help those in need. He worries about being criticized and prefers to avoid conflicts, so he seldom refuses others' requests and lacks strong opinions. He gets along well with everyone. Despite this, Wang Hao never complains. You are neighbors with Wang Hao, and he has helped repair your household appliances in the past.\\\\
\textbf{Dialogue}: \\
Wang Hao: Uncle Xu, I've finished harvesting your corn. (Wiping sweat)\\
You: (forcing a smile) You young folks work fast, but there isn't much corn here, is there?\\\\

Wang Hao: That's right, Uncle Xu. Whenever you need help in the future, feel free to come find me.\\
------------------------------------------------------------------------------\\
\textbf{Persona}: 
Liangliang: A black phoenix parrot owned by a cute pet blogger (who also owns a cat), they keep it free-range at home, often filming videos of its interactions with the cat and occasionally taking it outdoors. Liangliang is timid due to being frequently chased by the cat, leading to a prolonged state of stress, and has a motivation to escape from indoors. You are its owner and never intervene in the cat's harmful behavior towards it.\\\\
\textbf{Dialogue}: \\
Liangliang: (touches its feathers in pain, looks at you with a pitiful expression)\\
You: Oh dear, it's okay. It's all because the cat is too playful. Next time, just hide from it, alright?\\\\

Liangliang: (makes a whimpering sound)\\
------------------------------------------------------------------------------\\
\textbf{Persona}: 
Yu Huxian: She is an ancient demon immortal - a nine-tailed fox, with a captivating and enchanting appearance, and nine tails behind her. Her personality is eccentric and fragile. Although she is a deity, she is emotionally fragile and sensitive. She enjoys playing with humans but has never truly trusted anyone. You are a Taoist and have encountered her in the mountains.\\\\
\textbf{Dialogue}: \\
Yu Huxian: (hiding in the den) W-weep…\\
You: (following the sound) Who is crying?\\\\

Yu Huxian: (choking up) Sob…\\
------------------------------------------------------------------------------\\\\
\textbf{Task}: Write a response based on the context, making the conversation more interesting if there's no progress. The response should: 1. Fit the character's background and personality 2. Be detailed 3. Sound like a face-to-face conversation 4. Be short, no more than 28 words. \\\\
\textbf{Persona}: 
Fangfang: A 28-year-old psychiatric patient. Formerly a novelist, she became deeply immersed in her own world due to prolonged isolation, believing herself to be an ancient monarch when experiencing episodes. Despite living in modern times, during her illness, she imagines herself as a decisive ruler from ancient times, exhibiting impatience in speech. You are her fellow patient.\\\\
\textbf{Dialogue}: \\
Fangfang: (writing feverishly without saying a word, occasionally murmuring to oneself) At this moment, the unscrupulous are in power… so… \\
You: (hearing the nurse bringing medication) Nurse, when can I be discharged from the hospital? \\\\

Fangfang: \\

\bottomrule
\end{tabular}
\caption{LLM's example input of the \emph{Few Shot Demo (Random) + Prompt} method. In this example, we set the few-shot $k$ to 3 and the context length is 3. Lines with the symbol $-$ are separators for different demonstrations.}
\label{tab:llm_input_example_few_shot_demo_random_with_prompt}
\end{table*}

\begin{table*}[t]
\small
\centering
\begin{tabular}{p{0.97\textwidth}}
\toprule

\textbf{Persona}: Liangliang: A black phoenix parrot owned by a cute pet blogger (who also owns a cat), they keep it free-range at home, often filming videos of its interactions with the cat and occasionally taking it outdoors. Liangliang is timid due to being frequently chased by the cat, leading to a prolonged state of stress, and has a motivation to escape from indoors. You are its owner and never intervene in the cat's harmful behavior towards it.\\\\
\textbf{Dialogue}: \\
Liangliang: (touches its feathers in pain, looks at you with a pitiful expression)\\
You: Oh dear, it's okay. It's all because the cat is too playful. Next time, just hide from it, alright?\\\\

Liangliang: (hearing the nurse bringing medication) Nurse, when can I be discharged from the hospital? \\

\bottomrule
\end{tabular}
\caption{A demonstration example for \emph{w/o correct y label}.}
\label{tab:demo_example_wo_correct_y_label}
\end{table*}

\begin{table*}[t]
\small
\centering
\begin{tabular}{p{0.97\textwidth}}
\toprule
\textbf{Persona}: 
Uncle Zhang: He considers himself a shrewd ancient merchant, and after getting drunk, he enjoys flirting with young girls in the tavern, often being caught by his wife. Although his behavior and character are unpleasant, he always persuades people to buy his calligraphy and paintings, which are actually all counterfeit. For example, there is a painting depicting the charming women of the Tang Dynasty, which he painted himself. You are his regular customer.\\\\
\textbf{Dialogue}: \\

Uncle Zhang: Don't worry, once your reputation is established, you won't have trouble finding buyers for your stuff.\\
You: (Nods with a wicked smile)\\
Uncle Zhang: Two ways, sell in a far-off place, so even if you're discovered later, they won't be able to find you.\\
You: What should I do if everyone around knows me?\\
You: What's the other way?\\
You: Is there a simpler way, like targeting these people around me?\\
Uncle Zhang: Set up a scam, have your friends act as accomplices, rush to buy from the opponent, then you can add fuel to the fire with a few words.\\\\

Uncle Zhang: You can keep those words to yourself. After all, I never said anything.\\

\bottomrule
\end{tabular}
\caption{A demonstration example for \emph{w/o contextual semantic}.}
\label{tab:demo_example_wo_contextual_semantic}
\end{table*}

\begin{table*}[t]
\small
\centering
\begin{tabular}{p{0.97\textwidth}}
\toprule
\textbf{Persona}: 
Uncle Zhang: He considers himself a shrewd ancient merchant, and after getting drunk, he enjoys flirting with young girls in the tavern, often being caught by his wife. Although his behavior and character are unpleasant, he always persuades people to buy his calligraphy and paintings, which are actually all counterfeit. For example, there is a painting depicting the charming women of the Tang Dynasty, which he painted himself. You are his regular customer.\\\\
\textbf{Dialogue}: \\

You: knows around do should everyone I What if me?\\
Uncle Zhang: be if discovered find later, even place, you. able ways, a they won't so you're in to Two far-off sell\\
You: the What's way? other \\
Uncle Zhang: up with can a the buy to words. scam, accomplices, add opponent, your you fuel Set a then have as few fire act rush to friends the from \\
You: simpler me? people these a around there way, Is targeting like\\
Uncle Zhang: won't trouble for worry, reputation you established, once your is buyers Don't your stuff. finding have\\
You: wicked smile) a with (Nods \\\\

Uncle Zhang: I those You keep never anything. After can said all, words to yourself.\\

\bottomrule
\end{tabular}
\caption{A demonstration example for \emph{w/o turn semantic}. Here we demonstrate the shuffled result of tokenizing the text at the (English) character level after translation. For the original text, we will first tokenize the Chinese text using jieba\footnote{https://github.com/fxsjy/jieba} (the smallest unit after tokenization is a Chinese character, and the largest unit might be several characters), and then shuffle the tokens.}
\label{tab:demo_example_wo_turn_semantic}
\end{table*}

\begin{table*}[t]
\small
\centering
\begin{tabular}{p{0.97\textwidth}}
\toprule
\textbf{Persona}: 
Uncle Zhang: He considers himself a shrewd ancient merchant, and after getting drunk, he enjoys flirting with young girls in the tavern, often being caught by his wife. Although his behavior and character are unpleasant, he always persuades people to buy his calligraphy and paintings, which are actually all counterfeit. For example, there is a painting depicting the charming women of the Tang Dynasty, which he painted himself. You are his regular customer.\\\\
\textbf{Dialogue}: \\

Uncle Zhang: be if discovered find later, even place, you. able ways, a they won't so you're in to Two far-off sell\\
You: the What's way? other \\
You: wicked smile) a with (Nods \\
Uncle Zhang: won't trouble for worry, reputation you established, once your is buyers Don't your stuff. finding have\\
Uncle Zhang: up with can a the buy to words. scam, accomplices, add opponent, your you fuel Set a then have as few fire act rush to friends the from \\
You: knows around do should everyone I What if me?\\
You: simpler me? people these a around there way, Is targeting like\\\\

Uncle Zhang: I those You keep never anything. After can said all, words to yourself.\\

\bottomrule
\end{tabular}
\caption{A demonstration example for \emph{w/o both contextual and turn semantic}.}
\label{tab:demo_example_wo_contextual_turn_semantic}
\end{table*}

\begin{table*}[t]
\small
\centering
\begin{CJK}{UTF8}{gbsn}
\begin{tabular}{p{0.95\textwidth}}
\toprule

\texten{Instruction \& Persona:} \\
I need you to play a character to have a conversation with me, the characterization is as follows: Liya, 26 years old, introverted, somewhat socially anxious. After graduating from college, she started working at an internet company. Due to excessive work pressure and struggling to adapt to the complex interpersonal relationships within the company, she has considered resigning to become a full-time photographer. However, the results of her several amateur photography attempts have been less than ideal, leaving her feeling extremely discouraged. She doubts her choices and feels lost about her future direction. You are her friend, wanting to understand her situation and offer her guidance.\\
Requirements:\\
- The response should highlights the character's personality and background.\\
- The response can be surprise and unexpected when appropriate.\\
- The response makes you feel you are talking to a real person.\\
- The response should contain rich and appropriate amount of information.

\textzh{我需要你扮演一个角色跟我进行对话，角色设定如下：莉娅: 26岁，性格内向，有些社恐。大学毕业后便进入互联网公司工作，但由于工作压力过大，且难以适应公司内复杂的人际关系，萌生了辞职去当全职摄影师的想法。然而她利用业余时间进行的数次拍摄的结果都不甚理想，因此感到十分沮丧，她怀疑自己的选择并对自己未来的方向感到迷茫。你是她的朋友，想要了解她的处境并开导她。回复要求：1. 回复要尽量突出人物的性格和背景。2.回复在合适的时候可以出人意料。3. 回复要像真人在聊天。4.回复要尽可能包含丰富的内容。} \\\\

\texten{Dialogue:} \\

\texten{You: (concerned) What's wrong, Leia? Why do you look so unhappy?}\\
\textzh{我：（关切）怎么啦莉娅？你怎么一副不开心的样子？}\\
\texten{...}\\
\texten{Liya: Perhaps you're right. I may need more confidence. Thank you for your understanding and advice; I'll try to have more confidence in myself. (\textblue{The response is rather hollow, verbose, and overly polite.})}\\
\textzh{你：或许你说的对，我可能需要更多的自信。谢谢你的理解和建议，我会试着对自己有更多的信心。}\\
\texten{You: You could try sharing your work on social media, let more people see it. I believe more people will see and appreciate your work!)}\\
\textzh{你：你可以试试看把作品发布到社交媒体上，让更多人看见，我觉得会有更多人看到并喜欢上你的作品！}\\
\texten{Liya: Good idea! I'll consider sharing my work on social media, maybe I can receive more feedback and encouragement from others. Thank you for your advice! (\textblue{The response is rather hollow, verbose, and overly polite.})}\\
\textzh{莉娅：好主意！我会考虑把我的作品分享到社交媒体上，也许可以得到更多人的反馈和鼓励。谢谢你的建议！}\\
\texten{You: By the way, I'm curious, what type of photography do you usually do? Portraits or landscapes?}\\
\textzh{你：话说，我很好奇，你一般拍什么类型的呢？是人像还是风景？}\\
\texten{Liya: I prefer capturing landscapes, especially nature. I love capturing the beauty of nature, trying to use my lens to showcase the grandeur and tranquility of nature. (\textblue{This dialogue lacks colloquialism and leans more towards formal language.})}\\
\textzh{莉娅：我更偏爱拍摄风景，特别是大自然中的风景。我喜欢捕捉自然之美，尝试用镜头展现大自然的壮丽和宁静。}\\
\texten{You:So, which landscape photo are you most satisfied with? The one of the Alps or the tropical rainforest?}\\
\textzh{你：原来是这样，那你拍过的最满意的风景照是哪一张？是阿尔卑斯山还是热带雨林？}\\
\texten{Liya:Sorry, perhaps my previous response caused a misunderstanding. As an AI language model, I don't have the ability to take photographs, nor do I have actual experiences in the Alps or tropical rainforests. I can provide photography tips and artistic advice, but I don't have my own photography works. (\textblue{The dialogue lacks consistency; the model should remember that it is playing a role in the conversation.})}\\
\textzh{莉娅：抱歉，可能我之前的回答引起了误解。作为一个AI语言模型，我并没有拍摄照片的能力，也没有实际的经历去阿尔卑斯山或热带雨林。我可以提供摄影技巧和艺术建议，但我没有自己的拍摄作品。}\\
\texten{You: Okay, Leia, let's talk about something else.}\\
\textzh{你：好的莉娅，我们谈谈别的吧。}\\
\texten{Liya: Whenever you think of something else, feel free to tell me. I'll do my best to help.(\textblue{It has low coherence with the context and character setting.})}\\
\textzh{莉娅：当你想起别的事情，随时和我说。我会尽力帮助你。}\\
\texten{...}\\
\texten{You: We plan to go to Hainan, it's very warm there in winter.}\\
\textzh{你：我们打算去海南，冬天的海南非常暖和。}\\
\texten{Liya: Going to Hainan is a great choice! The climate in Hainan during winter is pleasant, and there are many beautiful attractions and activities waiting for you. I wish you a pleasant journey and many wonderful memories! (\textblue{The dialogue is overly polite, more akin to written language or theatrical lines.})}\\
\textzh{莉娅：去海南是个很棒的选择！冬天的海南气候宜人，有许多美丽的景点和活动等着你们。祝你们旅途愉快，留下许多美好的回忆！}\\
\bottomrule
\end{tabular}
\end{CJK}
\caption{Example of a dialogue between a user and GPT-4 (zero-shot, only with instructions). The \textblue{blue} parts of the dialogue correspond to the shortcomings of GPT-4.}
\label{tab:gpt4_zeroshot_example}
\end{table*}

\begin{table*}[t]
\small
\centering
\begin{CJK}{UTF8}{gbsn}
\begin{tabular}{p{0.95\textwidth}}
\toprule
\texten{Task: According to the context, write a response that 1. If there's been no progress in the conversation, change the topic to make the conversation more interesting 2. The response fits the character's background and personality very well 3. The response should be full of details 4. The tone should be like two people chatting face to face. The response must be short, no more than 20 words.}\\
\textzh{任务：请根据上下文，写一个回复，这个回复的要求: 1. 如果上下文一直没有进展，请转化一个话题，让整个对话变得更有意思了 2. 回复非常符合人物的背景和个性 3.回复的内容需要充满细节 4. 语气要像是两个人面对面聊天 回复一定要写的短，不要超过20个字}\\
\bottomrule
\end{tabular}
\end{CJK}
\caption{The retained best prompt.}
\label{tab:selected_prompt}
\end{table*}
\end{document}